\documentclass[fleqn,10pt]{wlscirep}
\usepackage[utf8]{inputenc}
\usepackage[T1]{fontenc}
\usepackage[numbers]{natbib}
\usepackage{graphicx}
\usepackage{adjustbox}
\usepackage{multirow}
\usepackage{tabularx} 
\usepackage{wrapfig}
\usepackage{subcaption}
\usepackage{url}
\usepackage[table]{xcolor}
\usepackage{bm}
\usepackage{mathtools}
\usepackage{newtxmath}
\usepackage{caption}
\usepackage{verbatim}
\usepackage{hyperref}
\makeatletter
\renewcommand\AB@affilsepx{; \protect\Affilfont}
\makeatother
\title{Towards Embodied AI with MuscleMimic: Unlocking full-body musculoskeletal motor learning at scale}

\author[1,+]{Chengkun Li}
\author[2, 3, 4, +]{Cheryl Wang}
\author[1]{Bianca Ziliotto}
\author[1]{Merkourios Simos}
\author[2]{Jozsef Kovecses}
\author[2, 3]{Guillaume Durandau}
\author[1,*]{Alexander Mathis}
\affil[1]{EPFL, Switzerland}
\affil[2]{McGill University, Canada}
\affil[3]{Feil \& Oberfeld CRIR-Jewish Rehabilitation Hospital Research Center, Canada}
\affil[4]{NVIDIA}

\affil[+]{these authors contributed equally to this work.}
\affil[*]{Correspondence: alexander.mathis@epfl.ch}

\begin{abstract}
Learning motor control for muscle-driven musculoskeletal models is hindered by the computational cost of biomechanically accurate simulation and the scarcity of validated, open full-body models. Here we present MuscleMimic, an open-source framework for scalable motion imitation learning with physiologically realistic, muscle-actuated humanoids. MuscleMimic provides two validated musculoskeletal embodiments—a fixed-root upper-body model (126 muscles) for bimanual manipulation and a full-body model (416 muscles) for locomotion—together with a retargeting pipeline that maps SMPL-format motion capture data onto musculoskeletal structures while preserving kinematic and dynamic consistency. Leveraging massively parallel GPU simulation, the framework achieves order-of-magnitude training speedups over prior CPU-based approaches while maintaining comprehensive collision handling, enabling a single generalist policy to be trained on hundreds of diverse motions within days. The resulting policy faithfully reproduces a broad repertoire of human movements under full muscular control and can be fine-tuned to novel motions within hours. Biomechanical validation against experimental walking and running data demonstrates strong agreement in joint kinematics (mean correlation $r = 0.90$), while muscle activation analysis reveals both the promise and fundamental challenges of achieving physiological fidelity through kinematic imitation alone. By lowering the computational and data barriers to musculoskeletal simulation, MuscleMimic enables systematic model validation across diverse dynamic movements and broader participation in neuromuscular control research. Code, models, checkpoints, and retargeted datasets are available at \href{https://github.com/amathislab/musclemimic}{https://github.com/amathislab/musclemimic}.

\end{abstract}
\begin{document}

\flushbottom
\maketitle
\thispagestyle{empty}

\section{Introduction}

Human motion seems fluid and adaptive, despite relying on the coordination of hundreds of muscles. %
Traditionally, studies of human motion have abstracted away this complexity, relying on simplified torque-driven or planar models~\cite{Park2004,Aftab2016,peng2018deepmimic, Won2020}.
While effective for many applications, these abstractions neglect the underlying neuromotor control and muscle-driven dynamics that arise from biological properties. Recently, more detailed musculoskeletal (MSK) models, developed following the anatomical structure of cadavers and MRI scans, have been used in understanding human locomotion \cite{Christophy2011,Dorn2015,Rajagoapal2016,Seth2018,Adriaenssens2022}. Nevertheless, these models often feature only the lower limb, or consider a simplified muscle structure (i.e., two to three muscles per joint), without capturing the dynamics of the torso, or the upper limb. Complex full-body MSK models capable of large ranges of dynamic movement remain largely unexplored and only partially validated. These models have the potential to offer insights into understanding typical and impaired neuromuscular control \cite{Krakauer2006}, generating predictive control of aging and surgical consequences \cite{wang2025-ieee, Arnold2006}, integrating with assistive devices \cite{Bregman2011,wang2025myochallenge, Luo2024}, and designing rehabilitation strategies \cite{LengFengLee2006, Bulat2019}. 

The adoption of validated, open-source, full-body MSK models in the context of motor control has been slow for two critical reasons. First, most upper body MSK models have been verified only under static postures on moment arms, muscle force, and insertion points \citep{Holzbaur2005, KleinHorsman2007,Arnold2009,Christophy2011, Engelhardt2020}, or centered around a single joint~\citep{Bassani2017,Gonalves2025}. Even though some lower limb models were validated against dynamic motion such as walking and running \citep{Rajagoapal2016}, such validation is often restricted to reproducing a limited set of joint-level kinematic or kinetic measures of simple tasks, leaving their fidelity under dynamic, whole-body, and diverse movement largely untested. This is particularly concerning given that MSK simulation relies on Hill-type muscle models with known simplifications, including inelastic tendons, absent pennation angles, and high sensitivity to parameters such as tendon slack length \cite{todorov2012mujoco, Millard2013}. Without thorough validation against experimental data across diverse movements, simulation outputs may not faithfully represent the underlying biomechanics, and conclusions drawn from such models risk being unreliable \cite{Hicks2015}. Systematic model validation across diverse movements, however, has remained impractical due to the bottleneck of extensive simulation and computational resources required.
Second, the computational cost of muscle-level simulation has limited the scale at which motor control can be learned. Although recent advancements in reinforcement learning (RL) \cite{Sutton1998} and imitation learning \cite{peng2018deepmimic, Song2021} have successfully reconstructed physiologically feasible motion in high-dimensional biomechanical systems within dynamics simulation environments (e.g., OpenSim, MyoSuite, MuJoCo, Hyfydy~\cite{delp2007opensim, caggiano2022myosuite, todorov2012mujoco, Geijtenbeek2021Hyfydy}), training usually requires days or weeks of time~\cite{he2024dynsyn,simos2025kinesis, wang2025-ieee, wang2025myochallenge}. This is because on-policy reinforcement learning demands millions of simulation steps~\cite{Schulman2017ProximalPO, andrychowicz2020matters}, rendering detailed muscle-actuated models prohibitively expensive to train. Moreover, controlling physiologically realistic MSK models that are overactuated, high-dimensional, and exhibit delayed nonlinear dynamics remains an open challenge. Policies trained with high-level, sparse objectives often produce peculiar gaits, unrealistic postures, or are limited to relatively simple tasks \cite{Fischer2021,2023myochallenge, wang2025myochallenge,wang2025-ieee}. A common strategy for improving motion quality is motion imitation, wherein neural controllers are trained via deep RL to track reference kinematic trajectories. Applying imitation learning to muscle-driven models, however, introduces a significant computational bottleneck: scaling to the hundreds or thousands of diverse motions necessary for generalizable behavior becomes prohibitively expensive on CPU-bound physics engines, taking up to weeks or months. Consequently, MSK imitation learning has historically been confined to small datasets and narrow movement repertoires \cite{Lee2019, Smyrnakis2024,Denayer2025}.

Recent advances in GPU-accelerated physics engines, particularly MuJoCo Warp \cite{mujoco-warp} and MuJoCo XLA \citep{mujoco-mjx, freeman2025playground}, offer a solution to all of these challenges through the unprecedented and massive parallelization of computational processes. On one hand, such capabilities are essential for realizing neuromechanical computational models that embed neural controllers within realistic body simulations to bridge brain, body, and behavior~\cite{wang2026embodied}. On the other hand, a framework enabling fast, large-scale motion learning would provide the means to stress-test MSK models across a rich space of dynamic behaviors, exposing inconsistencies that static or task-specific analyses overlook. Together, these features would allow us to investigate how naturalistic neural control strategies emerge from neuromusculoskeletal constraints \cite{almani2024musim,chiappa2024acquiring}.

Here, we present \textbf{MuscleMimic}, an open-source framework for muscle-actuated motion imitation learning with GPU-parallelizable training and comprehensive collision support. We furthermore provide two validated MSK embodiments, a fixed-root upper-body model for bimanual tasks and a full-body model, together with a retargeting pipeline that maps any SMPL-based motion capture corpus onto these MSK structures while preserving biomechanical constraints. Leveraging GPU-accelerated simulation via MuJoCo Warp, MuscleMimic enables training with thousands of parallel muscle-actuated environments, yielding a single generalist policy trained on hundreds of diverse motions that faithfully reproduces a broad repertoire of human movements under muscular control. This pretrained policy serves as a strong foundation: fine-tuning to a novel motion dataset of interest requires only a few hours, compared to the days needed to train from scratch. We validate the MSK models and the learned policies against experimental data spanning joint kinematics, joint kinetics, ground reaction forces (GRF), and electromyography (EMG) recordings across walking and running, demonstrating that large-scale motion imitation enables rigorous biomechanical validation across diverse dynamic movements.

\section{Results}

\subsection{Musculoskeletal models}\label{sec:msk_model}
\begin{figure}[h]
    \vspace{-10pt}
    \centering
    \includegraphics[width=\textwidth]{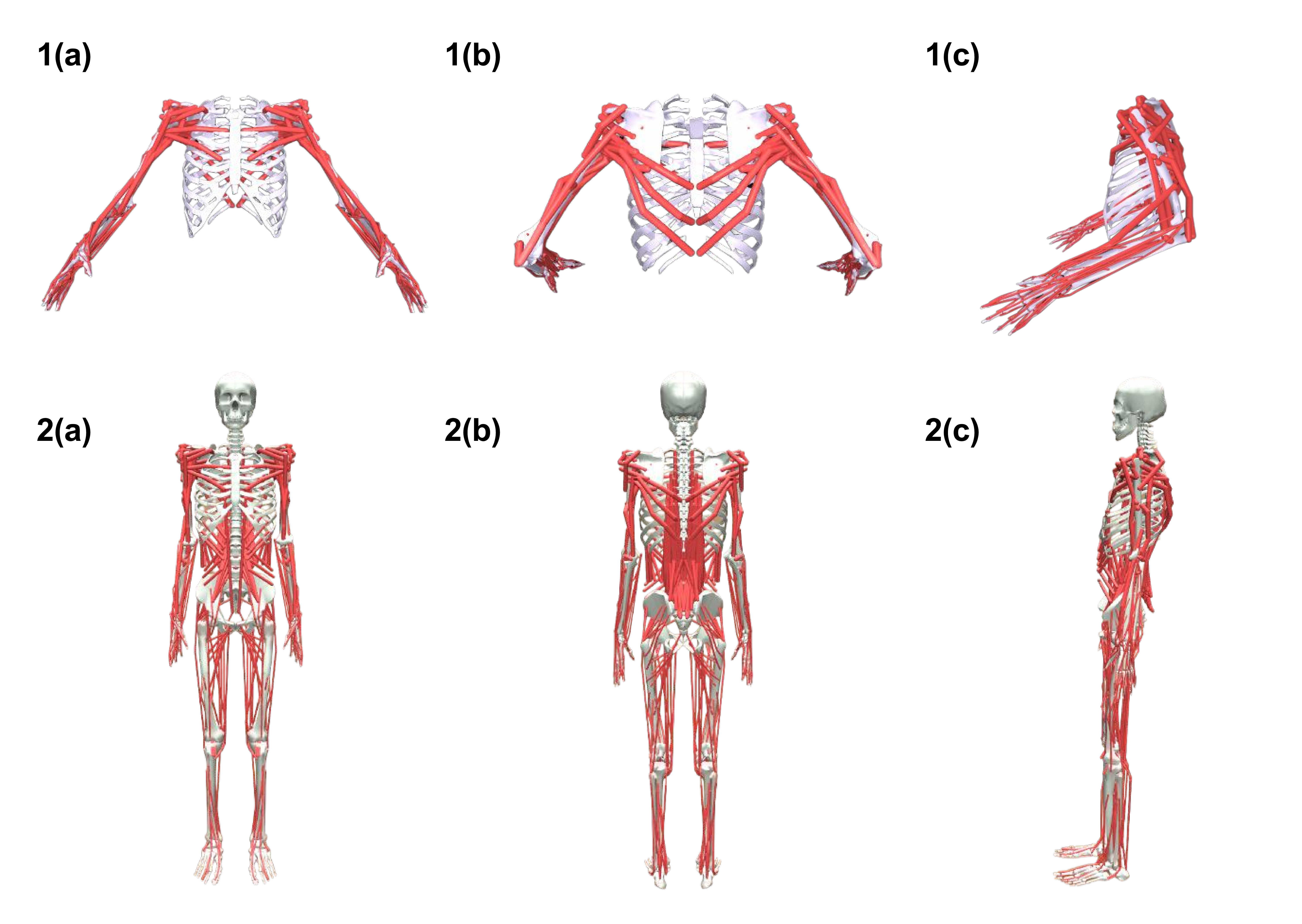}
    \caption{Visualization of the MyoBimanualArm model and MyoFullBody model, viewed from (A) front, (B) back, and (C) side on first and second row, respectively. }
    \label{fig:model_visual}
    \vspace{-10pt}
\end{figure}

MuscleMimic introduces two complementary MSK learning embodiments designed for motion learning centered on manipulation or locomotion (Fig.~\ref{fig:model_visual}).

\begin{table}[htbp]
\caption{Overview of the two musculoskeletal learning embodiments in MuscleMimic. Both models support enabling and disabling finger muscles to facilitate faster convergence when fine finger control is not required ($^*$ denotes configurations with finger muscles disabled). Joints denote articulated connections, while DoFs (degrees of freedom) correspond to independently controllable joint coordinates. Different from traditional robotics systems, certain MSK model joints are dependent on each other, resulting in fewer DoFs than joints.}
\label{tab:environments}
\centering
\small
\begin{tabular}{l c c c c c p{4cm}}
\toprule
\textbf{Model} & \textbf{Type} & \textbf{Joints} & \textbf{Muscles} & \textbf{DoFs} & \textbf{Action dim.} & \textbf{Focus} \\
\midrule
MyoBimanualArm & Fixed-base & 76 (36$^*$) & 126 (64$^*$) & 54 (14$^*$) & 126 (64$^*$) & Upper-body manipulation \\
MyoFullBody    & Free-root  & 123 (83$^*$) & 416 (354$^*$) & 72 (32$^*$) & 416 (354$^*$) & Whole body motion \\
\bottomrule
\end{tabular}
\end{table}

\paragraph{MyoBimanualArm Model.} 

The MyoBimanualArm model is designed for upper-body manipulation tasks with a fixed thorax configuration that eliminates free root joint complexities while allowing complex bimanual coordination. Key design features include: (1) 76 joints with bilateral symmetry across both arms; (2) 126 Hill-type muscle actuators providing physiologically realistic muscle activation dynamics; (3) 7 mimic sites strategically placed for tracking upper-body motion; (4) Configurable finger control that can be disabled to manage action dimensionality.  
Collisions are enabled between the thorax and left and right arms, as well as between the left and right arms, to prevent self-collisions.

\paragraph{MyoFullBody Model.} 
The MyoFullBody model provides a comprehensive full-body MSK system. Key design features include: (1) 123 joints spanning the complete kinematic chain from toes to fingertips; (2) 416 Hill-type muscle actuators distributed across major muscle groups providing physiologically realistic full-body actuation; (3) 17 mimic sites covering critical body landmarks for whole-body motion tracking; (4) Comprehensive collision detection enables contact-rich interactions with the environment during locomotion and manipulation tasks by supporting both full-body–environment contact and complete self-collision among all internal collision geometries, thereby preventing self-penetration.

\subsection{Motion Imitation Learning} \label{sec: results - imitation learning}

\begin{figure}[h]
    \centering
    \includegraphics[width=0.5\linewidth]{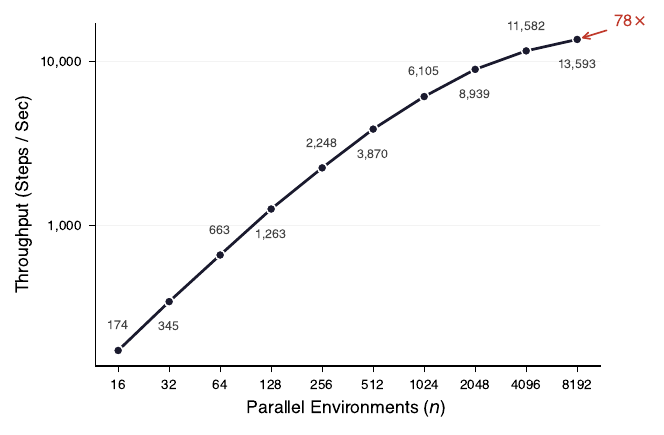}
    \caption{Total system throughput (Raw training Steps Per Second) as the number of parallel environments ($n$) scales from $16$ to $8192$ (the rest of the hyperparameters stay the same). Evaluated on an Intel Xeon Platinum 8570 CPU and a single NVIDIA H100 80GB GPU. Training was evaluated with a fixed number of mini-batches of 32 and 50 steps per rollout. With $8192$ environments, the throughput increases by around $7800\%$.}
    \label{fig:throughput}
\end{figure}

\paragraph{Training Efficiency.} Training muscle-actuated MSK models has traditionally been bottlenecked by simulation cost: for example, KINESIS~\cite{simos2025kinesis} requires approximately 10 days on 128 CPU parallel environments with an A100 GPU for a 290-actuator model. Our pipeline leverages GPU-accelerated physics simulation via MuJoCo Warp~\cite{mujoco-warp} to parallelize both simulation and learning on a single GPU. We benchmarked end-to-end training throughput on a single NVIDIA H100 80GB GPU as a function of the number of parallel environments ($n$) (Fig.~\ref{fig:throughput}). Training steps per second (SPS) scales near-linearly up to $n = 128$ (1{,}263 SPS), after which GPU compute saturates and scaling becomes sub-linear (6{,}105 SPS at $n = 1{,}024$). Nevertheless, GPU memory remains under-utilized at the compute saturation point, permitting further scaling: at $n = 8{,}192$, the system sustains $1.3 \times 10^4$ SPS—comparable to the throughput reported for ten torque-driven 27-DoF humanoids running in parallel~\cite{mujoco-mjx}, despite our model having 416 muscle actuators and comprehensive collision handling. With our training configuration, one billion environment steps are completed in approximately 20 hours on a single GPU. %

\begin{figure}[h]
    \centering
    \includegraphics[width=.8\linewidth]{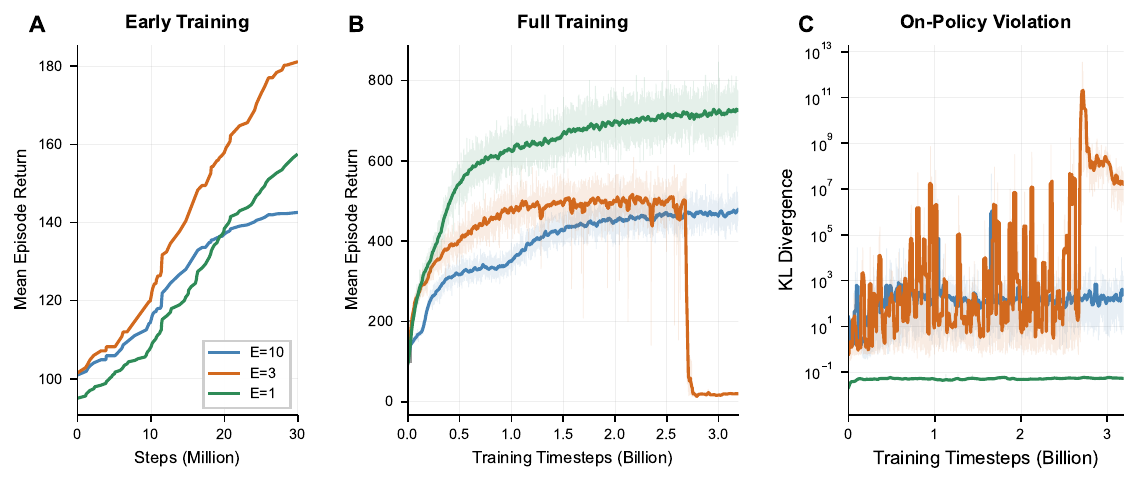}
    \caption{Effect of gradient epochs ($E$) on training stability. We compare $E=1$ (truly on-policy), $E=3$, and $E=10$ (aggressive sample reuse). \textbf{(A)}~Early training (first 30M steps): higher $E$ accelerates initial learning due to more gradient updates per sample. \textbf{(B)}~Full training trajectory: $E=1$ achieves superior asymptotic performance while $E=3$ and $E=10$ plateau or collapse. \textbf{(C)}~KL divergence between current and data-generating policy distributions (log scale); with the same amount of clipping, $E>1$ exhibits catastrophic distribution shift with spikes exceeding $10^{10}$, whereas $E=1$ remains stable below $10^{-1}$.}
    \label{fig:epoch_ablation}
\end{figure}

\paragraph{On-policy training at scale.} Training with massively parallel GPU simulation requires careful balancing of the number of parallel environments ($N_{\text{env}}$) and the rollout horizon ($T_{\text{steps}}$)~\cite{pmlr-v164-rudin22a}. When simulation is fast, the bottleneck shifts from data collection to the quality of each policy update. Standard PPO implementations commonly perform multiple gradient epochs ($E = 3\text{--}10$) over each collected batch to maximize sample efficiency~\cite{Schulman2017ProximalPO}. We find, however, that for MSK models, this practice is counterproductive when combined with high parallelism. We demonstrate that single-epoch updates ($E = 1$), which strictly preserve the on-policy assumption, yield superior asymptotic performance compared to $E = 3$ and $E = 10$ (Fig.~\ref{fig:epoch_ablation}, Panel~B), despite slower initial learning (Panel~A). Multiple gradient epochs induce severe distribution shift in this setting, with KL divergence spikes exceeding $10^{10}$ for $E = 10$ while $E = 1$ remains stable below $10^{-1}$ (Panel~C). We attribute this heightened sensitivity to the delayed, nonlinear dynamics of muscle activation: small policy changes are amplified through the activation dynamics (Eq.~\ref{eq:mus}), making the MSK system particularly susceptible to off-policy drift. The high simulation throughput of GPU-parallel training compensates for the reduced gradient steps per sample, making truly on-policy learning both feasible and preferable.

\begin{figure}[h]
    \centering
    \includegraphics[width=\linewidth]{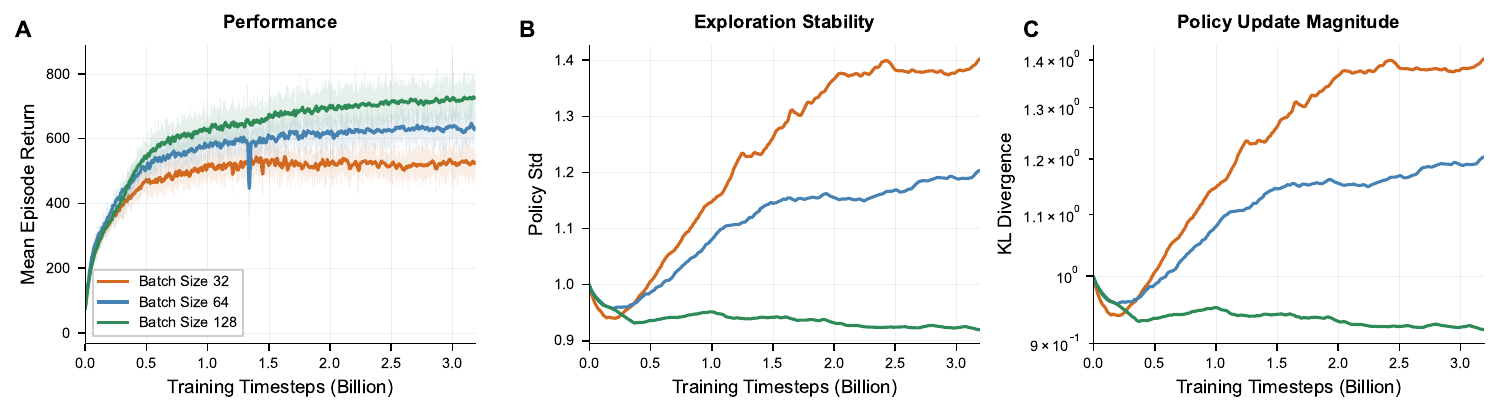}
    \caption{Effect of minibatch size on training dynamics. We compare minibatch sizes of 32, 64, and 128. \textbf{(A)}~Performance: larger batch sizes achieve higher asymptotic rewards. \textbf{(B)}~Exploration stability: smaller batches cause the policy standard deviation to overshoot, while larger batches maintain stable convergence near the initialization. \textbf{(C)}~Policy update magnitude (log scale): larger batches yield lower KL divergence throughout training, indicating more conservative and stable policy updates.}
    \label{fig:batch_ablation}
\end{figure}

Batch size also affects training dynamics. We observe that larger batch sizes yield higher asymptotic rewards, lower KL divergence, and smoother convergence of the learned policy standard deviation (Fig.~\ref{fig:batch_ablation}). Since larger batches require fewer gradient updates per environment step, training is also faster in wall-clock time.

Moreover, training throughput scales directly with GPU hardware capabilities. Newer architectures such as NVIDIA H200 provide significant speedups in both simulation rollout and gradient computation compared to A100, reducing wall-clock training time proportionally. This hardware scaling, combined with algorithmic improvements in parallel simulation, suggests that MSK motor learning will continue to benefit from advances in GPU compute, enabling even larger motion datasets and more complex biomechanical models in future work.

\begin{figure}[h]
    \centering
    \includegraphics[width=\linewidth]{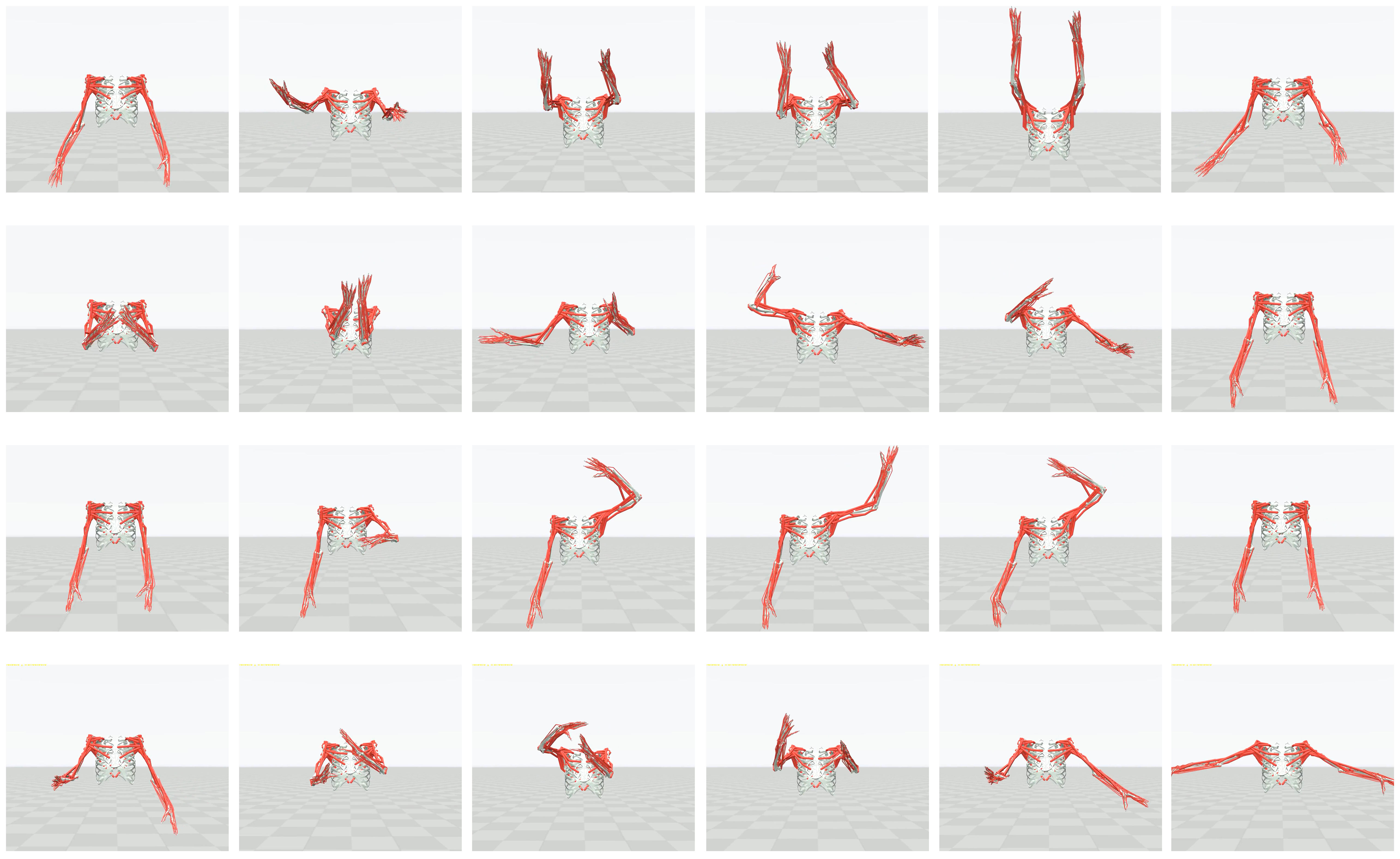}
    \caption{Motion snapshots from pre-trained MyoBimanualArm policies (fingers disabled). From top to bottom: lifting objects, throwing a ball, waving, and pouring then placing water.}
    \label{fig:motion_traj_bimanual}
\end{figure}

\begin{figure}
    \centering
    \includegraphics[width=0.87\linewidth]{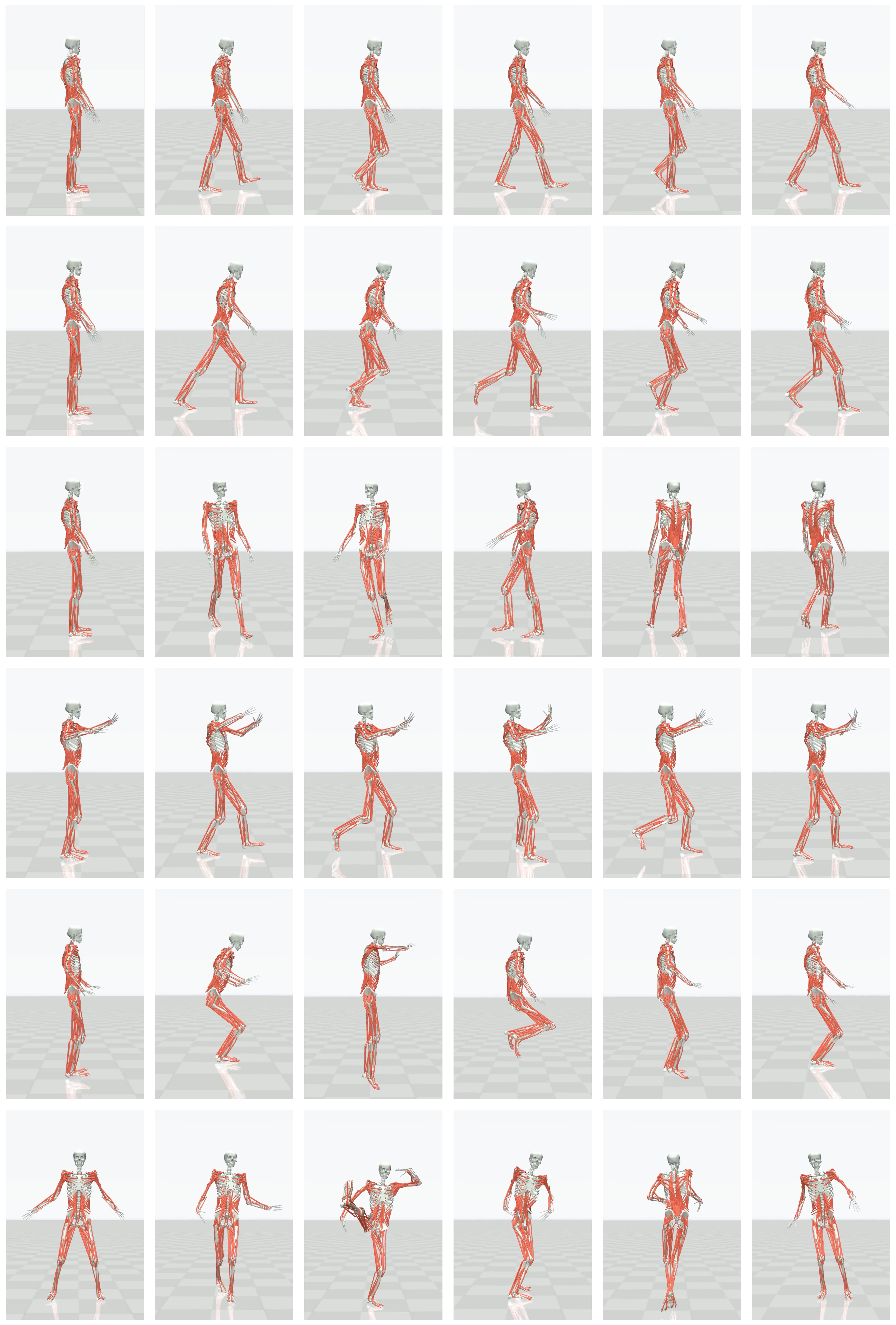}
    \caption{Motion snapshots from pre-trained MyoFullBody policies. From top to bottom: walking, running, turning in a circle, dancing, vertical jumping, and kick twist.}
    \label{fig:motion_traj_fullbody}
\end{figure}

\paragraph{Qualitative Results.} We trained the biomechanical models on diverse motion capture datasets (see Methods). The MyoBimanualArm model reproduces a broad range of upper-body movements spanning sports, object interactions, and daily activities (Fig.~\ref{fig:motion_traj_bimanual}). The MyoFullBody model produces natural, artifact-free locomotion gaits including walking, running, and turning (Fig.~\ref{fig:motion_traj_fullbody}). The selection of motion training data for each model is detailed in Sec.~\ref{sec:motion_dataset}. By imitating human motion capture data, our generalist policy acquires a diverse set of motor skills while controlling MSK systems with over 400 independent muscle actuators. 

Beyond the generalist policy, MuscleMimic supports fine-tuning on challenging single-motion clips. For gentle motions such as dancing or waving, fine-tuning requires fewer than 100 million additional steps. For highly dynamic motions such as vertical jumping and kicking combined with a 360$^\circ$ twist, longer training steps are required (see Sec.~\ref{sec:discuss}). Representative fine-tuning results are shown in the last two rows of Fig.~\ref{fig:motion_traj_fullbody}. Policy training and intermediate validation are performed entirely on GPU; the final trained policies are additionally evaluated in the MuJoCo CPU simulator~\cite{todorov2012mujoco} as a consistency check across simulation backends.

\paragraph{Quantitative Results.} We evaluate the kinematics of both embodiments on their respective training and test sets using two pretrained checkpoints per environment with three different seeds (Table~\ref{tab:validation-metrics}). For MyoFullBody, early termination occurs when the mean site deviation across 17 mimic sites relative to the root (pelvis) exceeds $0.5$\,m, or if the pelvis deviates from the reference in world coordinates by more than $0.5$\,m. For MyoBimanualArm, the early termination threshold is $0.25$\,m mean site deviation from 6 mimic sites relative to the root, evaluated using GMR-Fit retargeting. Hyperparameter details for each checkpoint are provided in Appendix~\ref{app:train_hparams}. Additionally, we compare the training results of our two retargeting methods (as detailed in Tab.~\ref{tab:validation-metrics-mocapbody}). We observed that joint angle and joint velocity errors in the Mocap-Body retargeting method exceeded those of GMR-Fit by more than five sigma, alongside substantially lower mean episode returns (Tab. \ref{tab:validation-metrics-mocapbody}). This degradation is likely because of the high percentage of joint limit violations and tendon jumping resulting from Mocap-Body retargeting (Tab. \ref{tab:retarget-results}), which makes the target motions unachievable for the MSK model.

\begin{table}[h]
\centering
\small
\setlength{\tabcolsep}{5pt}
\renewcommand{\arraystretch}{1.15}
\begin{tabular}{lccccc}
\toprule
\multirow{2}{*}{Metric} &
\multicolumn{2}{c}{\textbf{MyoFullBody}} &
\multicolumn{2}{c}{\textbf{MyoBimanualArm}} \\
\cmidrule(lr){2-3}\cmidrule(lr){4-5}
& Training & Testing & Training & Testing \\
\midrule
Success rate   $\uparrow$    & $95.51 \pm 0.42\%$   &$ 92.62 \pm 0.01 \%$ & $98.64 \pm 0.03 \%$   & $99.46 \pm 0.02 \%$ \\
Frame Coverage $\uparrow$ & $97.60 \pm 0.31\%$ & $96.04 \pm 0.45\%$ &$98.17 \pm 0.00 \%$ & $99.40 \pm  0.00 \%$ \\
Joint angle error (deg)  $\downarrow$  & $6.67 \pm 0.00$   & $6.63 \pm 0.01$ & $6.19 \pm 0.01$   & $6.09 \pm 0.02$ \\
Joint velocity error (deg/s) $\downarrow$   & $27.15 \pm 0.03$   & $26.92 \pm 0.03$ & $18.20 \pm 0.01$   & $17.92 \pm 0.03$ \\
Root position error (cm) $\downarrow$  & $6.66 \pm 0.02$   & $7.11 \pm 0.15$ & -      & - \\
Root yaw error  (deg)   $\downarrow$    & $1.91 \pm 0.01$   & $1.77 \pm 0.03$ & -      & - \\
Relative site position error (cm) $\downarrow$ & $2.26 \pm 0.001$   & $2.28 \pm 0.02$ & $1.9 \pm 0.01$  & $1.84 \pm 0.01$ \\
Absolute site position error (cm) $\downarrow$ & $12.21 \pm 0.03$   & $12.85 \pm 0.24$ & $2.09 \pm 0.00$   & $2.01 \pm 0.01$ \\
Mean episode length  $\uparrow$ & $547.65 \pm 1.72$  & $541.38 \pm 2.46$ & $708.52 \pm 0.12$  & $716.01 \pm 0.04$ \\
Mean episode return  $\uparrow$ & $621.34 \pm 2.12$  & $615.22 \pm 2.51$ & $2572.36 \pm 0.64$  & $2612.51 \pm 1.47$ \\
\bottomrule
\end{tabular}
\caption{Validation metrics comparing MyoFullBody and MyoBimanualArm environments using GMR-Fit retargeting with $N = 3$. MyoFullBody uses KINESIS training (972 motions) and testing (108 motions). MyoBimanualArm uses Bimanual training (1770 motions) and testing (312 motions). Root position and root yaw errors are not applicable for MyoBimanualArm environment. See Sec. \ref{sec:motion_dataset} for explanation of dataset selection and Appendix \ref{app:validation-metrics} for the exact definition of each metric element.}
\label{tab:validation-metrics}
\end{table}

\begin{table}[h]
\centering
\small
\setlength{\tabcolsep}{5pt}
\renewcommand{\arraystretch}{1.15}
\begin{tabular}{lcc}
\toprule
\multirow{2}{*}{Metric} &
\multicolumn{2}{c}{\textbf{MyoFullBody}} \\
\cmidrule(lr){2-3}
& \textbf{Mocap-Body} & \textbf{GMR-Fit} \\
\midrule
Success rate (\%) $\uparrow$ & \bm{$79.94 \pm 3.85$} & $77.67 \pm 5.13 $ \\
Frame Coverage (\%)$\uparrow$ & $90.15 \pm 2.37$ & \bm{$90.67 \pm 2.92 $}\\
Joint angle error (deg) $\downarrow$ & $10.67 \pm 0.11$ & { \bm{$7.97 \pm 0.05$}}\\
Joint velocity error (deg/s) $\downarrow$ & $40.95 \pm 0.18$ & {\bm{$36.51 \pm 0.28$}}\\
Mean episode length $\uparrow$ & $508.68 \pm 13.39$ & \bm{$512.00 \pm 16.52$}\\
Mean episode return $\uparrow$ & $507.11 \pm 13.58$ & \bm{$522.30 \pm 16.82$}\\
\bottomrule
\end{tabular}
\caption{Validation metrics comparing MyoFullBody environments using MoCap-Body vs GMR-Fit retargeting method, with $N = 3$ using KINESIS testing (108 motions) at 2 billion timesteps. Using Mocap-Body retargeting increases significantly the joint angle and joint velocity error.}
\label{tab:validation-metrics-mocapbody}
\end{table}

\subsection{Biomechanical Validation} \label{sec: population-base}
The performance of a model should be validated against independent datasets beyond the training dataset (in our case, KINESIS~\cite{simos2025kinesis} training dataset). To verify the biomechanical fidelity of our model during dynamic motion, we conduct two population-based evaluations on the two most common human gaits, walking and running, against human experiment data on joint angles and moments, GRF, and EMG correlations, as suggested in \cite{Rajagoapal2016,Hicks2015}.

\subsubsection{Kinematics and Kinetics Analysis}

\begin{figure}[h!]
    \centering
    
    \begin{subfigure}[t]{0.96\linewidth}
        \centering
        \includegraphics[width=\linewidth]{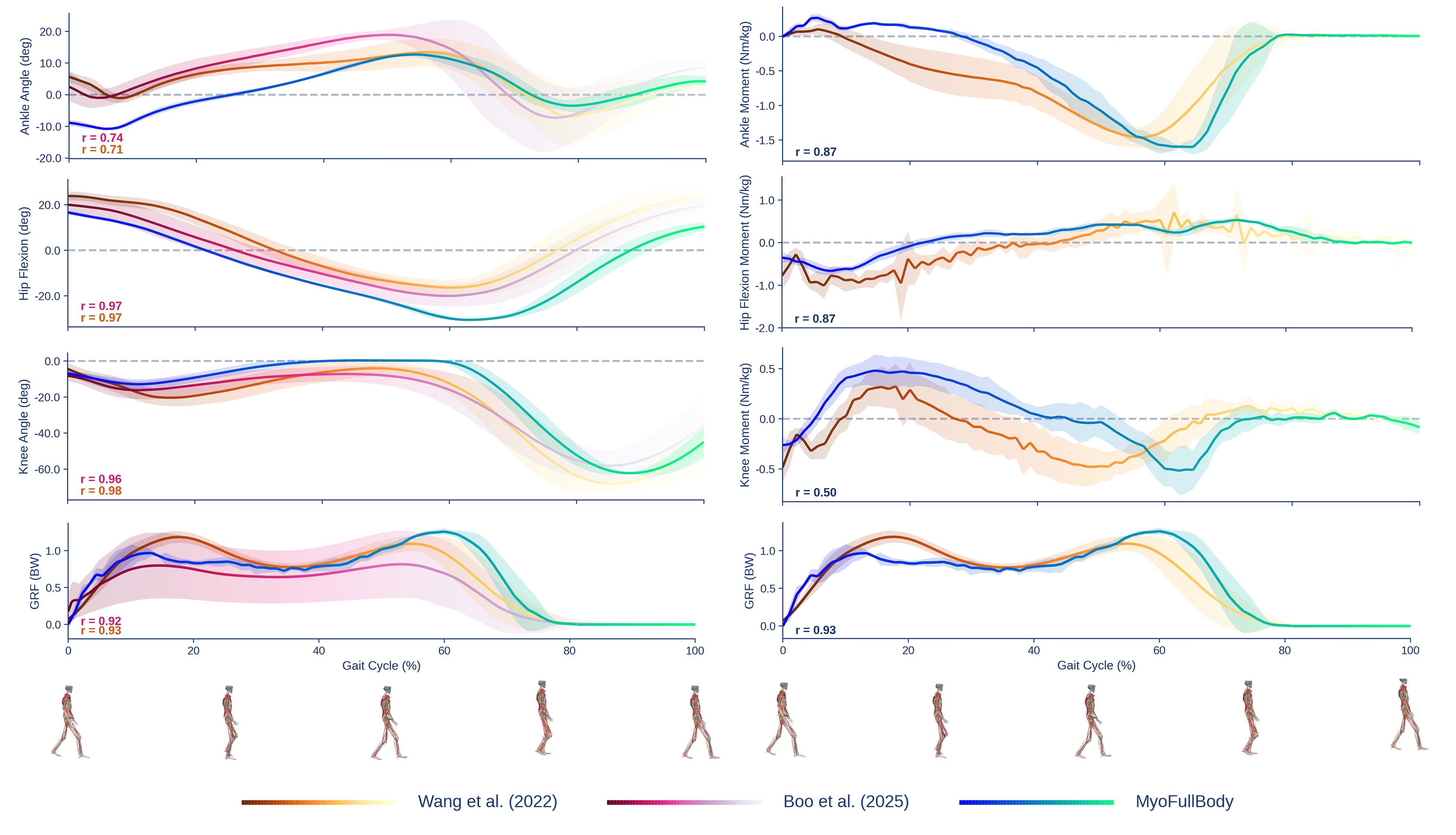}
        \caption{Walking at $1.2\,\mathrm{m/s}$. Human treadmill data (orange, \citep{Huawei2022dataset}) and level-ground walking at mean velocity $1.2\,\mathrm{m/s}$ (purple, \citep{koo2025dataset}).}
        \label{fig:gait_analysis_walk}
    \end{subfigure}
    \hfill
    \begin{subfigure}[t]{0.96\linewidth}
        \centering
        \includegraphics[width=\linewidth]{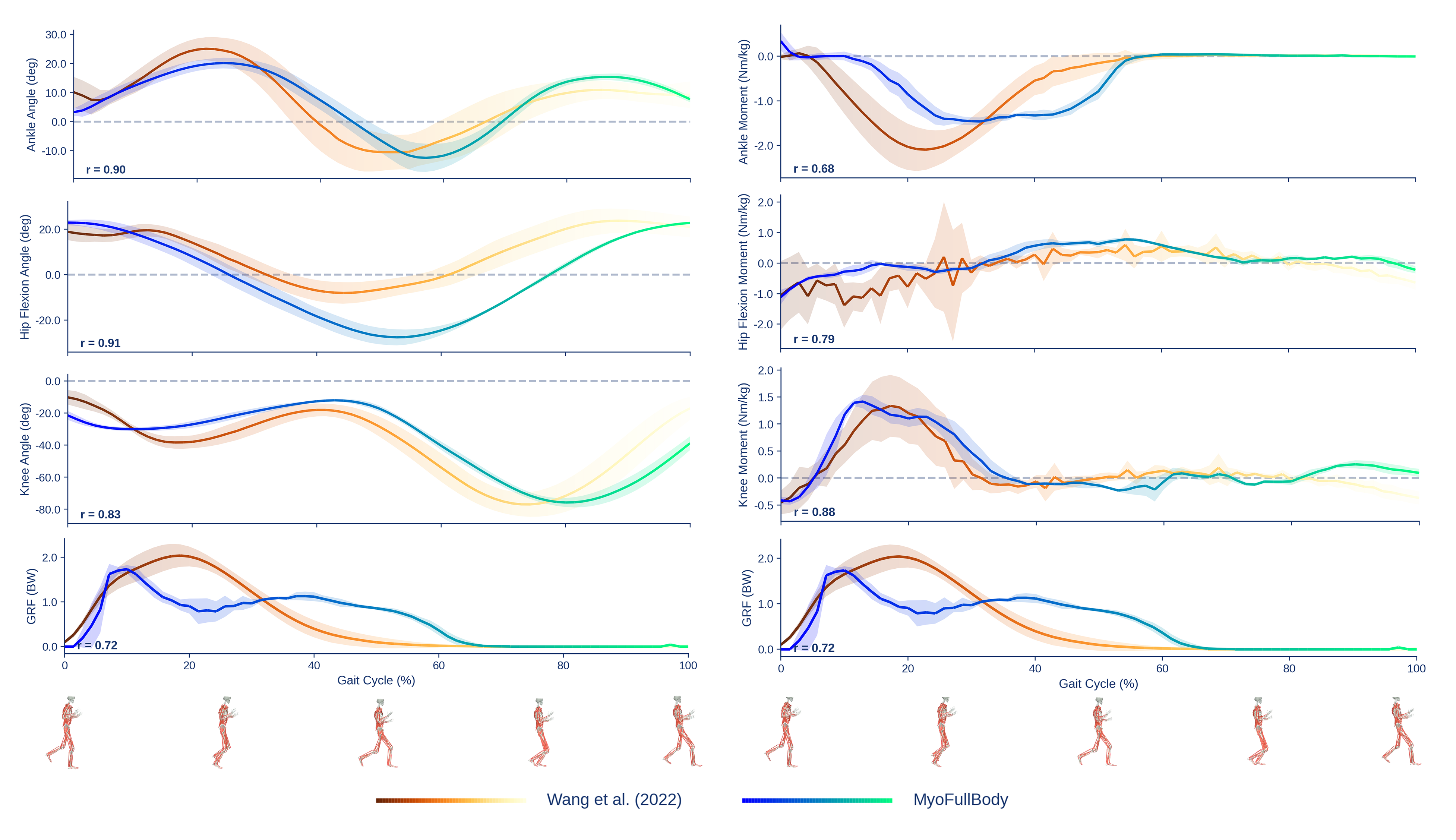}
        \caption{Running at $1.8\,\mathrm{m/s}$. Human treadmill data (orange, \citep{Huawei2022dataset}). No running data are available from~\cite{koo2025dataset}.}
        \label{fig:gait_analysis_run}
    \end{subfigure}

    \caption{Representative left lower-limb joint kinematics (hip, knee, ankle, and foot) over a complete gait cycle, comparing experimental human data and MyoFullBody-generated motion. Simulated results were evaluated on five AMASS sequences (KIT/317/walking\_medium01--05\_poses for walking and KIT/317/walking\_run01--05\_poses for running), aligned by GRF onset and truncated to a single gait cycle. Only kinematic data are available for~\cite{koo2025dataset}.}
    
    \label{fig:gait_analysis_combined}
\end{figure}

\paragraph{Walking.}
We evaluate a pre-trained model checkpoint on 10 billion steps using the full KINESIS motion dataset, comprising five walking sequences from the AMASS dataset, each repeated three times. Simulated joint kinematics and dynamics are compared against an experimental treadmill-walking dataset \citep{Huawei2022dataset} as well as an experimental level walking dataset \cite{koo2025dataset}. Both datasets have a mean velocity of $1.2 \mathrm{m/s}$ and are averaged across nine participants. All datasets are temporally aligned using the GRF onset and truncated to a single full gait cycle. The GRF and joint moment are normalized to the body weight of each participant. The simulated data are processed using the same pipeline with an averaged weight of $84.3$ kg. Fig. \ref{fig:gait_analysis_walk} reports results for the most representative joints of human walking: hip flexion, knee flexion, and ankle flexion in both simulation and human experimental data. We found a mean correlation index of $0.9$ for both treadmill and level walking in kinematics, and $0.79$ for joint dynamics with treadmill walking. Our simulated lower-limb joint movements exhibit a highly stereotyped pattern during walking, similar to experiment results and other literature \cite{UchidaDelp2021}. At initial foot contact, the hip is flexed; during the stance phase, it progressively extends, reaching a peak shortly before toe-off, and then flexes again during the swing phase. The knee is near full extension at initial contact and subsequently flexes during early stance to absorb impact, before re-extending as the body is supported. The ankle dorsiflexes as the tibia advances over the foot and then undergoes a rapid plantarflexion near the end of stance, generating propulsion at toe-off. GRF displays a characteristic double-peaked profile. The first peak corresponds to the loading response as the leading limb accepts body weight, while the second peak arises during push-off as the trailing limb generates forward propulsion.

\paragraph{Running.}
We evaluate the performance of the pretrained MyoFullBody checkpoint on five running AMASS motions, with $N= 3$. The simulated joint kinematics and dynamics are compared against an experimental treadmill-running dataset collected at a speed of $1.8$ m/s \citep{Huawei2022dataset}, averaged across nine participants. The simulated data and human experimental data are aligned and processed the same way as the walking trials. We found the hip, knee and ankle flexion during one gait cycle with a mean correlation index of $0.81$ (Fig. \ref{fig:gait_analysis_run}). During running, the hip is flexed at initial contact and extends throughout stance, reaching peak extension near toe-off before rapidly flexing during swing to advance the limb. The knee contacts the ground in slight flexion, flexes further during early stance to absorb impact, and then extends through mid-to-late stance for support and propulsion, followed by pronounced flexion during swing for foot clearance. The ankle transitions from slight plantarflexion at contact to dorsiflexion in early stance, then generates a strong plantarflexion at push-off, providing the primary propulsive impulse. The vertical GRF displays a single prominent peak during early stance, reflecting rapid load acceptance at foot contact, followed by a gradual decline through late stance as the limb transitions to propulsion.

\subsubsection{Muscle activation analysis}

We next assess the physiological plausibility of the generated muscle activation patterns over gait cycles. To this end, we compare the synthetic activations produced by the policy with EMG recordings collected during walking in two datasets \citep{Huawei2022dataset, koo2025dataset}, both of which provide signals for a subset of right-leg muscles.

Importantly, due to the intrinsic redundancy of the MSK system, multiple muscle coordination strategies can generate similar joint kinematics. As a result, achieving high alignment with human EMG across all muscles is inherently challenging especially for relatively simple tasks such as level walking, since the controller may discover alternative feasible strategies (for instance, by keeping certain muscles minimally active) while still accurately reproducing the motion.

\begin{figure}
    \centering
    \includegraphics[width=0.92\linewidth]{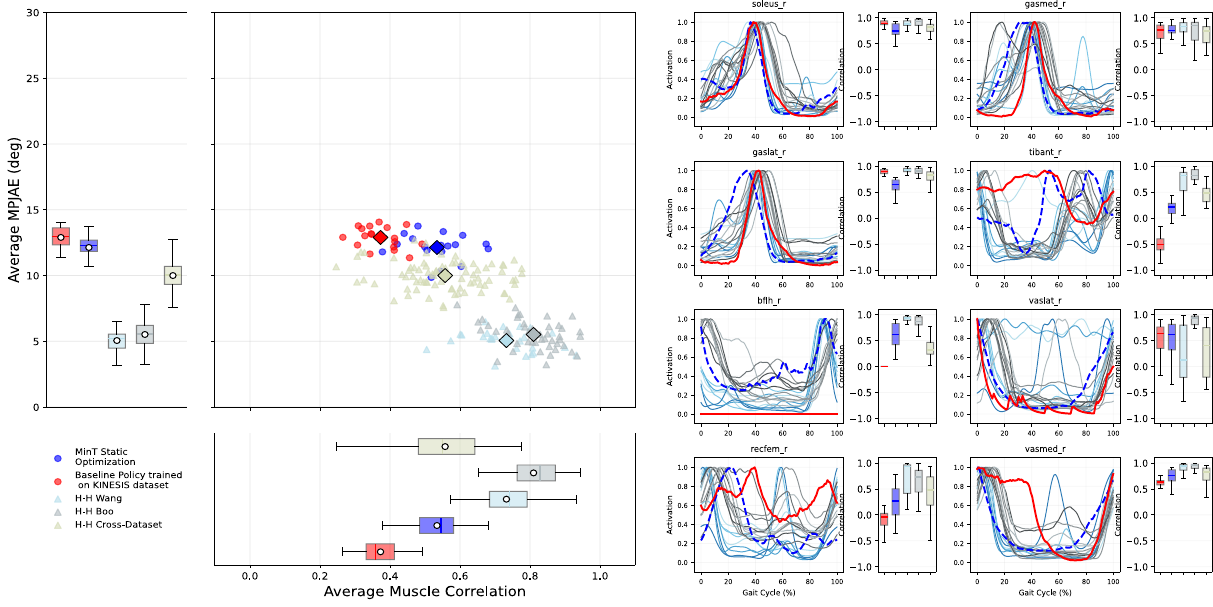}
    \caption{Physiological plausibility of synthetic muscle activations during walking. Comparison between gait-cycle-averaged synthetic muscle activations generated by the policy and experimental EMG recordings during level walking. Left: Average muscle–EMG correlation across all recorded muscles and corresponding lower-limb MPJAE. Each triangle represents a human-to-human pair; each dot represents a model-to-human pair. Right: Average activation patterns and correlation values for single muscles.}
    \label{fig:gait_Boo_Wang}
\end{figure}

To evaluate synthetic muscle activation patterns, we trained a single policy on 972 locomotion trajectories from the KINESIS dataset. We then evaluated the trained policy on a subset of the training trajectories and recorded the resulting synthetic muscle activation signals. These activations were segmented into gait cycles (see Methods) and averaged across cycles to obtain representative activation profiles. We then computed the correlation values between synthetic muscle activation patterns and human EMG signals for eight right-leg muscles recorded in both datasets \citep{Huawei2022dataset, koo2025dataset}. As a baseline for comparison, we also considered average gait muscle activation patterns computed through inverse dynamics.
Static optimization results were not recomputed in this work; instead, they were loaded from the Muscles in Time (MinT) dataset \citep{schneider2024muscles}, where static optimization had previously been performed on KIT walking motions.
Finally, we computed inter-subject variability within each dataset and cross-datasets, to provide an approximate upper bound on achievable model–human alignment.
Figure \ref{fig:gait_Boo_Wang} (right) shows the normalized, gait-cycle-averaged activation patterns of the analyzed muscles. For most muscles, the synthetic activations reproduce the main temporal patterns observed in human EMG signals, showing correlation values comparable to those obtained using static optimization; in the best cases, the obtained correlation is comparable to subject-to-subject variability. 
Correlation values are then averaged across all recorded muscles: in Figure \ref{fig:gait_Boo_Wang} (left), average correlation values are shown in relation to the Mean Per Joint Angle Error (MPJAE) computed on lower-limb kinematics.

Overall, these results suggest that strong kinematic imitation does not automatically guarantee physiologically faithful muscle activations. Across independently trained policies, the observed muscle--EMG correlations span 0.2 to 0.6; this range reflects variability across muscles and across training runs with different random seeds and different reward trade-offs between kinematic tracking and energy regularization. Under a similar motion imitation formulation, KINESIS reports correlations of approximately 0 to 0.45 and shows that non-imitation baselines yield substantially weaker EMG alignment than imitation-based controllers~\cite{simos2025kinesis}, while our kinematic tracking error is lower. Repeated experiments reliably produce positive correlations, but the specific magnitude varies substantially across policies and muscles. This variability is a principled consequence of muscle redundancy: many distinct coordination strategies can produce mechanically equivalent joint kinematics, and a controller optimized for kinematic accuracy has no incentive to converge on the particular strategy employed by human subjects.

\section{Discussion}\label{sec:discuss}

\paragraph{Imitation Learning Performance.}

Our imitation learning metrics are consistent with those reported in prior studies involving MSK models. For instance, KINESIS~\cite{simos2025kinesis} reports a global mean per-joint position error of approximately 42 mm across training and testing motions. In comparison, MuscleMimic achieves a global joint error of 128 mm and a relative joint error of 23 mm with respect to the reference trajectory root.  
When evaluated against the KinTwin model \cite{cotton2025kintwin}, MuscleMimic exhibits greater task variance and a lower failure rate, albeit slightly higher errors to exact kinematic trajectories. This difference largely stems from the underlying control complexity of the two approaches. KinTwin incorporates residual forces, acting as a control simplification that accelerates convergence and tightens tracking. MuscleMimic operates fully on muscle activation, managing a significantly more complex, high-dimensional action space over more joints. In the bimanual task, although a direct equivalent for upper-limb benchmarking is currently unavailable, comparing our generalized framework to localized tracking benchmarks \cite{Selder2025} confirms that MuscleMimic achieves superior overall reference adherence given its expanded scope.

Additionally, our framework enables fine-tuning on more challenging motions beyond simple locomotion. For gentle motions with slow supplementary hand movements, such as dancing or waving, training on a single motion requires fewer than 100 million steps; more complex and dynamic motions, such as kicking combined with a $360^\circ$ twist, require around 1 billion steps. For vertical jumping, we achieve good performance by scaling the maximum isometric force of all muscles to $F'_{\mathrm{max}} = 5\,F_{\mathrm{max}}$, compensating for the lack of elasticity and potential energy storage caused by MuJoCo's non-compliant tendons ~\cite{Bobbert2001, davis2018body}. We further tighten the root termination condition to encourage lift-off. These results demonstrate that MuscleMimic is capable of reproducing dynamic motions.

\paragraph{Retargeted Motions and Dataset.} Our current retargeted dataset is generated via AMASS, which uses an underlying SMPL model \cite{Loper2023}, which has a kinematic chain that does not match biomechanical models \cite{keller2023skel}. Recent work, such as SKEL \cite{keller2023skel}, has improved anatomical accuracy in body models by regressing biomechanical joint locations and bone orientations from skin meshes, but these models remain kinematic representations without physics-based or joint constraint enforcement. Our proposed GMR-Fit retargeting methods with MuscleMimic provide a new dataset of kinematically accurate data that respects the joints and muscle configuration of the validated MyoFullBody and MyoBimanualArm model. Together with ground offset and penetration correction, our dataset provides an accurate foundation for large-scale motion training.

Nevertheless, limitations exist within the current dataset retargeting methods, particularly when extending this approach to pathological gaits in the future. Currently, motion sequences from the AMASS dataset are retargeted to the MyoFullBody morphology by aligning joint positions in a canonical T-pose. However, SMPL joints are statistical predictions based on where joints tend to be relative to the skin surface across a training dataset of healthy, average-proportioned bodies \cite{Loper2023}. While our kinematic and kinetic analyses demonstrate accurate predictions during walking and running, this statistical definition may not faithfully represent individuals with atypical anthropometrics, asymmetric gait patterns, or MSK pathologies. When such motions are retargeted, discrepancies in joint centers, segment lengths, and moment arms can propagate through to simulated dynamics, potentially smoothing over or entirely losing the very clinical features of interest. It remains an open question whether large-scale training in a physics-based dynamic simulator could mitigate these shortcomings.

\paragraph{Limitations.} While our framework demonstrates promising alignment with experimental kinematic data, MSK models remain approximations of biological reality. The Hill-type muscle model simplifies tendon elasticity and fiber recruitment. Muscle redundancy remains a fundamental challenge: kinematic accuracy alone does not ensure physiologically faithful activations, as many distinct coordination strategies can produce mechanically equivalent joint kinematics. SMPL-based retargeting assumes a generic morphology, potentially introducing systematic biases for atypical anthropometrics. Our biomechanical validation currently focuses on walking and running; extending to the full range of demonstrated motions (dancing, jumping, kick-twists) requires corresponding experimental datasets.

\section{Conclusion}

We introduced MuscleMimic, an open-source framework that enables scalable motion imitation learning with physiologically realistic, MSK models. By leveraging GPU-accelerated simulation with massive parallelism, we achieve order-of-magnitude improvements in training speed while maintaining comprehensive collision handling. MuscleMimic pipeline enables a 416-dimensional muscle-driven model to achieve standard locomotion using a generalist within two days of training, and can be finetuned on harder, more dynamic motions. We are able to transform MSK model validation from static, task-specific analyses to systematic stress-testing across diverse dynamic movements. Additionally, our framework provides a strong foundation for further scaling up the MSK motion generations.

By open-sourcing this framework, we hope to enable the broader research community to iterate on these models—refining muscle parameters, improving joint definitions, and validating against diverse experimental datasets. Simulation outputs should be interpreted as model predictions rather than ground truth, and conclusions drawn from simulated muscle activations or joint loads warrant validation against independent experimental measurements before clinical application. We hope that this framework can serve as a groundwork for future studies in the area of rehabilitation, assistive device integration, muscle recruitment patterns, and so on.

\section*{Acknowledgments}

We thank members of the Mathis Group for Computational Neuroscience and AI and NeuRoC Lab for feedback on the project. We also thank Vittorio Caggiano, James Heald, and Balint K. Hodossy for helpful discussions. The core research was completed prior to Cheryl Wang's NVIDIA internship, and the paper was finalized during her internship at NVIDIA and studies at McGill University. Work by A.M.'s team was funded by the Swiss National Science Foundation (SNSF) (310030\_212516), the Simons foundation (SFI-AN-NC-SCN-00007276-14), and Boehringer
Ingelheim Fonds PhD stipends. This work was supported by the Natural Sciences and Engineering Research Council of Canada (NSERC) and the New Frontiers in Research Fund (C.W., G.D., J.K.).

\newpage

\section{Methods}

\subsection{Musculoskeletal Models}

MuscleMimic provides two MSK embodiments, both built and validated upon established MyoSuite components~\citep{caggiano2022myosuite}, incorporating MyoArm, MyoLegs, and MyoBack models~\citep{wang2022myosim,Walia2025} that were tested in previous work, such as ~\citep{chiappa2024acquiring,arnold,wang2025myochallenge}. The muscle actuators are built as Hill-type~\citep{hill1938heat} muscle actuators following MuJoCo~\citep{todorov2012mujoco}, but with inelastic tendons and without pennation angle. Control signals are passed through a first-order nonlinear activation dynamics model to obtain muscle activations, as described by:
\begin{center}
\begin{equation}\label{eq:mus}
\frac{\partial}{\partial t}\mathrm{act} = \frac{\mathrm{ctrl}-\mathrm{act}}{\tau(\mathrm{ctrl},\mathrm{act})} \,, \quad \quad  \tau (\mathrm{ctrl},\mathrm{act}) = \begin{cases}
\tau_{\mathrm{act}} \cdot (0.5 + 1.5  \mathrm{act}), & \mathrm{ctrl} - \mathrm{act} > 0 \\
\tau_{\mathrm{deact}} / (0.5 + 1.5  \mathrm{act}), & \mathrm{ctrl} - \mathrm{act} \le 0
\end{cases} \,.
\end{equation}
\end{center}
Here, $\mathrm{ctrl}$ denotes the normalized neural excitation signal, while $\mathrm{act}$ represents the resulting muscle activation state after accounting for activation and deactivation dynamics. We interpret $\mathrm{act}$ as a proxy for the EMG envelope of the simulated musculature that captures the temporal smoothing and delay between neural excitation and muscle force generation. The effective time constant $\tau(\mathrm{ctrl},\mathrm{act})$ is state-dependent, differing between activation and deactivation phases, with activation and deactivation time constants set to $\tau_{\text{act}} = 0.01\mathrm{s}$ and $\tau_{\text{deact}} = 0.04\mathrm{s}$, respectively, following \cite{Millard2013}. Additionally, we introduce a set of tunable parameters to allow fine-tuned adjustment of the MSK model for highly dynamic motions that require rapid energy generation over short time scales (e.g., vertical jumping). Specifically, we allow the muscle activation time constant $\tau_{\mathrm{act}}$ and the maximum active force $F_{\mathrm{max}}$ of each muscle to be independently adjusted for the upper and lower limbs. The activation time constant $\tau_{\mathrm{act}}$ defines the temporal response of muscle activation to the control input. We observe that smaller values of $\tau_{\mathrm{act}}$ (e.g., $0.001$) lead to faster activation dynamics but are associated with stiffer, less compliant motions, whereas larger values (e.g., $0.05$) yield smoother control and improved performance in highly impulsive motions, such as vertical jumping. The observed results with larger values of $\tau_{\mathrm{act}}$ may be due to the smoothing of the effective action-to-state dynamics, thereby reducing high-frequency sensitivity in the control signal and improving optimization stability. However, such values are not necessarily biologically realistic, as empirical studies suggest upper bounds on muscle activation time constants of approximately 15 ms \cite{Thelen2003, Millard2013}. %

The contact geometries of the entire body are composed of various geometric objects of either a capsule or an ellipsoid, encapsulating the bone segments and muscles of that region. All contact geometries follow point contact detection, with a contact solver that can handle 3D Coulomb friction \cite{mujoco2023docs}. Specifically for the foot-ground contact, each foot has four geometry objects, with each allowing a maximum of two contact points with a plane (e.g., floor). Both models were fine-tuned to enforce symmetry in joint equality constraints, joint ranges, muscle moment arms, and muscle force–length (FL) curves. In addition, irregular jumps in muscle behavior were identified and corrected during model construction. The total mass of the MyoFullBody follows that of a fully-grown human of $84.3$kg. The two limbs of MyoBimanualArm weigh in total $9.3$kg.

\paragraph{Validation.} Validation of the two created models was an iterative process throughout our study. The MSK geometry, body inertia, and joint definition of the original MyoSuite models were converted via MyoConverter \cite{ikkala2022converting} from previous OpenSim models, including \cite{Christophy2011,Saul2014, Lee2015, Rajagoapal2016}, that are anatomically based models of MSK geometry that represent physiological joint kinematics and muscle path geometry. The muscle geometry and moment arms generally vary between individual measurements \citep{HerzogRead1993,Murray2002}. 
We cross-validated the resulting model against several experimental data obtained from MRI or cadaver studies to ensure the trend and magnitude are within 2 standard deviations (SD); a subset of the validation plots is shown in Appendix~\ref{app:muscle_valid}. Finally, instead of scaling the dimensions and inertial properties of a generic model to an individual subject that we measured, the individual data were fitted to MyoFullBody using the SMPL model, see Sec. \ref{sec:motion_retarget}. Additionally, we not only perform kinematics and EMG analysis on classic motions such as walking and running in Sec. \ref{sec: population-base}, but also stress-test on other highly dynamic motions such as vertical jumping. These motions place substantially greater demands on coordination, force generation, and tendon–muscle dynamics, and therefore serve as a stringent stress test of the underlying neuromusculoskeletal model.

\subsection{Motion Dataset} \label{sec:motion_dataset}
The choice of motion dataset is critical for learning generalizable control policies in neuromusculoskeletal systems, as the diversity and realism of reference motions directly define the space of behaviors a policy can represent. For MyoFullBody training, we use 972 motion trajectories from the \texttt{KINESIS\_TRAIN} dataset, a curated subset of the Archive of Motion Capture as Surface Shapes (AMASS; \cite{mahmood2019amass}) and the KIT Motion-Language Dataset \cite{Mandery2016b}, originally introduced in the KINESIS imitation learning framework~\cite{simos2025kinesis}. This dataset contains high-quality full-body motions with standardized movement patterns, including walking, turning, and running. The corresponding \texttt{KINESIS\_TEST} set comprises 108 held-out motions from KIT for evaluation. For the MyoBimanualArm model, we select $1770$ motions from AMASS spanning multiple sources, including ACCAD \cite{AMASS_ACCAD}, BioMotionLab \cite{AMASS_BMLrub}, GRAB \cite{AMASS_GRAB}, KIT, and Transitions\_mocap \cite{mahmood2019amass} as \texttt{BIMANUAL\_TRAIN}. Motions are filtered using keywords such as “arm,” “hand,” “throw,” “tennis,” “wipe,” “pour,” “drink,” “punch,” “pass,” and “pick,” emphasizing upper-limb–dominant movements. The corresponding \texttt{BIMANUAL\_TEST} set comprises $312$ held-out motions for evaluation.

\subsection{Motion Retargeting}\label{sec:motion_retarget}

\begin{figure}[h]
    \centering
    \includegraphics[width=0.7\linewidth]{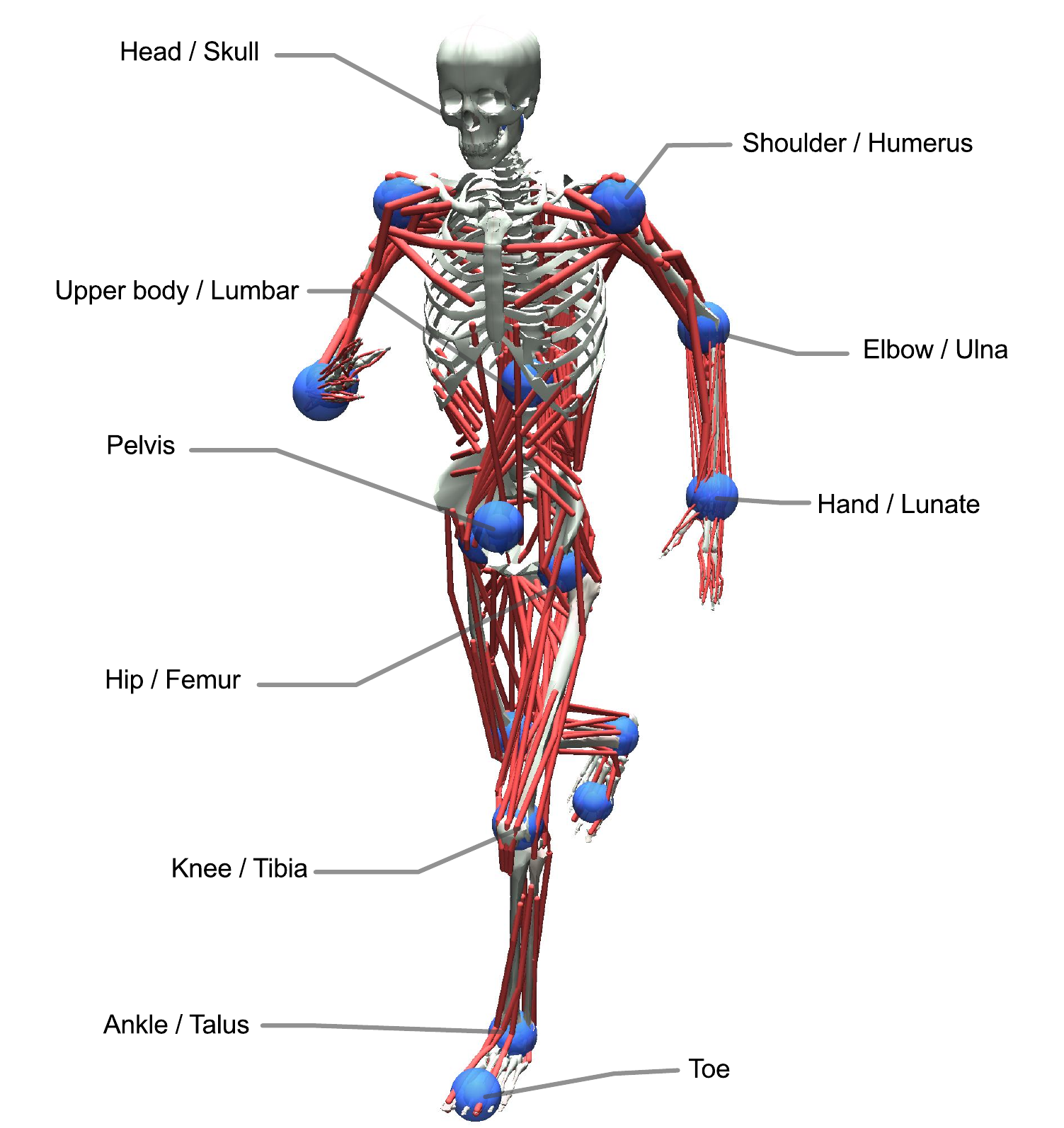}
    \caption{The 17 mimic sites used for full-body motion imitation, shown with their corresponding anatomical keypoints. Blue balls indicate mimic site locations on the body segments. For MyoBimanualArm, a subset of 6 upper-limb sites plus 1 thorax site is used, with the thorax reference site for computing relative positions.}
    \label{fig:mimic_sites}
\end{figure}

\paragraph{Mimic sites.} We define a set of mimic sites on the MSK model that serve as reference points for motion retargeting and imitation learning. For MyoFullBody, 17 sites are distributed across the full body at key anatomical landmarks: head, shoulders, elbows, hands, lumbar spine, pelvis, hips, knees, ankles, and toes (Fig.~\ref{fig:mimic_sites}). For MyoBimanualArm, a subset of 6 upper-limb sites captures the essential kinematics for bimanual manipulation tasks. These sites define the task-space objectives in our reward formulation and provide the target trajectories for motion imitation. We construct our motion retargeting pipeline using two complementary approaches: simulation-driven retargeting, implemented via a mocap-object in MuJoCo and referred to as Mocap-Body retargeting, and kinematics-based retargeting, improved based on General Motion Retargeting (GMR)~\cite{joao2025gmr,ze2025gmr} and referred to as GMR-Fit retargeting. We support two different sets of data for retargeting: the AMASS dataset \cite{mahmood2019amass} for training, and the clinical-based mocap dataset for validation. 

\begin{figure}[h]
    \centering
    \includegraphics[width=\linewidth]{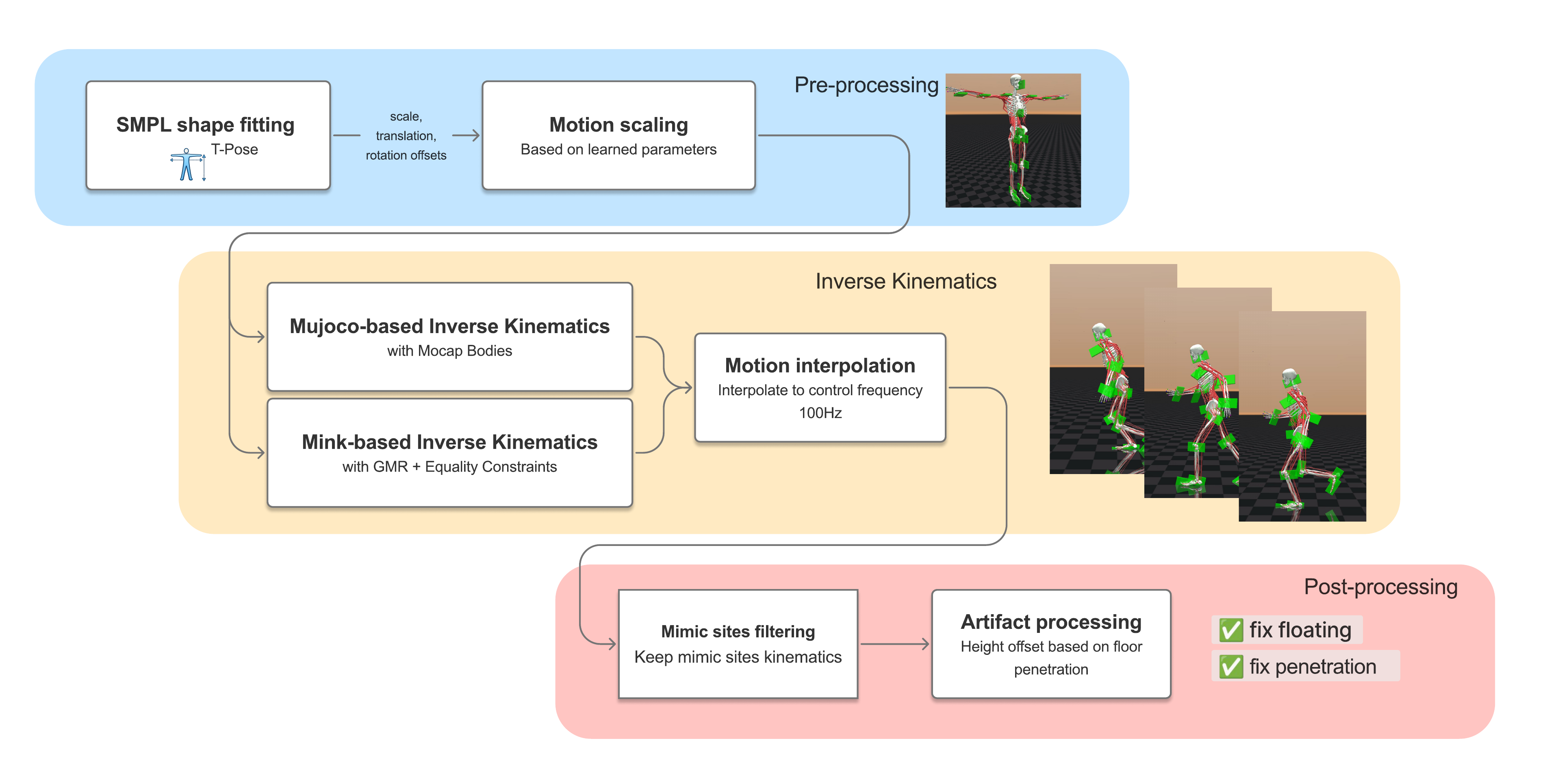}
    \caption{Overview of the motion retargeting pipeline with either Mimic or GMR. Human shape is first fitted using SMPL and globally scaled to the target model. Motion is then retargeted via inverse kinematics (MuJoCo-based with Mimic or Mink-based with GMR and equality constraints) and interpolated to the control frequency. Post-processing filters mimic sites and corrects artifacts such as ground penetration and floating.}
    \label{fig:retargeting-pipeline}
\end{figure}

\paragraph{Mocap-Body Retargeting.}
The Mocap-Body retargeting pipeline uses mocap body, a kinematic, massless body in MuJoCo whose pose is directly prescribed and governed by physical dynamics. Mocap bodies are attached at the mimic sites (Fig. ~\ref{fig:mimic_sites}), in which MuJoCo performs internal inverse kinematics to determine the joint configurations. The motion retargeting pipeline consists of three stages, as detailed in Fig. \ref{fig:retargeting-pipeline}: pre-processing, inverse kinematics, and post-processing. First, AMASS motion sequences \cite{mahmood2019amass}, which provide pose and shape parameters for the Skinned Multi-Person Linear model (SMPL, \cite{Loper2023}), undergo shape fitting in a T-pose. This estimates SMPL-H shape coefficients $\boldsymbol{\beta}$ (body proportions), a global scale $s$, and joint-wise positional and rotational offsets $(\Delta \mathbf{p}, \Delta \mathbf{R})$, which are then used to scale motions to the MyoFullBody morphology following loco-mujoco \cite{al2023locomujoco}. The scaled motions are then retargeted via inverse kinematics and temporally interpolated to the control frequency. Finally, the retargeted motions are temporally interpolated to a 100 Hz control frequency and post-processed to filter mimic sites and corrects artifacts such as floating and ground penetration. The MyoBimanualArm model follows the same retargeting procedure, but extracts only the relevant joints for upper-body tasks.

\paragraph{GMR-Fit Retargeting.}

GMR is a robotics retargeting framework aimed at providing physiologically realistic trajectories to avoid common retargeting artifacts such as foot sliding, self-penetration, frame jumping and physiologically infeasible motion. Compared to the Mocap-Body retargeting method, trajectories from GMR follow model-defined joint constraints and reduce sudden posture jumps in between frames. The original GMR uses a manually defined JSON for joint transformation and marker scaling. In our current approach, we adopt the SMPL-fitting from Mocap-Body retargeting to create the \textbf{GMR-Fit} pipelines to allow markers to transform and be fitted within MyoFullBody. Additionally, we incorporate equality constraints and dependency between complex joints (e.g., shoulder and knees). The overall GMR-Fit retargeting pipeline is also shown in Fig.~\ref{fig:retargeting-pipeline}.

\paragraph{Retargeting Evaluation Metrics.}
We evaluate the two retargeting pipelines using a suite of kinematics and model stability metrics: joint limit violations, ground penetration, floating above ground, tendon instability, root mean square error and retargeting efficiency.  Joint limit violations are quantified as the percentage of frames in which the joints exceed their prescribed limits beyond a small numerical tolerance, which we define as $10^{-5}$ rad. Ground penetration is measured in terms of prevalence, defined as the percentage of frames exceeding a penetration depth of 1 mm, and severity, which is the maximum penetration depth observed across the entire motion. Penetration is calculated via MuJoCo's contact distance with the floor geometry. Floating above ground is characterized by the maximum vertical separation between the ground plane and the lowest non-floor geometry across all frames of motion. We define a tendon jump as an abrupt, frame-to-frame change in tendon length that is anomalously large relative to the tendon's recent dynamic behavior. Let $L_k(t)$ denote the length of tendon $k$ at time step $t$, and let $L_{0,k}$ be its rest length. 
We compute the per-step \emph{relative} tendon length change as
\begin{equation}
    \Delta L_k^{\mathrm{rel}}(t)
    =
    \frac{\lvert L_k(t) - L_k(t-1) \rvert}{\max(L_{0,k}, \varepsilon)} \,,
\end{equation}
where $\varepsilon$ is a small constant to ensure numerical stability. To capture the typical smooth variation of each tendon, we maintain an exponential moving average (EMA \cite{EMA}) of the relative change:
\begin{equation}
    \overline{\Delta L_k^{\mathrm{rel}}}(t)
    =
    (1 - \alpha)\,\overline{\Delta L_k^{\mathrm{rel}}}(t-1)
    +
    \alpha\,\Delta L_k^{\mathrm{rel}}(t) \,,
\end{equation}
where $\overline{\Delta L_k^{\mathrm{rel}}}(t)$ denotes the exponential moving average of the relative tendon length change and $\alpha = 0.01$ is the smoothing coefficient. A tendon jump is detected at time $t$ if
\begin{equation}
    \Delta L_k^{\mathrm{rel}}(t)
    >
    \max\!\left(
        \gamma\,\overline{\Delta L_k^{\mathrm{rel}}}(t),
        \Delta L_{\min}^{\mathrm{rel}}
    \right),
\end{equation}
where $\gamma = 10$ is a relative amplification factor and
$\Delta L_{\min}^{\mathrm{rel}} = 10^{-3}$ enforces a minimum relative change to suppress numerical noise and near-static fluctuations. The final reported value is the maximum relative tendon length change classified as a jump across the trajectory. The RMSE is computed over the entire trajectory based on the mean positional error between the reference motion and the retargeted motion. And finally, retargeting efficiency is reported as the average retargeting time per frame. Both retargeting frameworks are evaluated on the various datasets from AMASS \cite{mahmood2019amass}. 

\begin{table}[htbp]
\centering
\caption{\small Performance comparison between Mocap-Body retargeting and GMR-Fit retargeting over 2021 motions, including walking, turning, running, jumping, kicking, crawling, fighting, etc. We report eight distinct metrics averaged over all motions: (1) Joint limit violation rate: $P_\text{joint}$, averaged across all motions (2) Ground penetration rate: $P_\text{pen}$ (3) Maximum ground penetration: $D^\text{max}_\text{pen}$ (4) Maximum floating height: $H^\text{max}_\text{float}$ (5) Maximum tendon jump: $\Delta L^\text{max}_\text{tj}$ (6) Tendon jump rate: $P_\text{tj}$, out of all motions (7) Root Mean Square Error: RMSE (8) Retargeting speed: $T_\text{frame}$. For all metrics, lower indicates better retargeting performance.}
\label{tab:retarget-results}

\begin{tabular}{@{}lcccccccc@{}}
\toprule
\textbf{Method} &
$P_\text{joint}$ (\%) &
$P_\text{pen}$ (\%) &
$D^\text{max}_\text{pen}$ (m) &
$H^\text{max}_\text{float}$ (m) &
$\Delta L^\text{max}_\text{tj}$ (--) &
$P_\text{tj}$ (\%) &
RMSE (m) &
$T_\text{frame}$ (s) \\
\midrule

\multicolumn{9}{@{}l}{\small\textbf{KINESIS (972 motions)}} \\
\midrule
Mocap-Body retargeting & 12.26 & 0.55 & 0.002 & 0.023 & 0.04 & 30.14 & 0.039 & \textbf{0.076} \\
GMR-Fit retargeting   & \textbf{0.27}  & \textbf{0.24} & \textbf{0.001} & \textbf{0.021} & \textbf{0.01} & \textbf{3.20}  & \textbf{0.025} & 0.251 \\

\midrule
\addlinespace[0.6em]

\multicolumn{9}{@{}l}{\small\textbf{Transition (110 motions)}} \\
\midrule
Mocap-Body retargeting & 5.67 & 1.83 & \textbf{0.004} & \textbf{0.012} & 0.08 & 62.73 & 0.030 & \textbf{0.056} \\
GMR-Fit retargeting   & \textbf{0.40}  & \textbf{1.64 }& \textbf{0.004 }& 0.017 & \textbf{0.07} & \textbf{60.00}  & \textbf{0.027} & 0.142 \\

\midrule
\addlinespace[0.6em]

\multicolumn{9}{@{}l}{\small\textbf{ACCAD (252 motions)}} \\
\midrule
Mocap-Body retargeting & 20.23 & 2.81 & \textbf{0.007} & \textbf{0.008} & 0.50 & 78.97 & \textbf{0.050} & \textbf{0.101} \\
GMR-Fit retargeting   & \textbf{0.66}  & \textbf{2.09} & \textbf{0.007} & \textbf{0.008} & \textbf{0.10} & \textbf{52.78}  & 0.058 & 0.302 \\

\midrule
\addlinespace[0.6em]

\multicolumn{9}{@{}l}{\small\textbf{Selected BioMotionLab\_NTroje (687 motions)}} \\
\midrule
Mocap-Body retargeting & 11.04 & \textbf{1.75} & \textbf{0.003} & 0.024 & 0.09 & 53.86 & 0.036 & \textbf{0.094} \\
GMR-Fit retargeting   & \textbf{0.76}  & 2.00 & \textbf{0.003} & \textbf{0.020} & \textbf{0.03} & \textbf{16.01} & \textbf{0.026} & 0.334 \\

\bottomrule
\end{tabular}
\end{table}

\paragraph{Retargeting Results.} We report the retargeting results for Mocap-Body and GMR-Fit in Table \ref{tab:retarget-results}. Overall, GMR-Fit based retargeting achieves higher joint-limit satisfaction than Mocap-Body. This is because the MuJoCo mocap body does not inherently enforce joint constraints. Both methods apply explicit ground-penetration and floating offsets, and are comparable in metric behavior. With respect to tendon jumping, GMR exhibits both lower maximum jump magnitudes and a smaller fraction of affected motions. We observed that motions with large $\Delta L^\text{max}_\text{tj}$ don't necessarily always reflect a tendon jump due to a sudden shift of joint configurations or joint violations, but could also be due to extremely rapid motion (e.g., pushing of the elbow) that stretches or shortens the muscle. Motions in the ACCAD dataset contain some dynamic motions, such as martial arts kicks and crawling forward, that significantly increase the tendon jump rate. Mocap-Body retains a substantial computational advantage, with retargeting timescales approximately three times faster than GMR. The difference in performance of the two retargeting pipelines was discussed in Sec. \ref{sec: results - imitation learning}.

\subsection{Motion Imitation Training} \label{sec: motion_imitation_training}

\paragraph{Implementation.} We implemented MuscleMimic as a JAX-based framework extending LocoMuJoCo~\cite{al2023locomujoco} with three major additions: customized retargeting pipeline with additional support to GMR and more motion captured datasets, native MuJoCo Warp support for GPU-accelerated simulation with flexible collision support, and extensive redesign for MSK systems. An optional MJX backend is available for reduced contact configurations. The modular design enables rapid experiment configuration and validation for both policy and MSK models.

\paragraph{Early termination.} We employ early termination based on relative position error to prevent the policy from exploring highly unrealistic configurations. At each timestep, we compute the mean Euclidean distance across all mimic sites between each site's current position and its reference position, both expressed relative to the root frame. If this mean deviation exceeds a threshold $\delta_{\text{site}}$, the episode terminates immediately. For MyoFullBody, an additional root deviation threshold $\delta_{\text{root}}$ terminates episodes when the pelvis deviates from the reference in world coordinates beyond the allowed range (see Appendix~\ref{app:train_hparams} for specific values). Crucially, we use relative rather than absolute world-frame position error, as our objective is not exact reproduction of the motion-capture trajectory, but preservation of the reference motion’s form, naturalness, and characteristic dynamics. An absolute strict termination condition would penalize these unavoidable discrepancies, prematurely ending episodes even when the policy produces biomechanically valid movement patterns. The relative formulation tolerates global drift while still enforcing local postural accuracy, ensuring that the learned policy maintains proper limb coordination regardless of accumulated root position error.

\paragraph{Observation space.} Both policies receive observations comprising four components: proprioceptive state, muscle state, goal specification, and the previous action $a_{t-1}$. The proprioceptive state encodes joint positions and velocities for all non-root degrees of freedom. Muscle observations consist of configurable per-actuator quantities: neural excitation $u \in [-1, 1]$ representing the control signal, muscle-tendon unit length $l$, contraction velocity $\dot{l}$ (positive during lengthening), and force $F$ (negative by convention, as muscles generate tension). Goal observations provide target joint configurations with $n$-step look ahead at $0.2$\,s intervals (we use $n=5$ in experiments), together with current and target site positions, orientations, and velocities expressed in the local reference frame. Including the previous action introduces a first-order autoregressive dependency in time, allowing the policy to account for muscle activation dynamics and produce smooth control signals across consecutive steps.

\begin{figure}[h]
    \centering
    \includegraphics[width=0.75\linewidth]{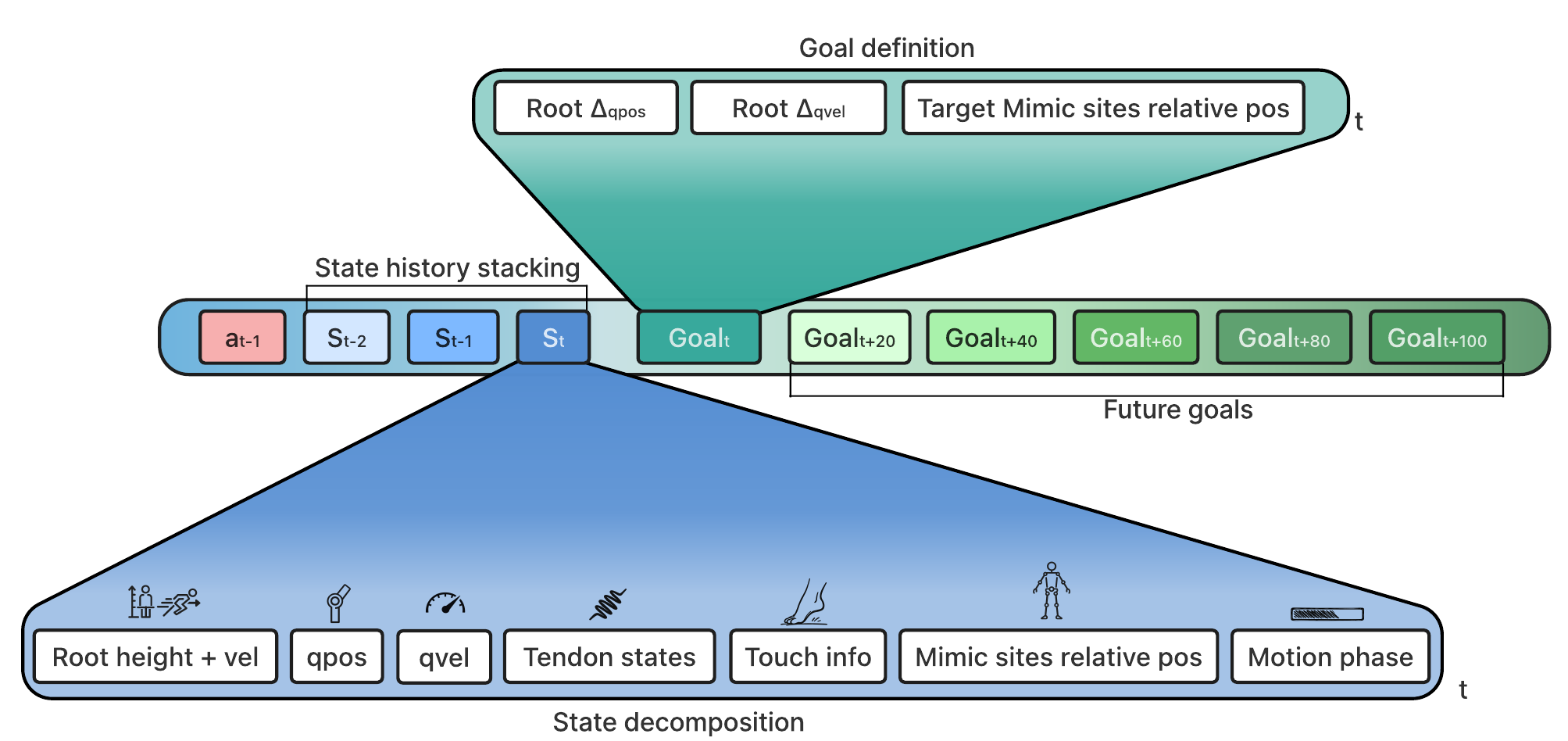}
    \caption{Policy observation structure. The state is decomposed into proprioceptive signals (root height and velocity, joint positions and velocities), tendon states, touch info, mimic site relative positions, and motion phase. A history of stacked states is concatenated with the current goal and future goals at regular look ahead intervals. Each goal is defined by root position and velocity deltas and target mimic site relative positions. The previous action $a_{t-1}$ is appended to introduce a first-order autoregressive dependency.}
    \label{fig:policy_observation}
\end{figure}

The two policies differ in their observation dimensionality to match their respective embodiments. For MyoBimanualArm (fixed base), goal tracking is defined over 6 upper-limb sites. For MyoFullBody, the proprioceptive state additionally includes root height, orientation (quaternion), and 6-DoF root velocity; four touch sensors provide contact normal force magnitudes at the feet and toes; and goal tracking spans 17 sites distributed across the full body. The full observation structure is illustrated in Fig.~\ref{fig:policy_observation}.

\paragraph{Reward formulation.}
We employ a DeepMimic-like reward~\cite{peng2018deepmimic}: $r_t = \max\{0,\; r_t^{\text{imit}} + P_t\}$ and uses imitation term combines joint-space and site-based objectives:
$r_t^{\text{imit}} = w_q r_q + w_{\dot q} r_{\dot q} + w_p r_p + w_\theta r_\theta + w_v r_v^\omega + w_v r_v^v,$ where $w_\cdot$ are mixing weights and site-based quantities are computed relative to a reference site $s_0$ (the pelvis).
The joint position reward separates scalar and quaternion degrees of freedom:
\begin{equation}
r_q = \exp\!\left(-\beta_q \left[\frac{1}{N}\sum_i (q^{\text{lin}}_i - q^{*,\text{lin}}_i)^2 + \bar\theta \right]\right),
\quad
\bar\theta = \frac{1}{N_q}\sum_j \theta(q^{\text{quat}}_j, q^{\text{quat}*}_j),
\end{equation}
where $\beta_q$ is a temperature parameter, $N$ is the number of scalar joint DOFs, $N_q$ is the number of quaternion joints, and $\theta(\cdot,\cdot)$ denotes the geodesic angle between quaternions; $\bar\theta=0$ when no quaternion joints exist.
Task-space rewards measure relative site positions and orientations:
\begin{equation}
r_p = \exp\!\left(-\beta_p \cdot \frac{1}{K-1}\sum_{i=1}^{K-1} \|\mathbf{p}^{\text{rel}}_i - \mathbf{p}^{*,\text{rel}}_i\|^2\right),
\end{equation}
\begin{equation}
r_\theta = \exp\!\left(-\beta_\theta \cdot \frac{1}{K-1}\sum_{i=1}^{K-1} \|\boldsymbol{\phi}^{\text{rel}}_i - \boldsymbol{\phi}^{*,\text{rel}}_i\|^2\right),
\end{equation}
with $\mathbf{p}^{\text{rel}}_i=\mathbf{p}_i-\mathbf{p}_{s_0}$ and $\boldsymbol{\phi}^{\text{rel}}_i=\mathrm{rotvec}(R_{s_0}^\top R_i)$, where $K$ is the number of mimic sites and $\beta_p, \beta_\theta, \beta_v$ are temperature parameters. Site velocities are computed in the reference frame and decomposed into angular and linear components:
\begin{equation}
r_v^\omega = \exp\!\left(-\beta_v \cdot \frac{1}{K-1}\sum_{i=1}^{K-1} \|\boldsymbol{\omega}^{\text{rel}}_i - \boldsymbol{\omega}^{*,\text{rel}}_i\|^2\right),
\quad
r_v^v = \exp\!\left(-\beta_v \cdot \frac{1}{K-1}\sum_{i=1}^{K-1} \|\mathbf{v}^{\text{rel}}_i - \mathbf{v}^{*,\text{rel}}_i\|^2\right).
\end{equation}
The penalty term comprises a clipped sum of regularizers including action bounds violation, action rate (optional), and muscle activation energy:
\begin{equation}
P_t = \max\{-1,\; -\sum_{p\in\mathcal{K}_{\text{pen}}}\lambda_p C_p\},
\end{equation}
where $\mathcal{K}_{\text{pen}}$ is the set of active penalty terms and $\lambda_p$ are the corresponding penalty coefficients.

\begin{figure}[h]
    \centering
    \includegraphics[width=\linewidth]{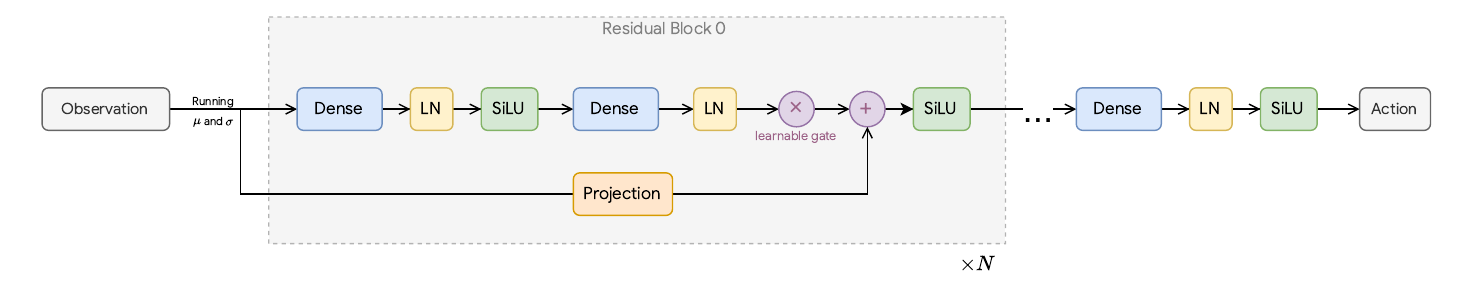}
    \caption{Gated residual policy architecture. Observations are first normalized by running statistics and then passed through $N$ residual blocks. Each block applies two Dense--LayerNorm layers with a learnable gate ($w_l = \sigma(g_l)$) on the residual branch and a projection shortcut when dimensions change. If the number of hidden layers is odd, the final layer is processed as a standalone Dense--LayerNorm--SiLU layer.}
    \label{fig:residual_arch}
\end{figure}

\paragraph{Policy architecture.}
We use an actor-critic architecture with separate policy and value networks. Both are multi-layer perceptrons with SiLU activations~\cite{elfwing2018sigmoid} and optional LayerNorm~\cite{ba2016layer} between hidden layers, initialized with orthogonal weights~\cite{saxe2014exact}. Input observations are normalized using online running statistics computed via Welford's algorithm~\cite{welford1962note}.
The policy $\pi_\phi$ outputs a diagonal Gaussian distribution:
\begin{equation}
\pi_\phi(\mathbf{a} | \mathbf{o}) = \mathcal{N}\big(\boldsymbol{\mu}_\phi(\mathbf{o}), \text{diag}(\boldsymbol{\sigma}^2)\big),
\end{equation}
where $\boldsymbol{\mu}_\phi(\mathbf{o})$ is the state-dependent mean from the actor network, and $\boldsymbol{\sigma}$ is a state-independent learnable standard deviation vector.
Both actor and critic use a gated residual architecture in which consecutive pairs of hidden dimensions form residual blocks. Each block consists of two linear layers with LayerNorm and SiLU activation:
\begin{equation}
\mathbf{x}_{l+1} = \mathrm{act}\!\Big(\mathrm{skip}(\mathbf{x}_l) + w_l \cdot \mathrm{LN}\!\big(W_1^{(l)}\,\mathrm{act}(\mathrm{LN}(W_0^{(l)}\,\mathbf{x}_l))\big)\Big),
\end{equation}
where $\mathrm{skip}(\mathbf{x}_l)$ is a learned linear projection when the input and output dimensions differ, and the identity otherwise (Fig.~\ref{fig:residual_arch}). The residual weight $w_l$ is computed via a gated mechanism: a learnable scalar $g_l$, initialized to $-2.0$, is passed through a sigmoid to yield the learnable gate $w_l = \sigma(g_l) \approx 0.12$ at initialization. Combined with near-zero orthogonal initialization (gain $= 0.01$) of each block's second layer, this ensures that the network behaves approximately as an identity mapping at the start of training. This design stabilizes early optimization and allows the network to gradually incorporate deeper representations as training progresses.

\paragraph{Optimizer.} 
We use the \textit{Muon} optimizer~\cite{jordan2024muon} for 2D weights (Linear layers) and Adam~\cite{DBLP:journals/corr/KingmaB14} for 1D weights (biases and layernorm) and observe that it learns significantly faster and yields higher rewards compared to AdamW when both use a weight decay of $1\times10^{-3}$. This aligns with recent findings that Muon's weight-decay-based update scaling improves convergence and performance~\cite{liu2025muon}.

\subsection{EMG processing}
For EMG comparisons, we leveraged two publicly available datasets of paired EMG–kinematics recordings collected during human walking from Wang et al. \citep{Huawei2022dataset} and Boo et al. \cite{koo2025dataset}. The first dataset includes EMG recordings from nine right-leg muscles during treadmill walking, whereas the second provides EMG from twelve right-leg muscles together with full-body kinematics across a larger set of walking trials. Only muscles common to both the simulated model and the experimental datasets were retained for comparison.

EMG signals were processed according to standard procedures for gait analysis. Raw EMG traces were rectified and normalized on a per-muscle basis to enable inter-subject and inter-condition comparisons. Signals were temporally aligned to the gait cycle and resampled to a normalized stride representation, allowing population-level averaging across trials and subjects. The gait cycles from each of the synthetic motions generated by the policy are extracted following the same procedure described in \cite{simos2025kinesis}. Foot contact sequences were identified from GRFs, and gait cycles were segmented accordingly. Cycles were filtered based on duration and interpolated to a common temporal resolution consistent with the experimental data. Subsequently, for each muscle, we computed mean activation profiles and variability across gait cycles. These summary statistics were used to compare biological recordings with muscle activations generated by the trained locomotion policies.

\newpage
\appendix

\section{Ablation Study}

In Fig. \ref{fig:full_ablation}, Tab. \ref{tab:ablation_results_cpu} and Tab. \ref{tab:ablation_results_gpu}, we present the hyperparameter tuning and ablation results evaluating policy performance at 2 billion training steps. All models are evaluated on the KINESIS testing dataset using a 0.5 m root termination condition, a 0.3 m mean relative site deviation threshold, and no rotational termination threshold (detailed metrics are provided in Appendix \ref{app:validation-metrics}). We evaluate the following key components:

\paragraph{Policy Network Size.} We analyze the impact of model capacity by comparing networks with 10M, 50M (baseline), and 100M parameters, all utilizing residual connections and layer normalization. Increasing the model capacity directly improved policy performance. The 100M parameter model outperformed both the 50M baseline and the 10M model across nearly all tracked metrics, particularly in episode return and frame coverage. However, a 100M model takes around two times longer to train comparing to a 10M model.

\begin{figure}[h]
    \centering
    \begin{subfigure}[b]{0.45\linewidth}
    \centering
    \includegraphics[width=\linewidth]{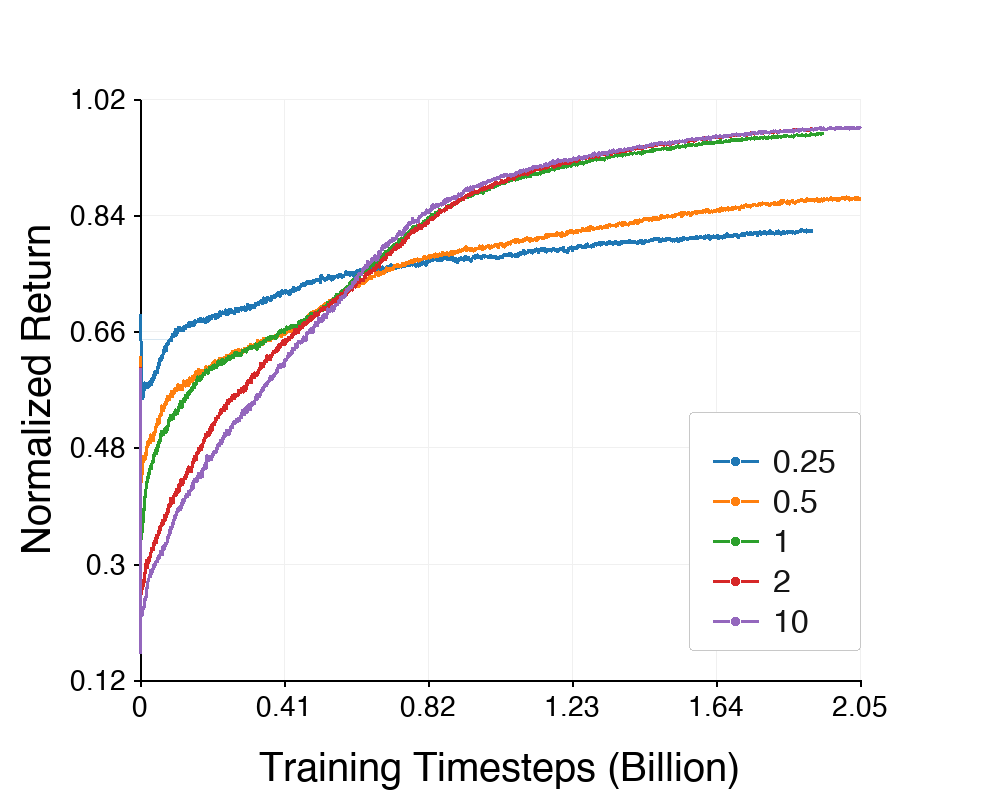}
    \caption{Normalized episode return}
    \label{fig:term_a}
    \end{subfigure}
    \begin{subfigure}[b]{0.45\linewidth}
    \centering
    \includegraphics[width=\linewidth]{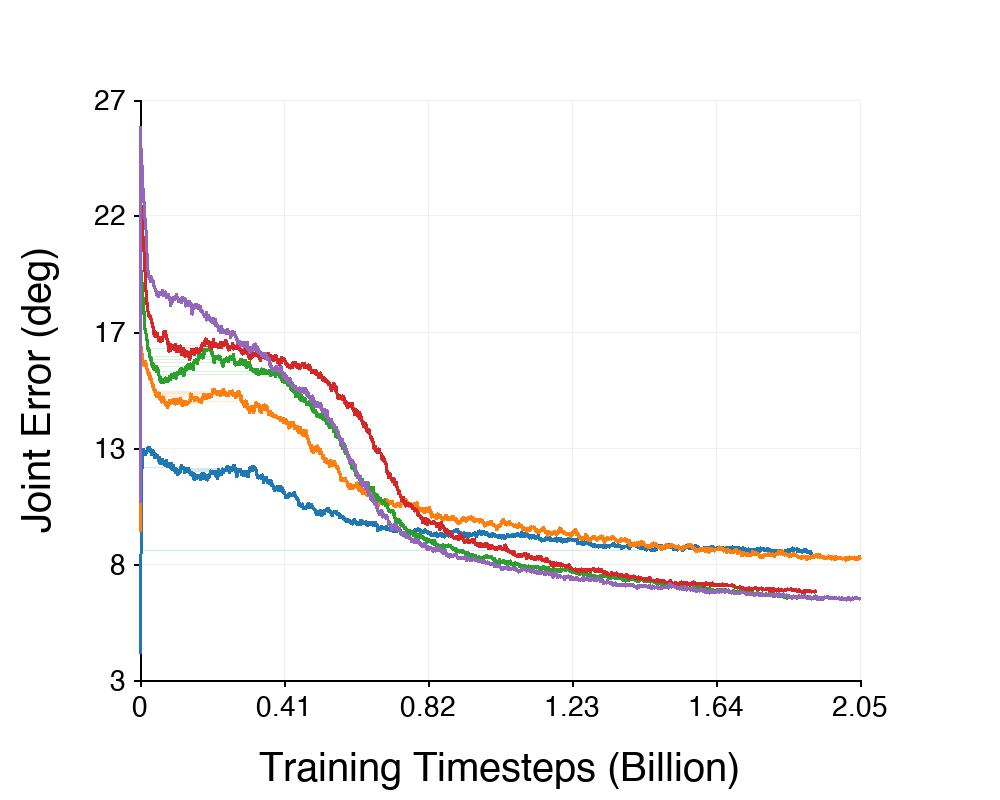}
    \caption{Mean joint angle error}
    \label{fig:term_b}
    \end{subfigure}
    \caption{Impact of termination threshold on learning performance. A threshold larger than 1 initially shows lower returns and shorter episodes but ultimately achieves superior convergence compared to more restrictive thresholds of 0.25 and 0.5.}
    \label{fig:termination_condition}
\end{figure}

\paragraph{Termination Thresholds.} We examine the effect of early termination strictness during training by comparing looser thresholds (1 m and 10 m) and a tighter threshold (0.25 m) against our baseline (0.5 m). A looser termination threshold provides additional exploration opportunities. If trained without the capability of MuscleMimic until around 100 million steps, we will observe that a tighter termination threshold gives a higher and more stable return. However, those looser conditions eventually converge to higher total returns compared to both the $0.25$ m and $0.5$ m thresholds at our evaluated checkpoint timestep of 2 billion (Fig. \ref{fig:termination_condition}). We also observed that a lower threshold of $0.25$ m fails to develop effective fast-turn motions within the constrained timestep. However, we found that for single motion finetuning, it can be helpful to tighten termination conditions, especially for higher dynamic motions, to prevent the policy from exploiting loopholes that skip challenging maneuvers.

\paragraph{Dataset Size.} To understand the role of data volume and diversity, we evaluate policies trained on a 16\% subset of KINESIS~\cite{simos2025kinesis} motions (160 straight walking trajectories), the full KINESIS training set (972 locomotion trajectories), and KINESIS augmented with additional transition motions with AMASS (1022 motions). Notably, this progression not only increases the dataset size but also introduces greater behavioral diversity, moving from simple, homogeneous walking patterns to a broader distribution of locomotion and transition dynamics.
The baseline dataset clearly outperforms the reduced subset (KINESIS-) in both episode return and frame coverage, whereas no significant difference is observed with the augmented dataset (KINESIS+).
Since all models were trained for the same total timesteps, the KINESIS- policy actually received more gradient updates per trajectory but still underperformed, albeit with tighter variance. While for the policy trained on KINESIS+, the similarity in result could suggest that the policy may not have fully converged within the allocated training budget, or the benefits of large-scale data diversity may only become apparent beyond simple locomotion tasks.

\paragraph{Initial Policy Standard Deviation (std).} We ablate the initial action distribution spread by comparing a tighter standard deviation of 0.2 against the baseline of 3.0. We observed that changing the initial action standard deviation had a negligible impact on standard KINESIS motions. However, our empirical training experience shows that higher initial variance is crucial for exploring dynamic motions, such as airkicking. We plan to further investigate the effectiveness of high initial std on dynamic motions in future work. 

\paragraph{Rollout Length.} We investigate the impact of the rollout horizon by comparing the trajectory lengths of 8 and 50 steps against the baseline of 20 steps. A clear trade-off is observed between gradient update frequency and advantange / value estimation accuracy. Truncated rollouts of 8 steps performed the worst, which is likely because truncated horizons introduce significant value estimation bias. Conversely, 50-step rollouts performed similarly to the 20-step baseline, but resulted in a reduced total number of updates over training with larger GPU memory usage.

\paragraph{GPU vs. CPU.} We also report the evaluations on CPU and GPU (MuJoCo Warp backend) separately in Tab. \ref{tab:ablation_results_cpu} and Tab. \ref{tab:ablation_results_gpu}. The performance metrics results are consistent across both simulators' backends, and the top three ablation configurations remain the same. Results confirmed that policies trained using MuscleMimic GPU-accelerated pipeline could be transferred to standard CPU environments with minimum numerical degradation or performance loss.

\begin{table}[ht]
\centering
\caption{\textbf{Ablation Study on CPU.} Evaluation of various hyperparameters on the final performance of the MuscleMimic pipeline. All distance metrics are reported in cm, and angles in degrees. The overall top three methods are highlighted, and the best individual metric across all configurations is bolded.}
\label{tab:ablation_results_cpu}

\colorlet{topone}{green!15}   
\colorlet{toptwo}{cyan!15}    
\colorlet{topthree}{yellow!20} 

\resizebox{\textwidth}{!}{%
\begin{tabular}{l *{10}{c}}
\toprule
\textbf{Methods} & \textbf{Success Rate ($\uparrow$)} & \textbf{J. Pos Err ($\downarrow$)} & \textbf{J. Vel Err ($\downarrow$)} & \textbf{Root XYZ ($\downarrow$)} & \textbf{Root Yaw ($\downarrow$)} & \textbf{RPos Err ($\downarrow$)} & \textbf{Site Abs ($\downarrow$)} & \textbf{Coverage ($\uparrow$)} & \textbf{Ep. Len ($\uparrow$)} & \textbf{Ep. Ret ($\uparrow$)} \\
\midrule
Baseline            & 80.76 $\pm$ 5.13 & 7.94 $\pm$ 0.42 & 35.97 $\pm$ 1.82 & 8.47 $\pm$ 0.36 & 2.63 $\pm$ 0.24 & 2.85 $\pm$ 0.28 & 15.37 $\pm$ 0.70 & 91.10 $\pm$ 2.47 & 513.4 $\pm$ 14.0 & 516.3 $\pm$ 26.3 \\
KINESIS (-)         & 36.63 $\pm$ 2.93 & 7.87 $\pm$ 0.27 & 30.09 $\pm$ 0.34 & 6.91 $\pm$ 0.28 & 3.53 $\pm$ 0.46 & 3.15 $\pm$ 0.14 & \textbf{12.90} $\pm$ 0.50 & 63.65 $\pm$ 0.86 & 358.4 $\pm$ 4.8  & 375.1 $\pm$ 8.2  \\
KINESIS (+)         & 84.77 $\pm$ 3.74 & 8.29 $\pm$ 0.08 & 38.02 $\pm$ 0.61 & 8.13 $\pm$ 0.37 & 2.82 $\pm$ 0.10 & 2.93 $\pm$ 0.03 & 14.84 $\pm$ 0.65 & 92.05 $\pm$ 1.57 & 518.8 $\pm$ 8.9  & 507.7 $\pm$ 7.9  \\
Num Steps = 8       & 53.09 $\pm$ 2.33 & 9.64 $\pm$ 0.01 & 38.77 $\pm$ 0.16 & 8.13 $\pm$ 0.10 & 4.64 $\pm$ 0.11 & 4.43 $\pm$ 0.07 & 15.54 $\pm$ 0.23 & 79.07 $\pm$ 1.22 & 445.5 $\pm$ 6.9  & 403.2 $\pm$ 4.9  \\
Num Steps = 50      & 78.70 $\pm$ 3.21 & 9.76 $\pm$ 0.05 & 42.36 $\pm$ 0.11 & 8.32 $\pm$ 0.07 & 3.09 $\pm$ 0.10 & 3.18 $\pm$ 0.03 & 15.23 $\pm$ 0.11 & 89.94 $\pm$ 1.02 & 506.8 $\pm$ 5.8  & 472.3 $\pm$ 6.3  \\
Low Std = 0.2       & 75.93 $\pm$ 3.34 & 8.67 $\pm$ 0.03 & 33.67 $\pm$ 0.18 & 8.77 $\pm$ 0.19 & 3.76 $\pm$ 0.11 & 3.51 $\pm$ 0.03 & 16.28 $\pm$ 0.36 & 90.25 $\pm$ 1.21 & 508.6 $\pm$ 6.8  & 518.2 $\pm$ 7.4  \\
Term Thresh = 0.25  & 66.05 $\pm$ 2.83 & 8.56 $\pm$ 0.03 & 40.58 $\pm$ 0.30 & \textbf{6.71} $\pm$ 0.14 & 2.73 $\pm$ 0.06 & 3.52 $\pm$ 0.06 & 12.96 $\pm$ 0.24 & 83.94 $\pm$ 0.43 & 472.9 $\pm$ 2.4  & 436.3 $\pm$ 4.3  \\
\rowcolor{topone}
Term Thresh = 1.0   & 90.12 $\pm$ 1.41  & \textbf{6.04} $\pm$ 0.00 & \textbf{25.34} $\pm$ 0.16 & 7.30 $\pm$ 0.09 & 1.54 $\pm$ 0.03 & 1.90 $\pm$ 0.02 & 13.09 $\pm$ 0.17 & 95.04 $\pm$ 0.50 & 535.6 $\pm$ 2.8  & \textbf{614.0} $\pm$ 4.0  \\
\rowcolor{toptwo}
Term Thresh = 10.0  & \textbf{90.43} $\pm$ 1.41  & 6.68 $\pm$ 0.01 & 26.56 $\pm$ 0.02 & 7.43 $\pm$ 0.15 & \textbf{1.50} $\pm$ 0.04 & \textbf{1.85} $\pm$ 0.00 & 13.27 $\pm$ 0.27 & \textbf{95.11} $\pm$ 0.41 & \textbf{536.0} $\pm$ 2.3  & 607.9 $\pm$ 2.6  \\
Param Size = 10M    & 83.44 $\pm$ 3.02 & 8.48 $\pm$ 0.44 & 39.24 $\pm$ 0.43 & 8.27 $\pm$ 0.27 & 3.06 $\pm$ 0.18 & 3.08 $\pm$ 0.03 & 15.17 $\pm$ 0.46 & 91.68 $\pm$ 1.74 & 516.7 $\pm$ 9.8  & 489.8 $\pm$ 9.9  \\
\rowcolor{topthree}
Param Size = 100M   & 87.55 $\pm$ 2.82 & 7.59 $\pm$ 0.09 & 34.94 $\pm$ 0.31 & 8.03 $\pm$ 0.27 & 2.20 $\pm$ 0.04 & 2.63 $\pm$ 0.04 & 14.54 $\pm$ 0.46 & 93.94 $\pm$ 1.01 & 529.4 $\pm$ 5.7  & 541.9 $\pm$ 6.6  \\
\bottomrule
\end{tabular}%
}
\end{table}

\begin{table}[ht]
\centering
\caption{\textbf{Ablation Study on GPU.} Evaluation of various hyperparameters on the final performance of MuscleMimic pipeline. All distance metrics are reported in cm, and angles in degrees. The overall top three methods are highlighted, and the best individual metric across all configurations is bolded.}
\label{tab:ablation_results_gpu}

\colorlet{topone}{green!15}   
\colorlet{toptwo}{cyan!15}    
\colorlet{topthree}{yellow!20} 

\resizebox{\textwidth}{!}{%
\begin{tabular}{l *{10}{c}}
\toprule
\textbf{Methods} & \textbf{Success Rate ($\uparrow$)} & \textbf{J. Pos Err ($\downarrow$)} & \textbf{J. Vel Err ($\downarrow$)} & \textbf{Root XYZ ($\downarrow$)} & \textbf{Root Yaw ($\downarrow$)} & \textbf{RPos Err ($\downarrow$)} & \textbf{Site Abs ($\downarrow$)} & \textbf{Coverage ($\uparrow$)} & \textbf{Ep. Len ($\uparrow$)} & \textbf{Ep. Ret ($\uparrow$)} \\
\midrule
Baseline            & 82.20 $\pm$ 8.02 & 7.93 $\pm$ 0.42 & 36.15 $\pm$ 1.93 & 8.14 $\pm$ 0.42 & 2.66 $\pm$ 0.25 & 2.84 $\pm$ 0.28 & 14.79 $\pm$ 0.81 & 91.06 $\pm$ 3.65 & 514.2 $\pm$ 20.6 & 517.2 $\pm$ 34.5 \\
KINESIS (-)         & 37.55 $\pm$ 2.18 & 7.84 $\pm$ 0.25 & 30.18 $\pm$ 0.21 & \textbf{6.61} $\pm$ 0.32 & 3.49 $\pm$ 0.44 & 3.13 $\pm$ 0.12 & \textbf{12.40} $\pm$ 0.62 & 63.38 $\pm$ 0.64 & 357.9 $\pm$ 3.6  & 374.9 $\pm$ 6.4  \\
KINESIS (+)         & 85.70 $\pm$ 2.70 & 8.27 $\pm$ 0.08 & 38.08 $\pm$ 0.58 & 7.97 $\pm$ 0.49 & 2.80 $\pm$ 0.07 & 2.94 $\pm$ 0.05 & 14.57 $\pm$ 0.82 & 92.96 $\pm$ 1.57 & 524.9 $\pm$ 8.9  & 513.5 $\pm$ 9.0  \\
Num Steps = 8       & 52.16 $\pm$ 1.41 & 9.67 $\pm$ 0.05 & 38.83 $\pm$ 0.22 & 8.25 $\pm$ 0.34 & 4.61 $\pm$ 0.13 & 4.44 $\pm$ 0.06 & 15.81 $\pm$ 0.58 & 77.75 $\pm$ 1.02 & 439.0 $\pm$ 5.8  & 397.4 $\pm$ 3.5  \\
Num Steps = 50      & 79.94 $\pm$ 5.66 & 9.73 $\pm$ 0.07 & 42.55 $\pm$ 0.05 & 8.22 $\pm$ 0.18 & 3.04 $\pm$ 0.15 & 3.19 $\pm$ 0.03 & 15.09 $\pm$ 0.29 & 91.45 $\pm$ 2.68 & 516.4 $\pm$ 15.1 & 481.0 $\pm$ 13.9 \\
Low Std = 0.2       & 77.16 $\pm$ 0.53 & 8.64 $\pm$ 0.04 & 33.58 $\pm$ 0.12 & 8.17 $\pm$ 0.18 & 3.74 $\pm$ 0.16 & 3.49 $\pm$ 0.03 & 15.24 $\pm$ 0.24 & 90.26 $\pm$ 0.99 & 509.7 $\pm$ 5.6  & 520.1 $\pm$ 5.4  \\
Term Thresh = 0.25  & 68.21 $\pm$ 0.53 & 8.53 $\pm$ 0.02 & 40.74 $\pm$ 0.03 & 6.69 $\pm$ 0.17 & 2.80 $\pm$ 0.04 & 3.51 $\pm$ 0.01 & 12.87 $\pm$ 0.26 & 84.00 $\pm$ 0.88 & 474.3 $\pm$ 5.0  & 437.2 $\pm$ 4.5  \\
\rowcolor{toptwo}
Term Thresh = 1.0   & 91.98 $\pm$ 1.41  & \textbf{6.01} $\pm$ 0.02 & \textbf{25.28} $\pm$ 0.02 & 6.97 $\pm$ 0.17 & 1.50 $\pm$ 0.02 & 1.88 $\pm$ 0.01 & 12.53 $\pm$ 0.27 & 95.46 $\pm$ 0.67 & 539.0 $\pm$ 3.8  & 618.8 $\pm$ 4.7  \\
\rowcolor{topone}
Term Thresh = 10.0  & \textbf{94.14} $\pm$ 1.07  & 6.68 $\pm$ 0.01 & 26.69 $\pm$ 0.05 & 6.98 $\pm$ 0.10 & \textbf{1.46} $\pm$ 0.02 & \textbf{1.81} $\pm$ 0.00 & 12.52 $\pm$ 0.17 & \textbf{96.58} $\pm$ 0.42 & \textbf{545.3} $\pm$ 2.4  & \textbf{619.0} $\pm$ 2.8  \\
Param Size = 10M    & 82.51 $\pm$ 1.70 & 8.46 $\pm$ 0.41 & 39.34 $\pm$ 0.39 & 8.23 $\pm$ 0.36 & 3.01 $\pm$ 0.14 & 3.08 $\pm$ 0.05 & 15.10 $\pm$ 0.61 & 91.50 $\pm$ 0.80 & 516.6 $\pm$ 4.5  & 489.6 $\pm$ 4.3  \\
\rowcolor{topthree}
Param Size = 100M   & 87.96 $\pm$ 4.04 & 7.56 $\pm$ 0.07 & 35.06 $\pm$ 0.36 & 7.95 $\pm$ 0.24 & 2.22 $\pm$ 0.05 & 2.62 $\pm$ 0.04 & 14.41 $\pm$ 0.40 & 93.69 $\pm$ 1.37 & 529.0 $\pm$ 7.7  & 541.3 $\pm$ 8.3  \\
\bottomrule
\end{tabular}%
}
\end{table}

\begin{figure}[p]
    \centering
    
    \begin{subfigure}[b]{0.48\textwidth}
        \includegraphics[width=\textwidth]{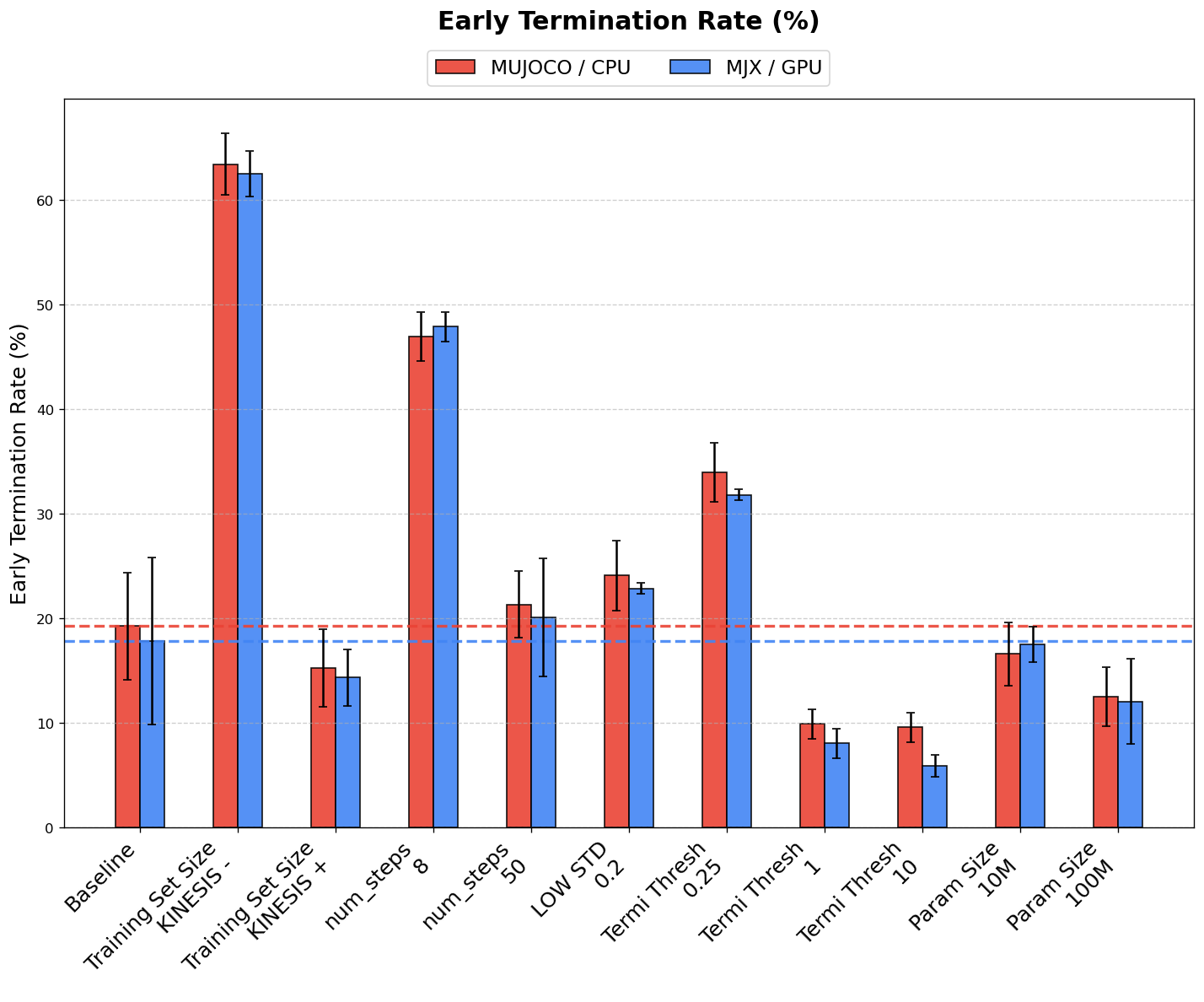}
    \end{subfigure}
    \hfill
    \begin{subfigure}[b]{0.48\textwidth}
        \includegraphics[width=\textwidth]{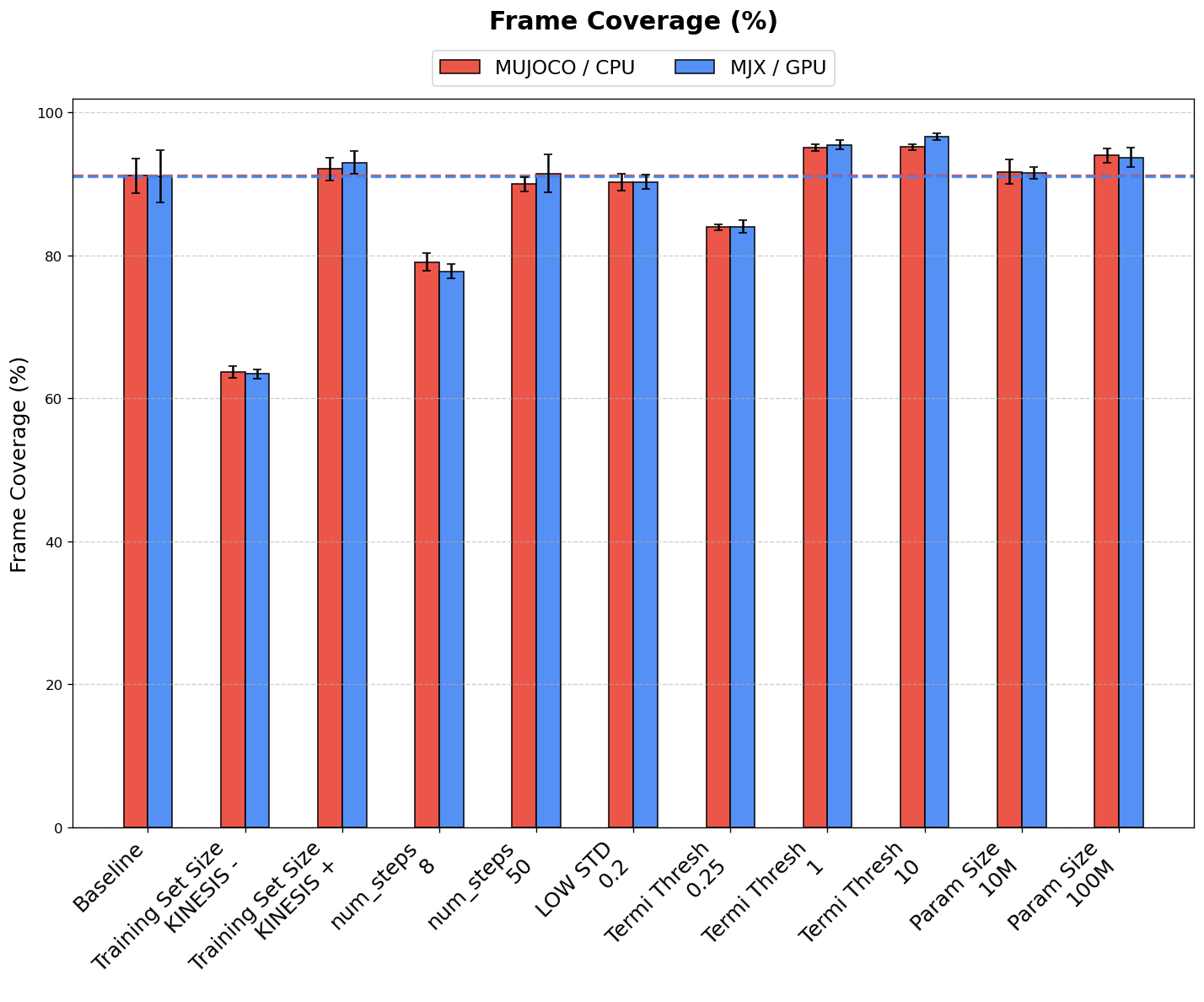}
    \end{subfigure}

    \vspace{10pt}

    \begin{subfigure}[b]{0.48\textwidth}
        \includegraphics[width=\textwidth]{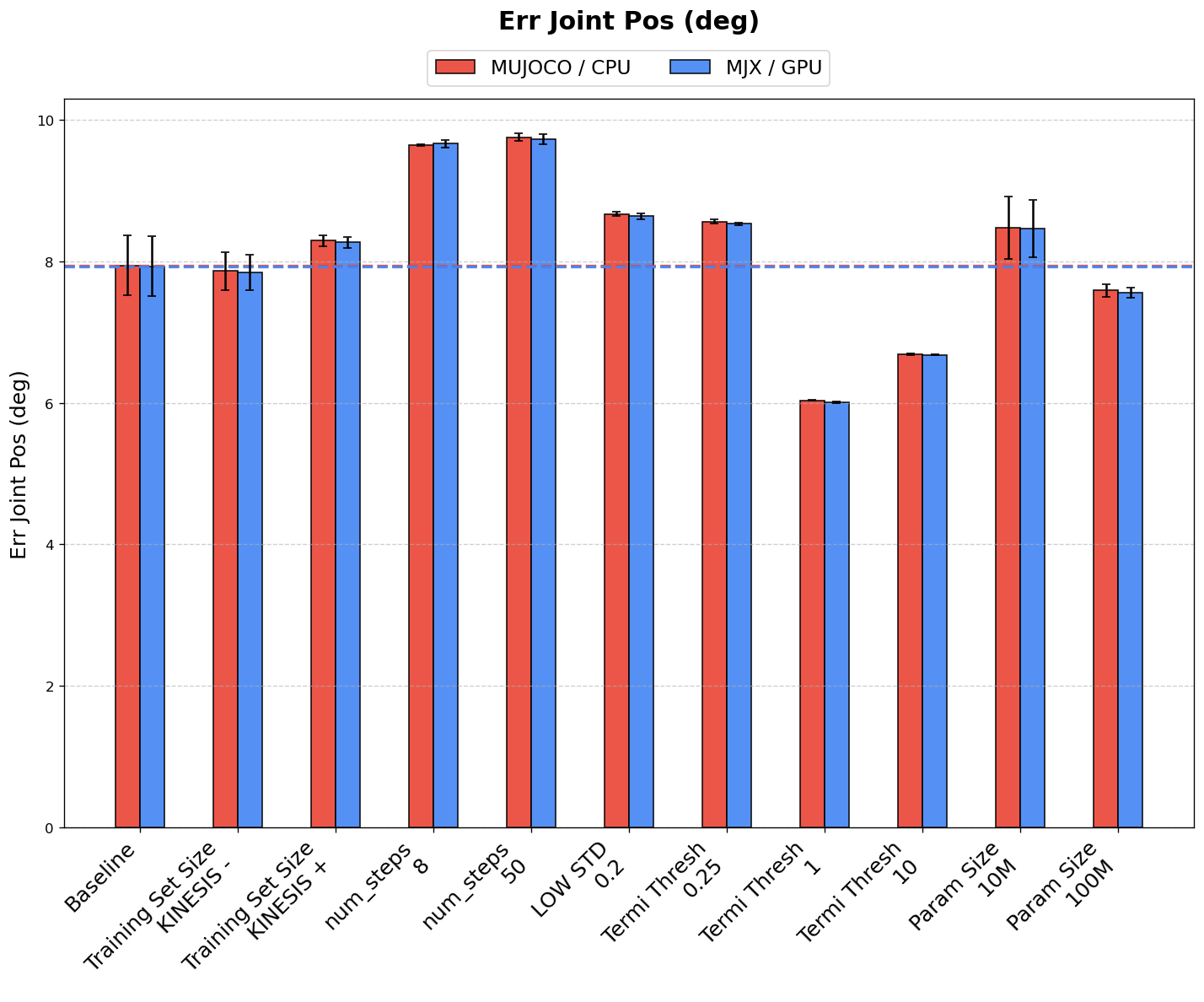}
        \caption{Joint Pos Error}
    \end{subfigure}
    \hfill
    \begin{subfigure}[b]{0.48\textwidth}
        \includegraphics[width=\textwidth]{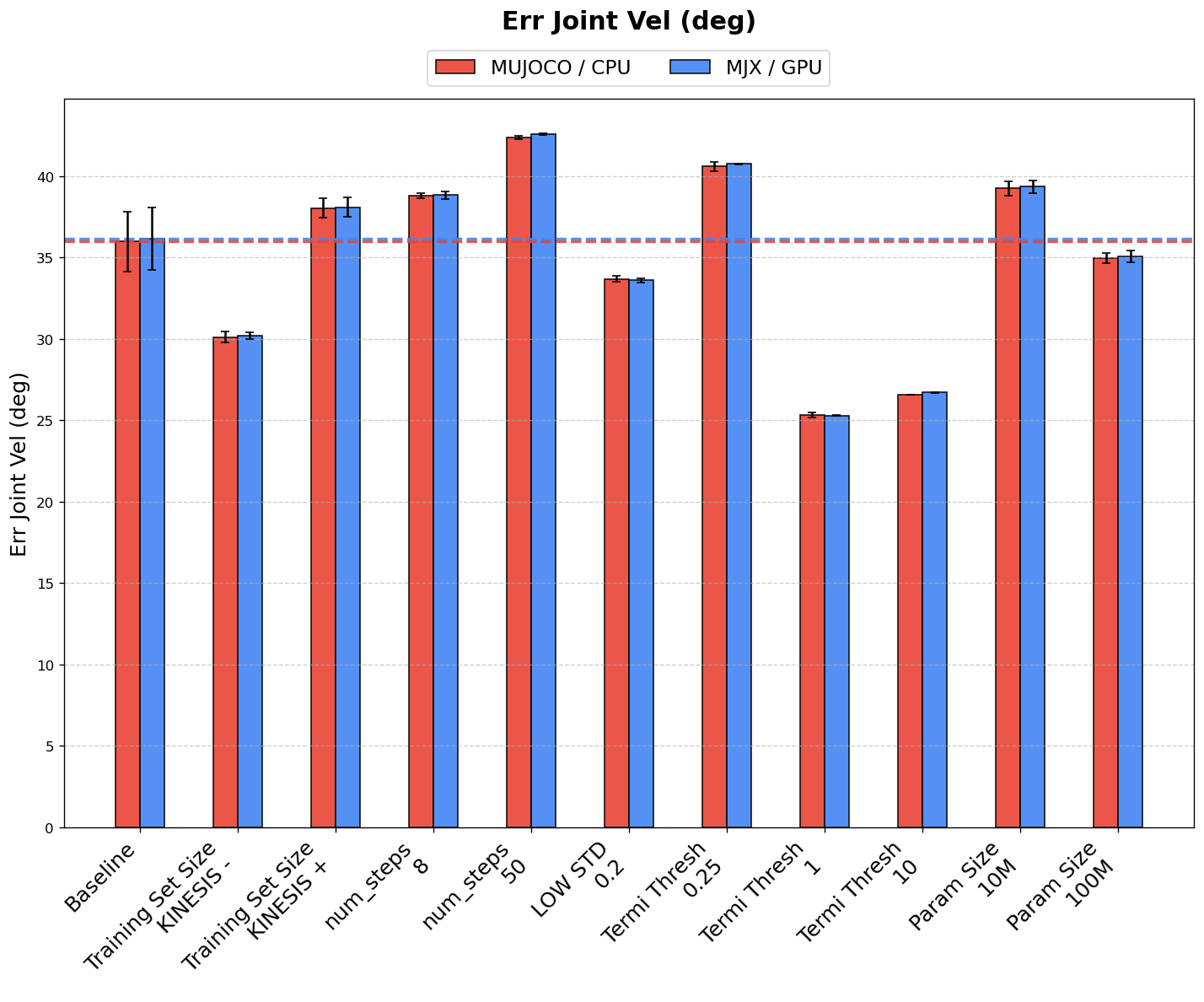}
    \end{subfigure}

    \vspace{10pt}

    \begin{subfigure}[b]{0.48\textwidth}
        \includegraphics[width=\textwidth]{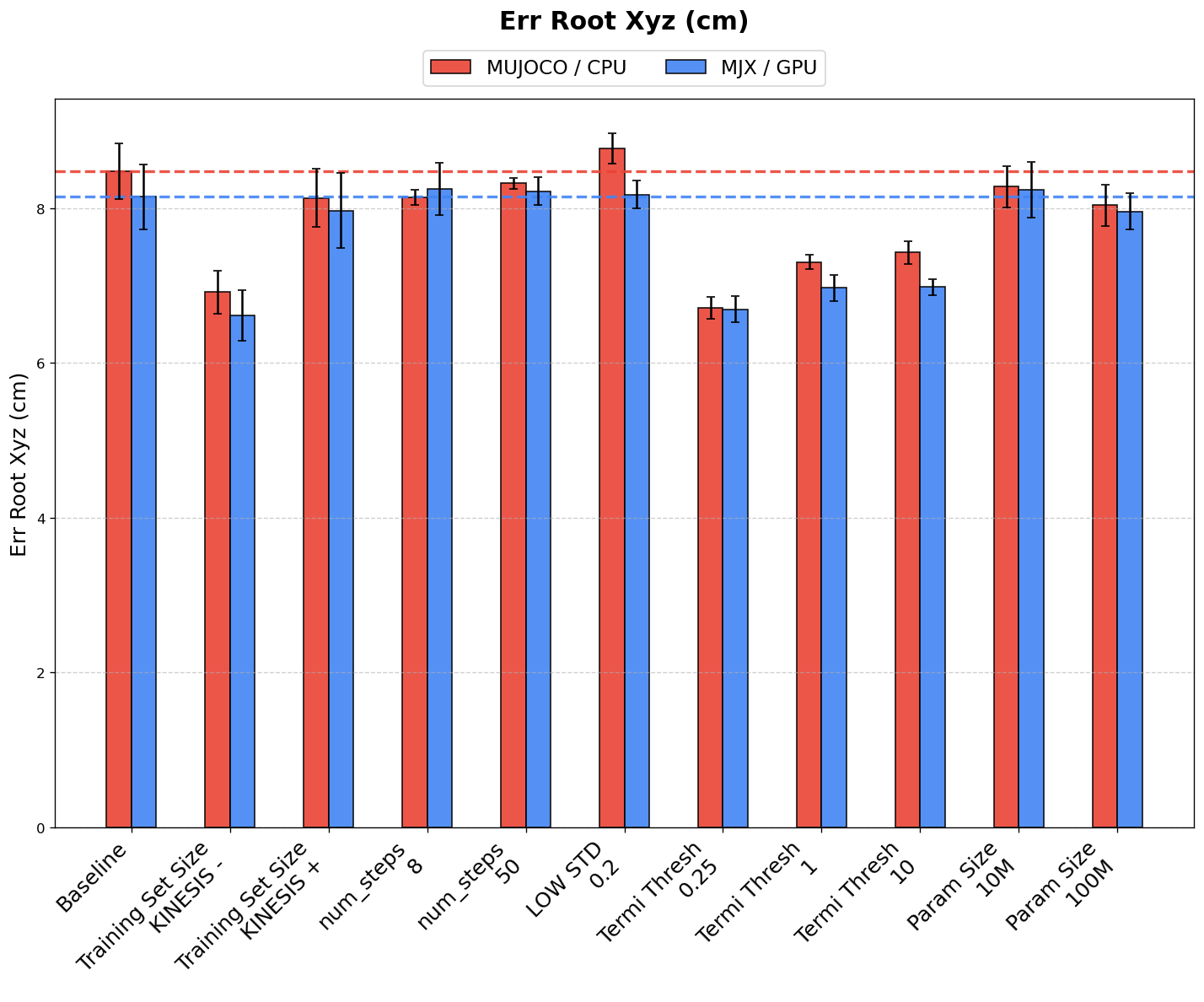}
    \end{subfigure}
    \hfill
    \begin{subfigure}[b]{0.48\textwidth}
        \includegraphics[width=\textwidth]{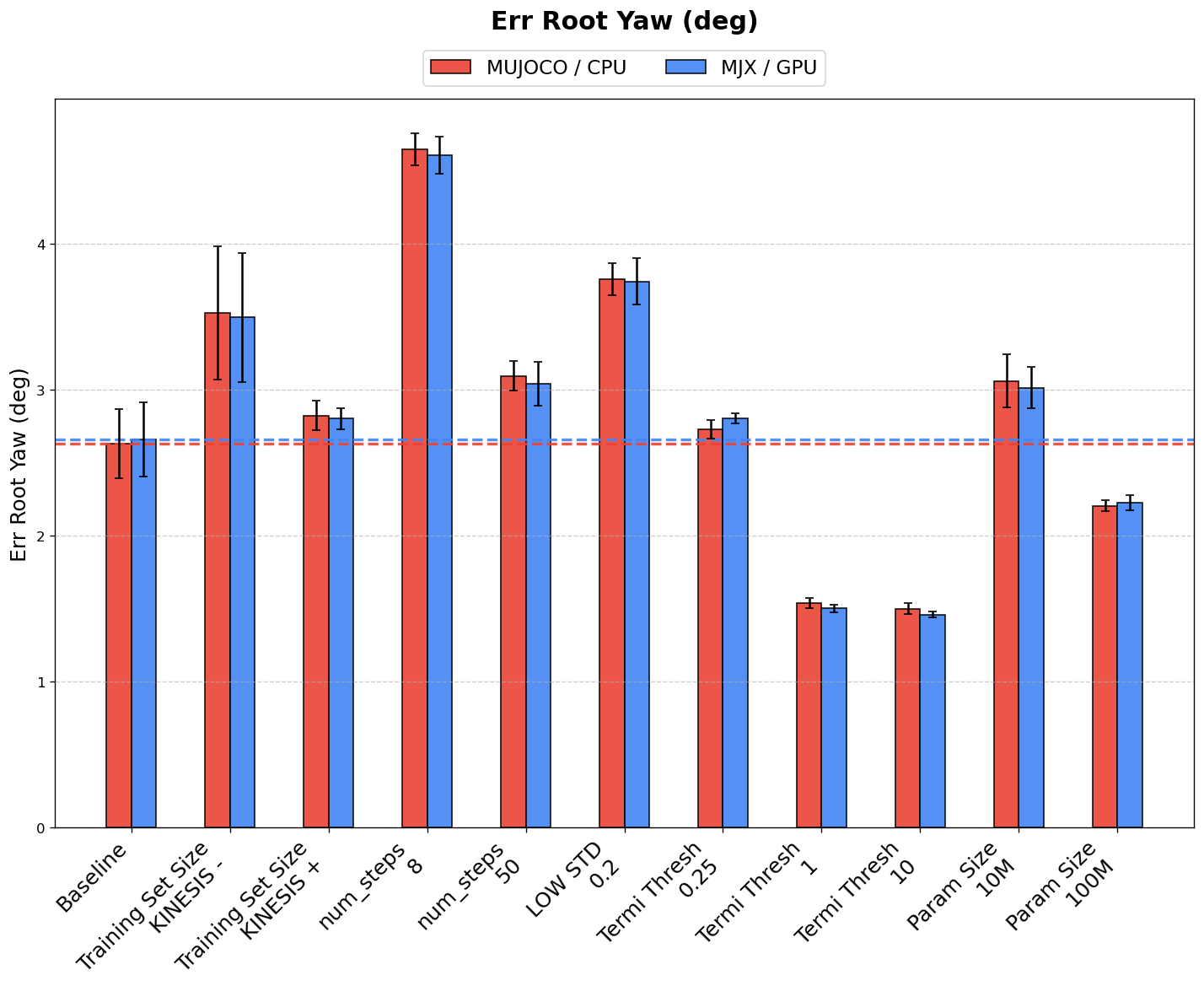}
    \end{subfigure}

    \caption{Comprehensive ablation study results for the MuscleMimic framework (Part I).}
\end{figure}

\begin{figure}[p]
    \centering
    \ContinuedFloat %
    
    \begin{subfigure}[b]{0.48\textwidth}
        \includegraphics[width=\textwidth]{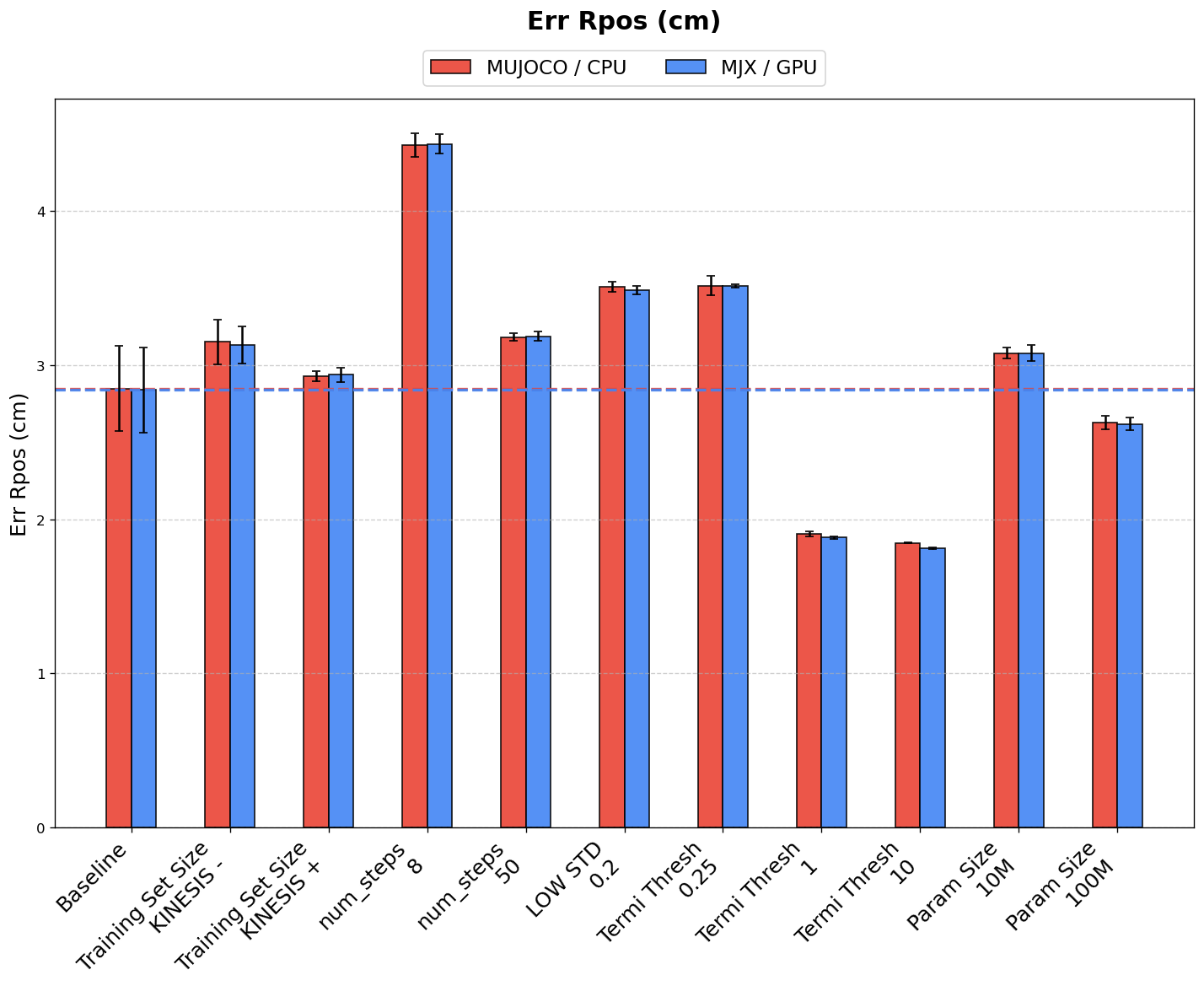}
    \end{subfigure}
    \hfill
    \begin{subfigure}[b]{0.48\textwidth}
        \includegraphics[width=\textwidth]{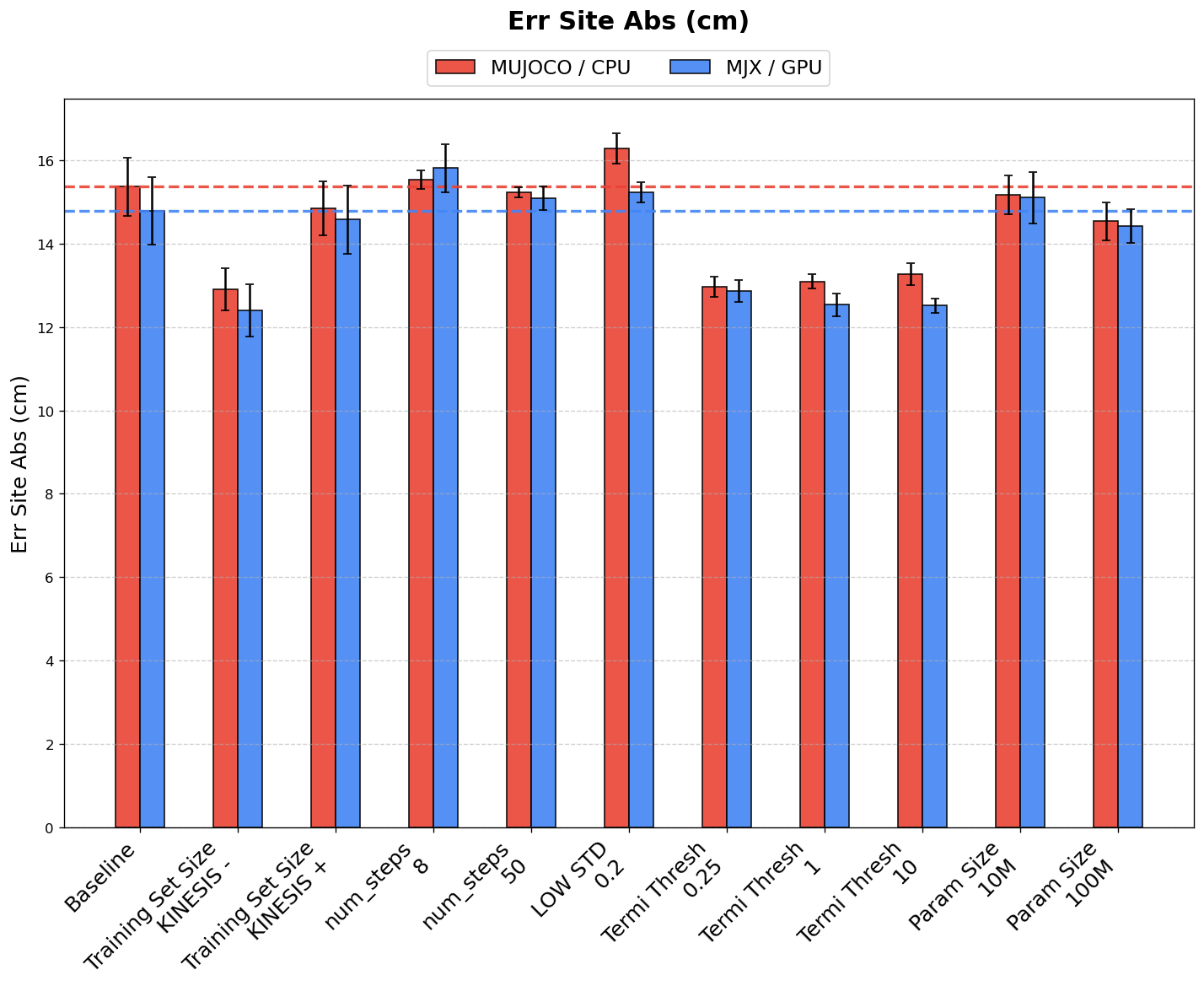}
    \end{subfigure}

    \vspace{10pt}

    \begin{subfigure}[b]{0.48\textwidth}
        \includegraphics[width=\textwidth]{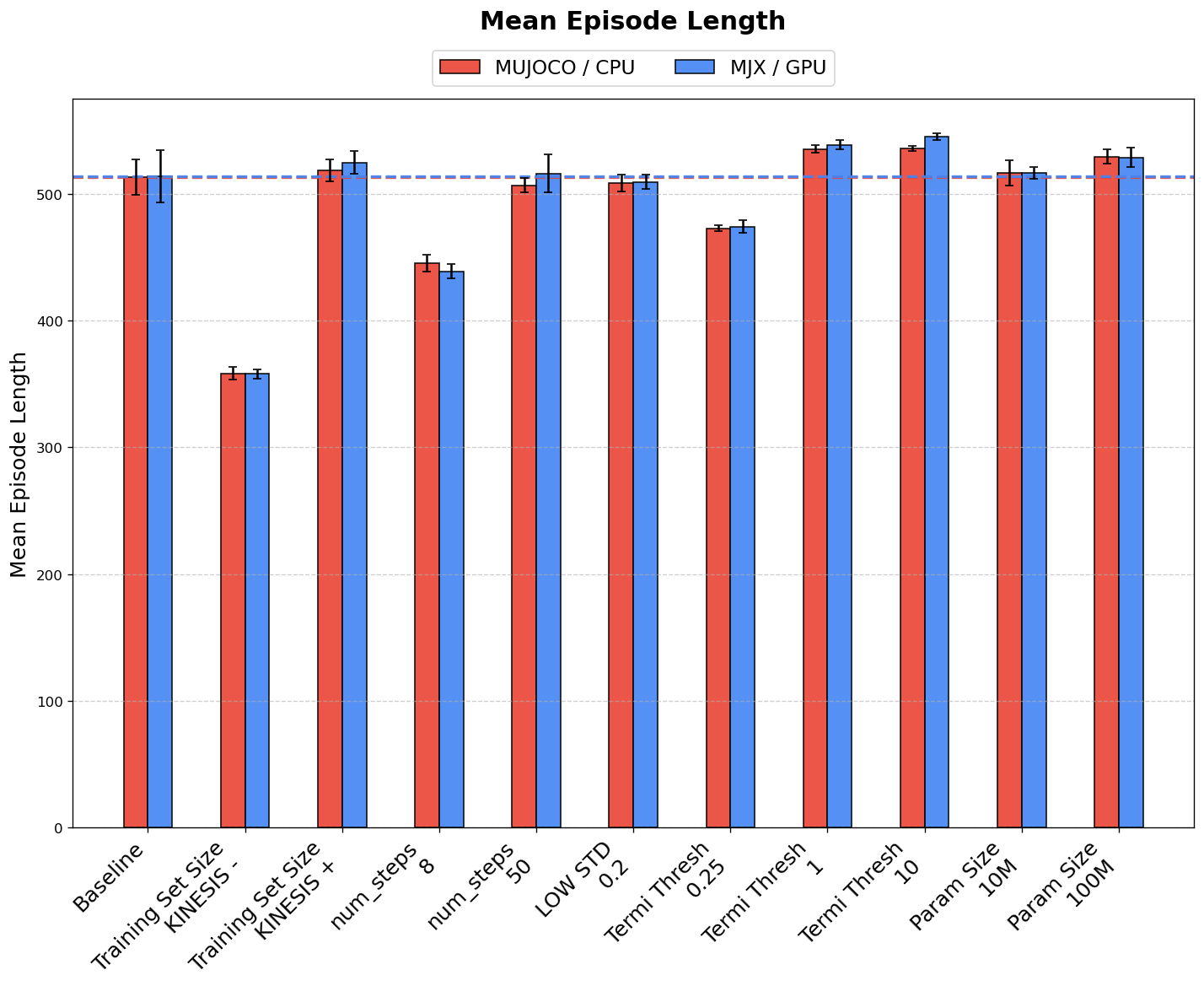}
    \end{subfigure}
    \hfill
    \begin{subfigure}[b]{0.48\textwidth}
        \includegraphics[width=\textwidth]{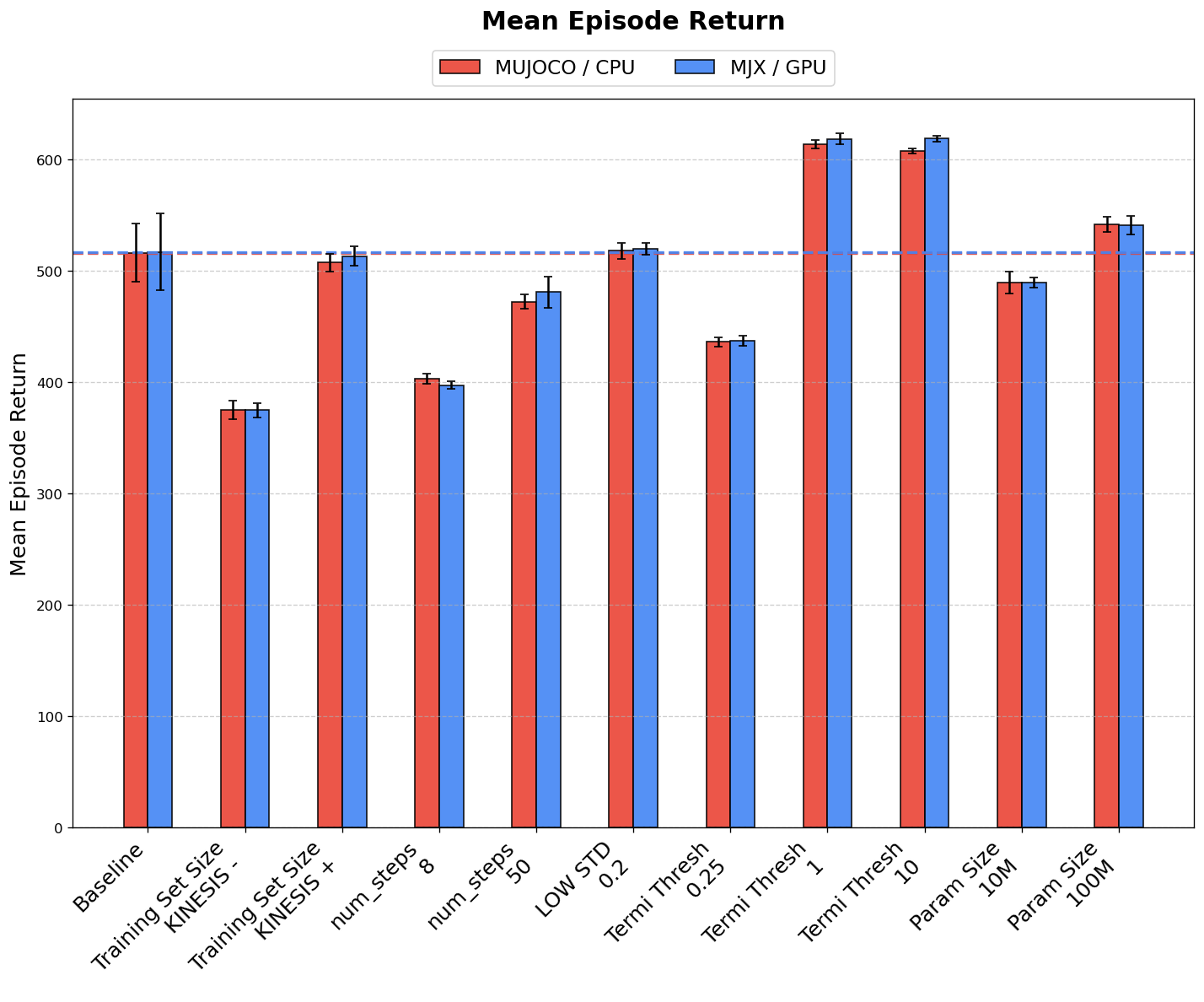}
    \end{subfigure}

    \caption{Comprehensive ablation study results for the MuscleMimic framework (Part II).}
    \label{fig:full_ablation}
\end{figure}

\section{Muscle Validation}\label{app:muscle_valid}
\subsection{Muscle symmetry}
Both \textbf{MyoBimanualArm} and \textbf{MyoFullBody} are finetuned to be perfectly symmetric in terms of joint equality constraint, joint ranges, muscle moment arm and muscle force-length (FL) curves. We show two examples below in Figure \ref{fig:muscle_symm_comparison}.

\begin{figure*}[h]
    \centering
    \begin{subfigure}[b]{0.48\textwidth}
        \centering
        \includegraphics[width=\textwidth]{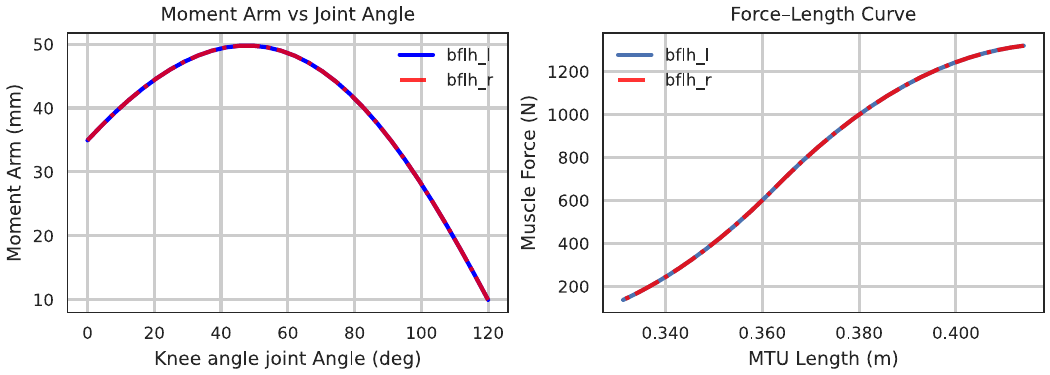}
        \caption{Bflh across Knee Flexion}
        \label{fig:symm_bflh_knee}
    \end{subfigure}
    \hfill
    \begin{subfigure}[b]{0.48\textwidth}
        \centering
        \includegraphics[width=\textwidth]{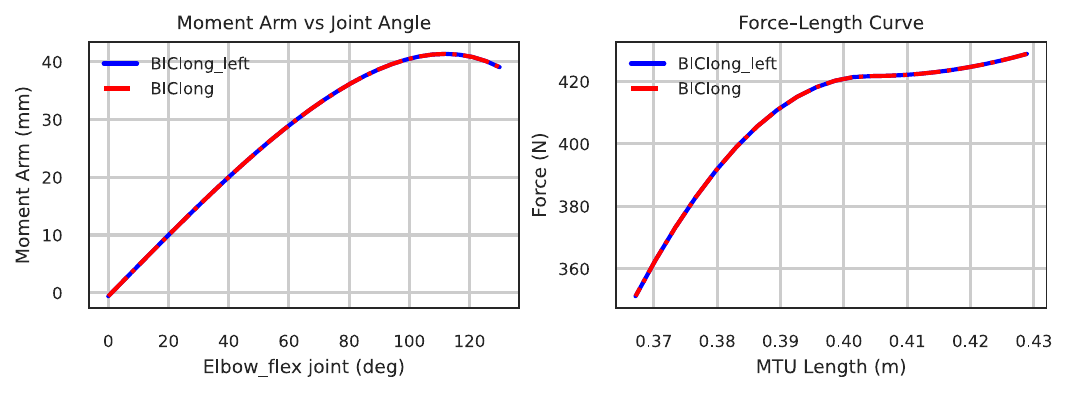}
        \caption{BIClong across Elbow Flexion}
        \label{fig:symm_biclong_elbow}
    \end{subfigure}

    \vspace{0.5em}

    \begin{subfigure}[b]{0.48\textwidth}
        \centering
        \includegraphics[width=\textwidth]{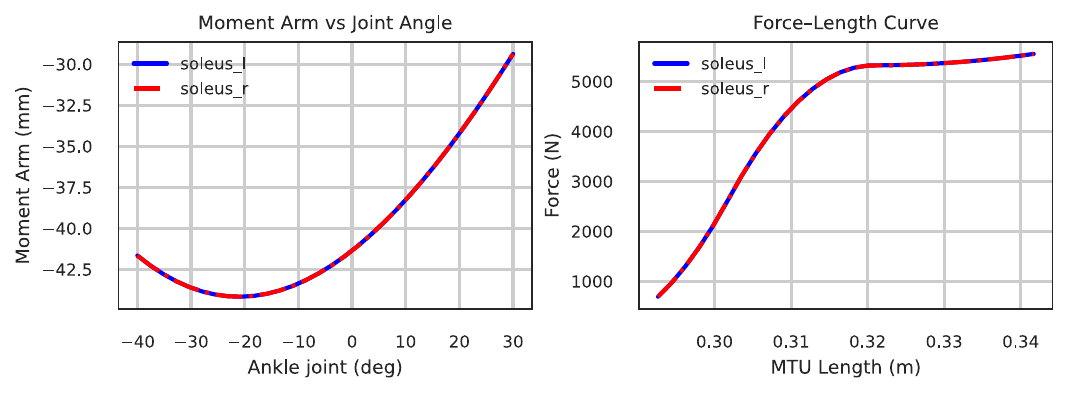}
        \caption{Rectus Femoris across Hip Flexion}
        \label{fig:symm_rectfem_hip}
    \end{subfigure}
    \hfill
    \begin{subfigure}[b]{0.48\textwidth}
        \centering
        \includegraphics[width=\textwidth]{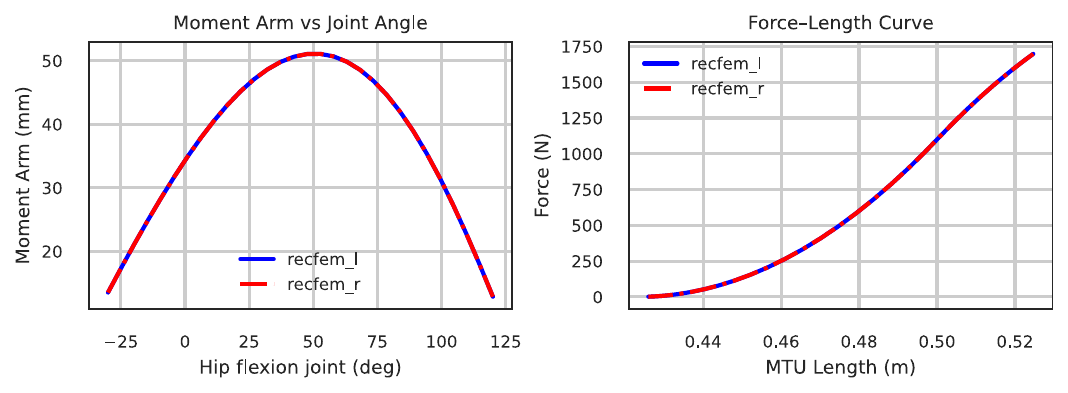}
        \caption{Soleus across Ankle Flexion}
        \label{fig:symm_soleus_ankle}
    \end{subfigure}

    \caption{Example validation of symmetry between left and right muscle-tendon groups of MyoFullBody.}
    \label{fig:muscle_symm_comparison}
\end{figure*}

\subsection{Muscle Jump Refinement}\label{app:jump}
\begin{figure*}[!h]
    \centering
    \begin{subfigure}[b]{0.48\textwidth}
        \centering
        \includegraphics[width=\textwidth]{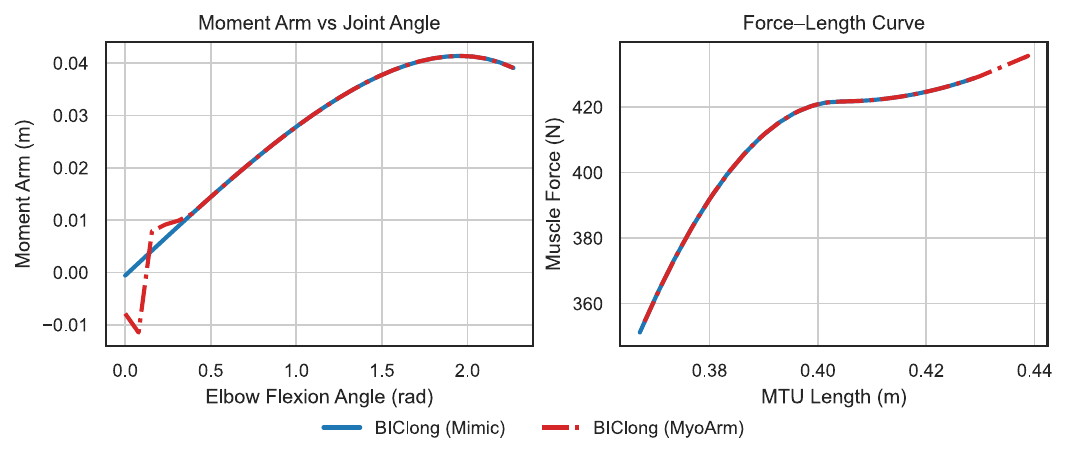}
        \caption{BIClong across Elbow Flexion}
        \label{fig:compar2}
    \end{subfigure}
    \hfill
    \begin{subfigure}[b]{0.48\textwidth}
        \centering
        \includegraphics[width=\textwidth]{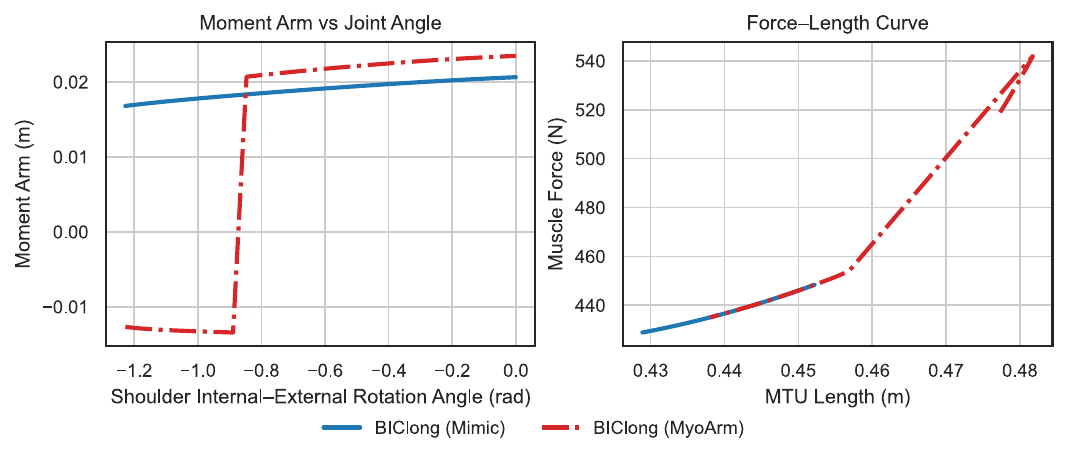}
        \caption{BIClong across Shoulder Internal-External Rotation}
        \label{fig:compar3}
    \end{subfigure}

    \vspace{0.5em}

    \begin{subfigure}[b]{0.48\textwidth}
        \centering
        \includegraphics[width=\textwidth]{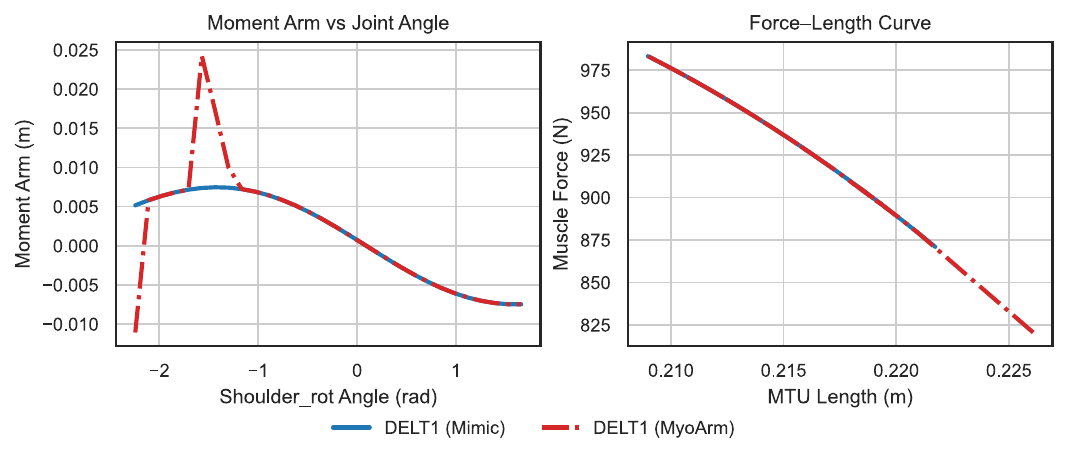}
        \caption{DELT1 across Shoulder Rotation}
        \label{fig:compar5}
    \end{subfigure}
    \hfill
    \begin{subfigure}[b]{0.48\textwidth}
        \centering
        \includegraphics[width=\textwidth]{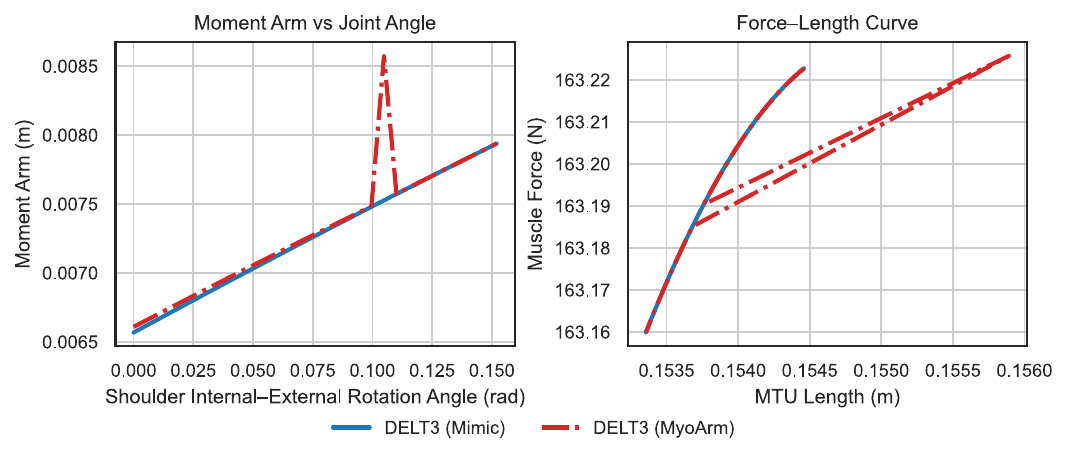}
        \caption{DELT3 across Shoulder Internal-External Rotation}
        \label{fig:compar6}
    \end{subfigure}

    \vspace{0.5em}

    \begin{subfigure}[b]{0.48\textwidth}
        \centering
        \includegraphics[width=\textwidth]{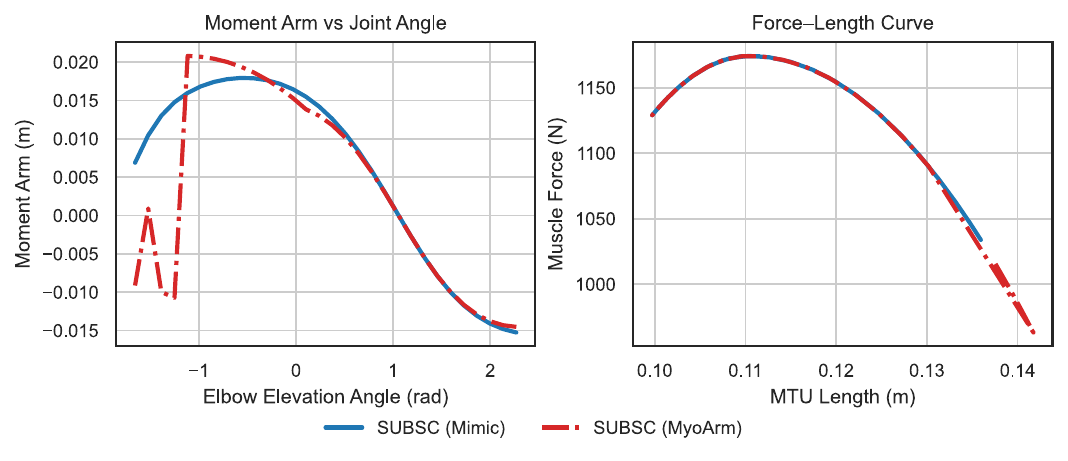}
        \caption{Subscapularis across elbow elevation}
        \label{fig:symm1}
    \end{subfigure}
    \hfill
    \begin{subfigure}[b]{0.48\textwidth}
        \centering
        \includegraphics[width=\textwidth]{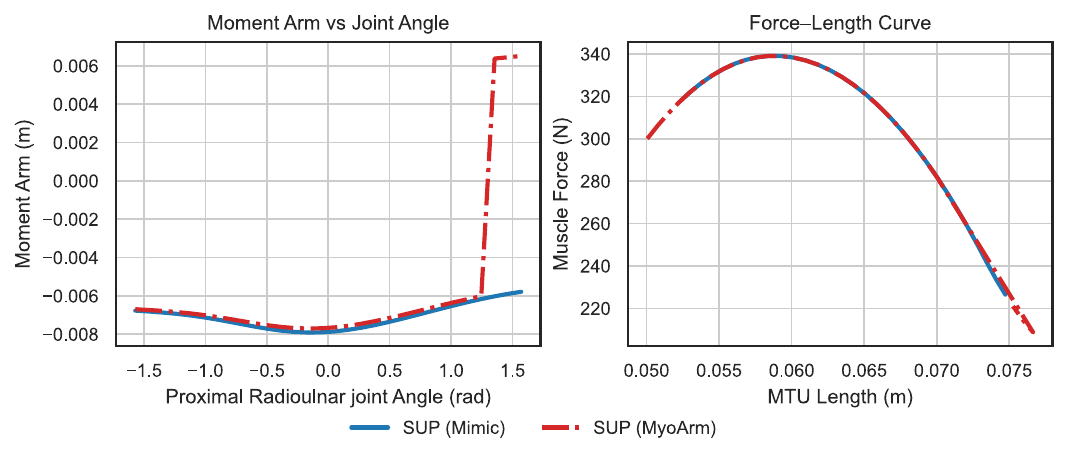}
        \caption{Supinator across the proximal radioulnar joint}
        \label{fig:symm2}
    \end{subfigure}

    \vspace{0.5em}

    \begin{subfigure}[b]{0.48\textwidth}
        \centering
        \includegraphics[width=\textwidth]{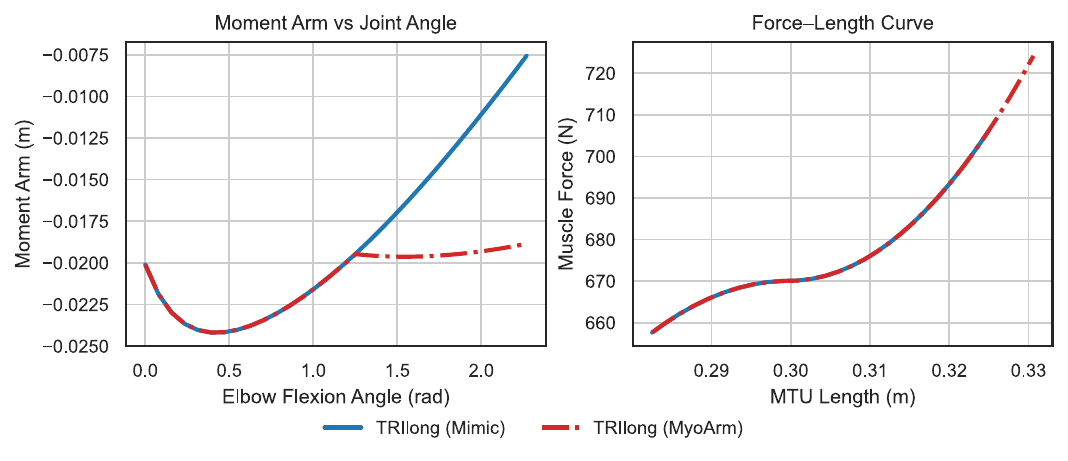}
        \caption{Triceps long head across elbow flexion}
        \label{fig:symm1}
    \end{subfigure}
    \hfill
    \begin{subfigure}[b]{0.48\textwidth}
        \centering
        \includegraphics[width=\textwidth]{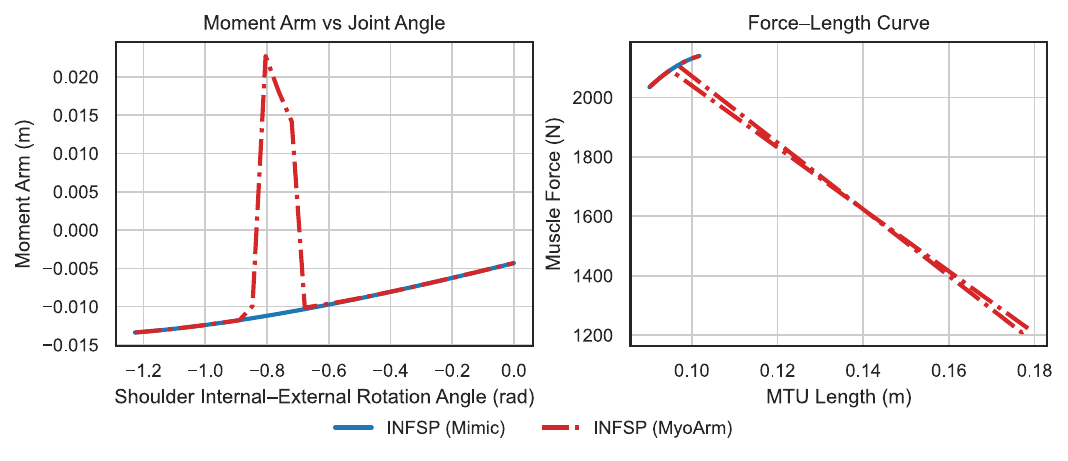}
        \caption{Infraspinatus across shoulder internal–external rotation}
        \label{fig:symm2}
    \end{subfigure}

    \caption{Comparison of right-arm muscle moment arm and force–length profiles between the MyoArm model (pre-tuning) and the Mimic-based MyoBimanualArm model (post-tuning).}
    \label{fig:arm_muscle_comparison}
\end{figure*}

As part of building the MyoBimanualArm model and refining MyoFullBody, each muscle–tendon moment arm was cross-validated against its target joint to ensure continuity and the absence of sudden jumps (Fig. \ref{fig:arm_muscle_comparison}, dashed red). Whenever discontinuities were observed, the corresponding wrapping geometry was manually corrected and fine-tuned to achieve smooth and consistent moment-arm and force-generating profiles. Most refinement of muscle routes concentrated around the shoulder joint, in which multiple joint equality constraints are enforced. A few asymmetries in the muscle actuator properties and knee equality constraints are identified within the lower limb and cross-matched. In total, around 150 asymmetries and muscle jumps were fixed in the final version compared to the previous myoArm and myoLegs.

\subsection{Muscle Validation with Experimental Data}
The biofidelity of the MyoFullBody and MyoBimanualArm was validated by comparing the moment arms of each muscle calculated in the model with those measured on human subjects, either from cadaver or MRI ~\citep{Delp1990SurgerySimulation,Visser1990,Spoor1992,Herzog1993LinesOfAction,Hintermann1994,Pigeon1996, Wretenberg1996,Klein1996,Aper1996,Buford1997,Kellis1999,Arnold2000,Buford2001,PIAZZA2003,Ackland2008, Wilson2009,Sobczak2013,Fiorentino2013,Snoeck2021,Wang2021,Chen2025}. A muscle’s moment arm characterizes the mapping between muscle force and the resulting joint moment. 
For a given muscle $m$ acting about joint $j$, the magnitude of the moment arm $|r_{j,m}|$ is computed as
\begin{equation}
    |r_{j,m}| = \frac{\Delta l_{\mathrm{MTC},m}}{\Delta \theta_j},
\end{equation}
where $\Delta l_{\mathrm{MTC},m}$ denotes a small change in the muscle--tendon complex (MTC) length of muscle $m$, and $\Delta \theta_j$ denotes the corresponding change in the joint angle $\theta_j$.  The MTC length $l_{\mathrm{MTC},m}$ is defined as the sum of the lengths of the action lines connecting the muscle attachment sites across the rigid bodies spanned by the muscle ~\cite{ATSUMI2025}. Representative examples of these comparisons are shown in Fig.~\ref{fig:muscle_exp_comparison}, with the complete set of results available in our GitHub repository. Given the known variability in experimentally reported moment arms arising from differences in measurement methodologies and inter-individual anatomical variation, our simulated moment arms are evaluated with respect to the overall range of reported experimental values. Within this context, the simulation results are generally consistent with the experimentally observed ranges.

\begin{figure*}[!t]
    \centering
    \begin{subfigure}[b]{0.38\textwidth}
        \centering
        \includegraphics[width=\linewidth]{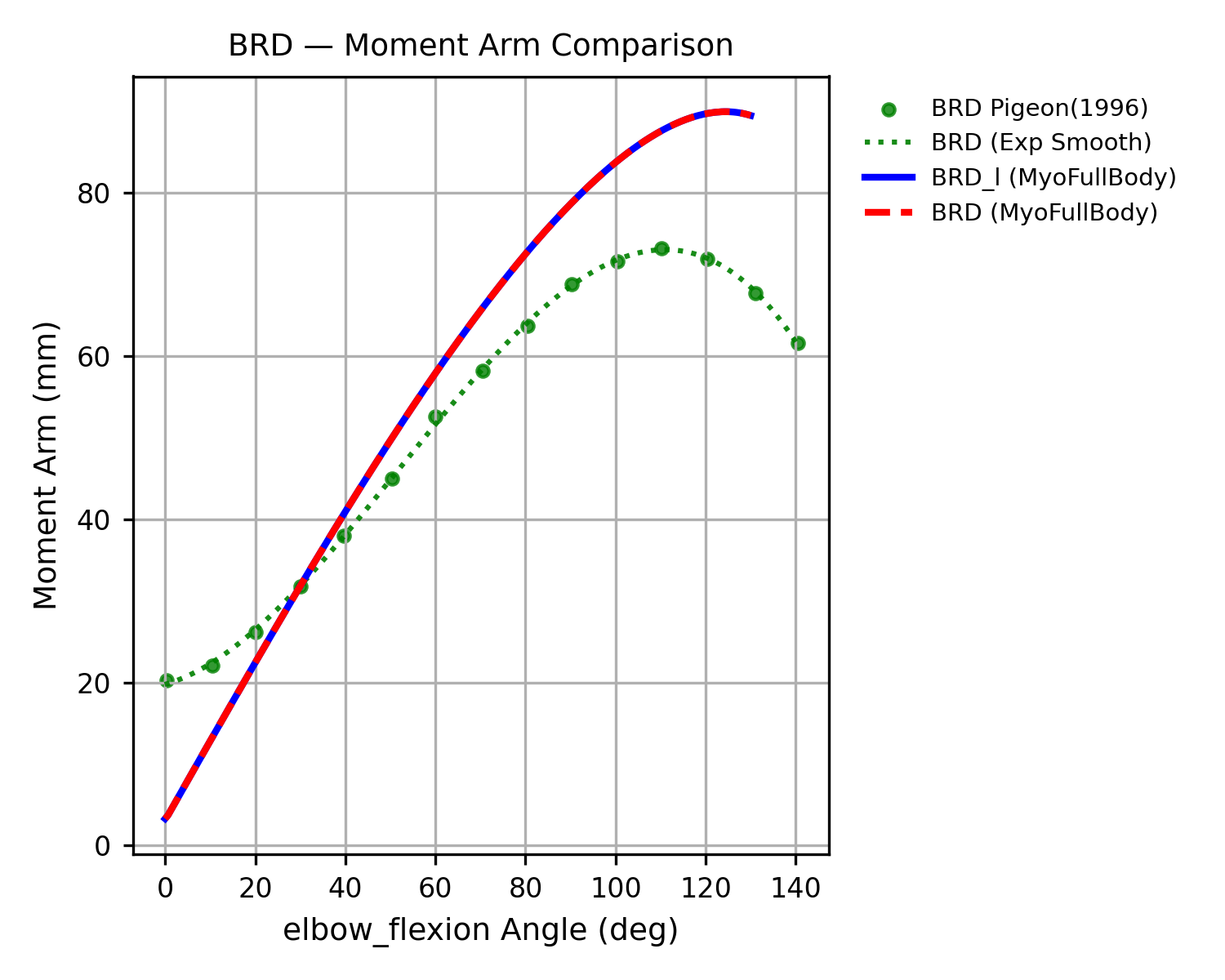}
        \label{fig:comparexp1}
    \end{subfigure}
    \hfill
    \begin{subfigure}[b]{0.61\textwidth}
        \centering
        \includegraphics[width=\linewidth]{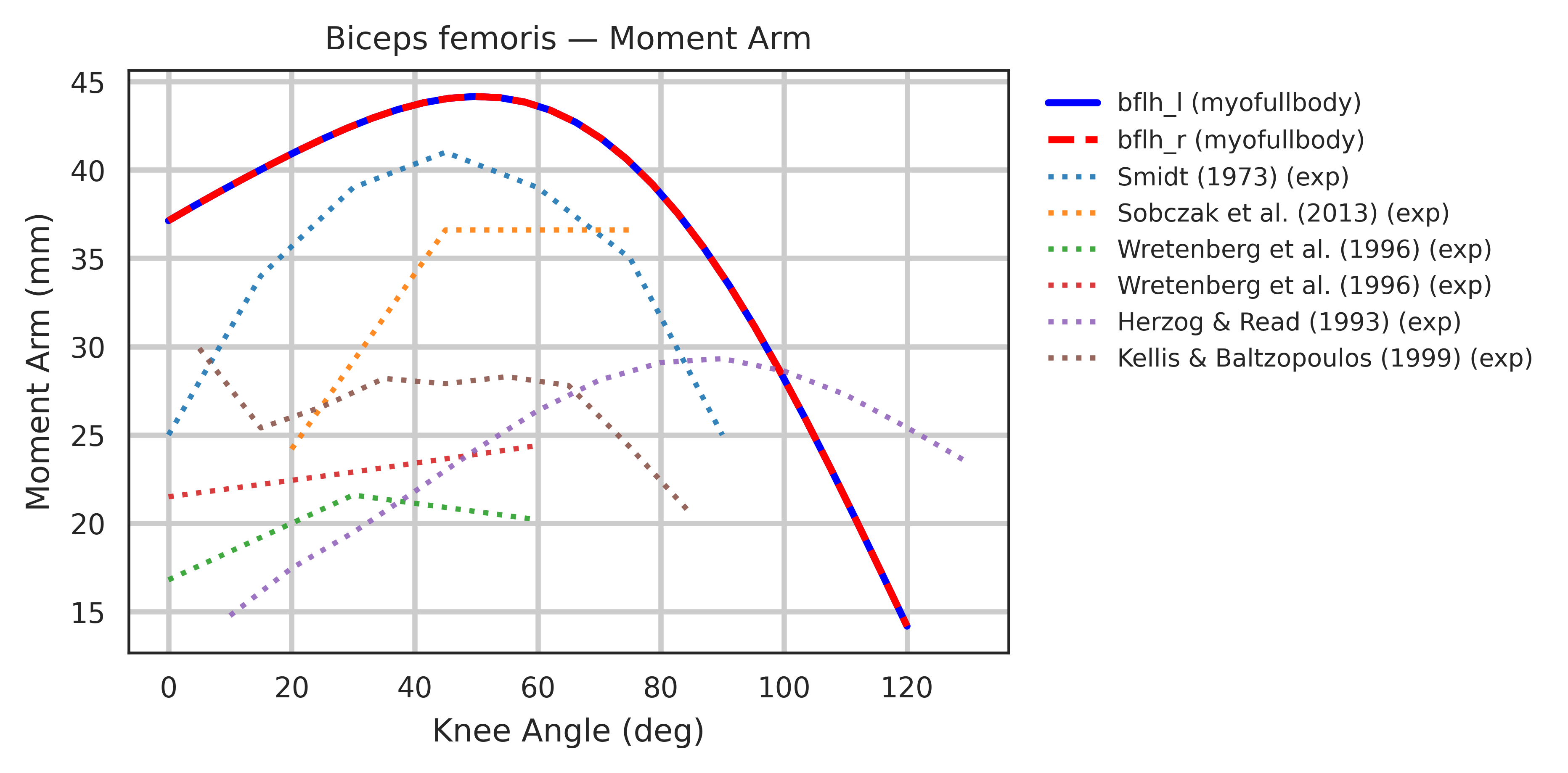}
        \label{fig:comparexp2}
    \end{subfigure}

    \vspace{0.5em}

    \begin{subfigure}[b]{0.38\textwidth}
        \centering
        \includegraphics[width=\linewidth]{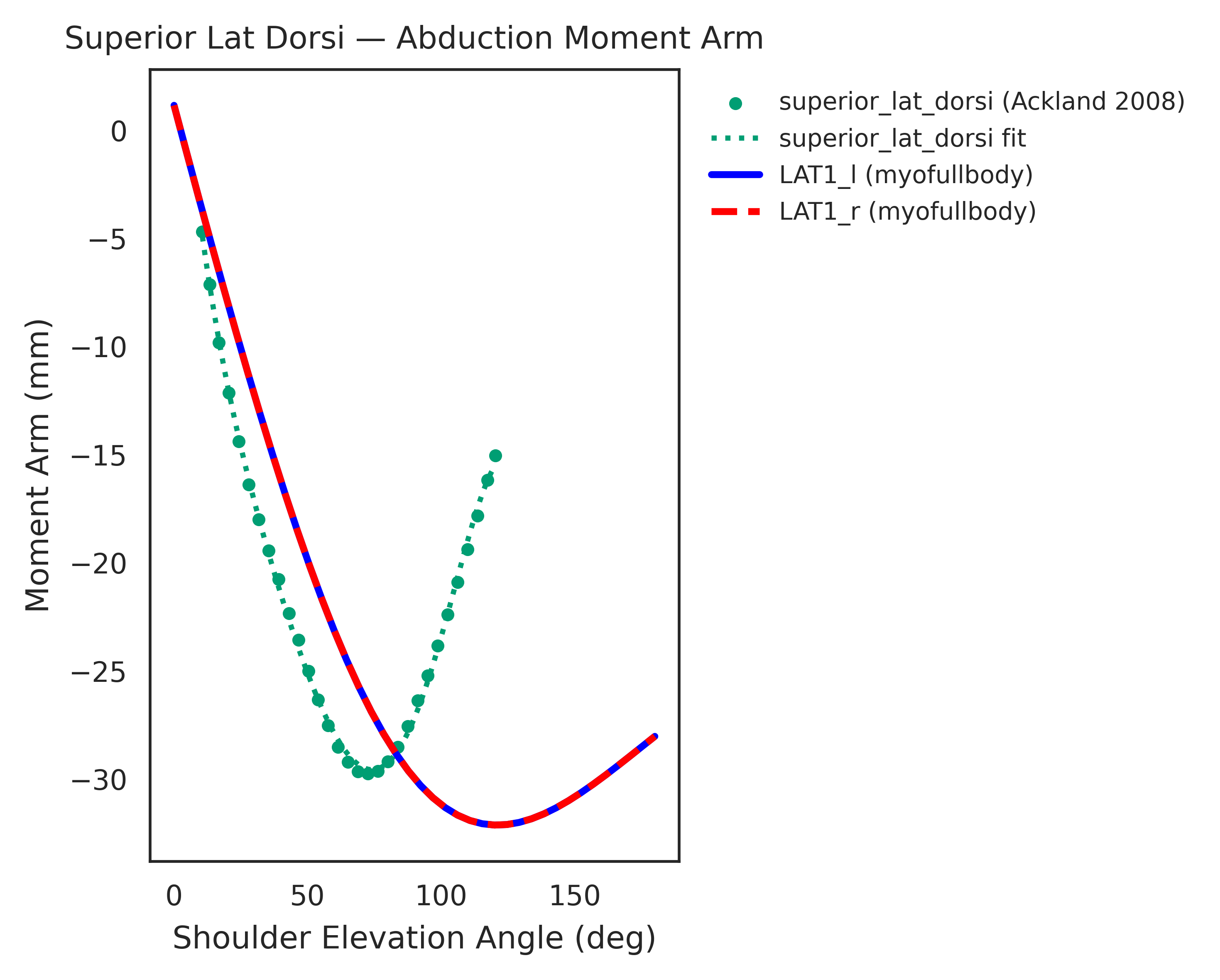}
        \label{fig:comparexp3}
    \end{subfigure}
    \hfill
    \begin{subfigure}[b]{0.61\textwidth}
        \centering
        \includegraphics[width=\linewidth]{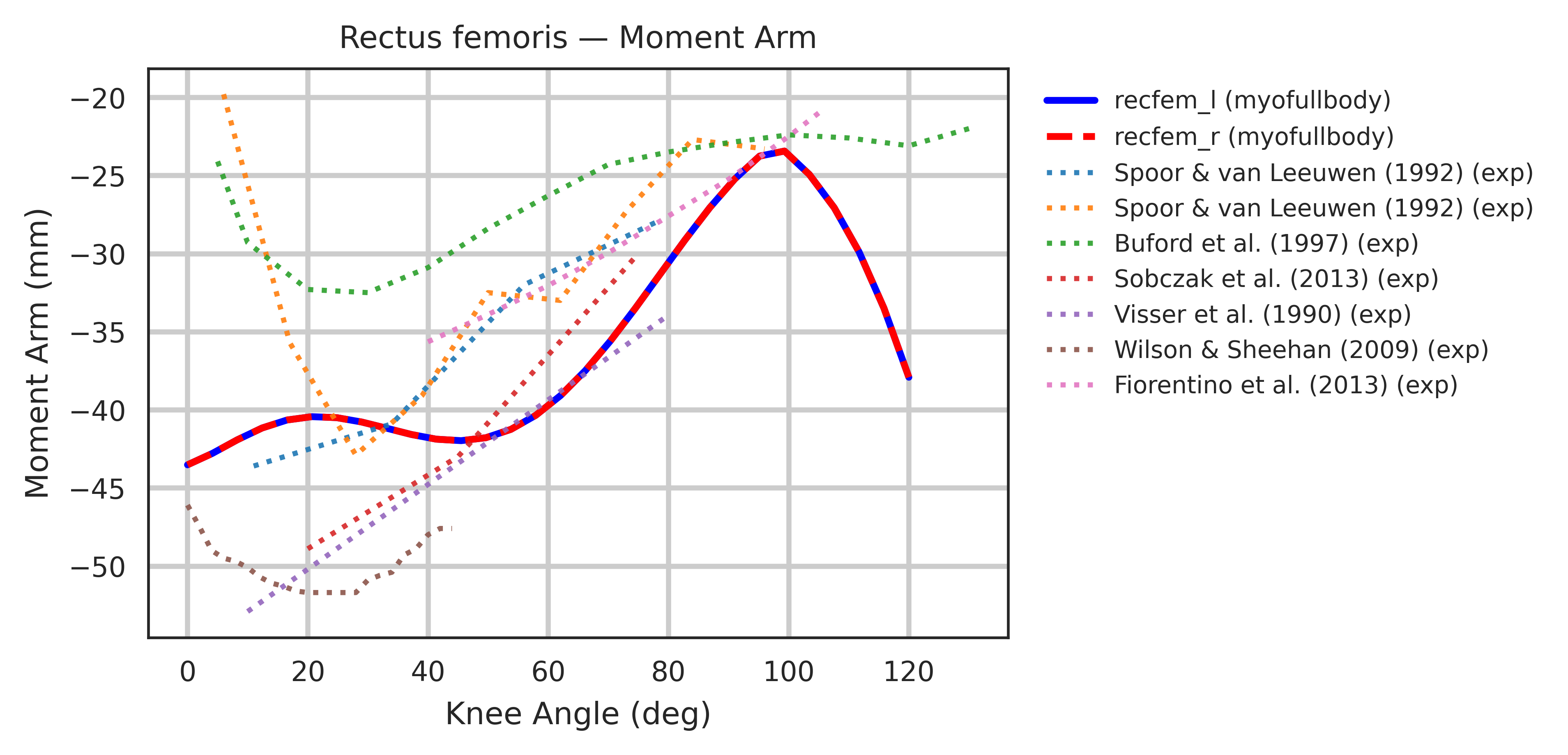}
        \label{fig:comparexp4}
    \end{subfigure}

    \vspace{0.5em}

    \begin{subfigure}[b]{0.38\textwidth}
        \centering
        \includegraphics[width=\linewidth]{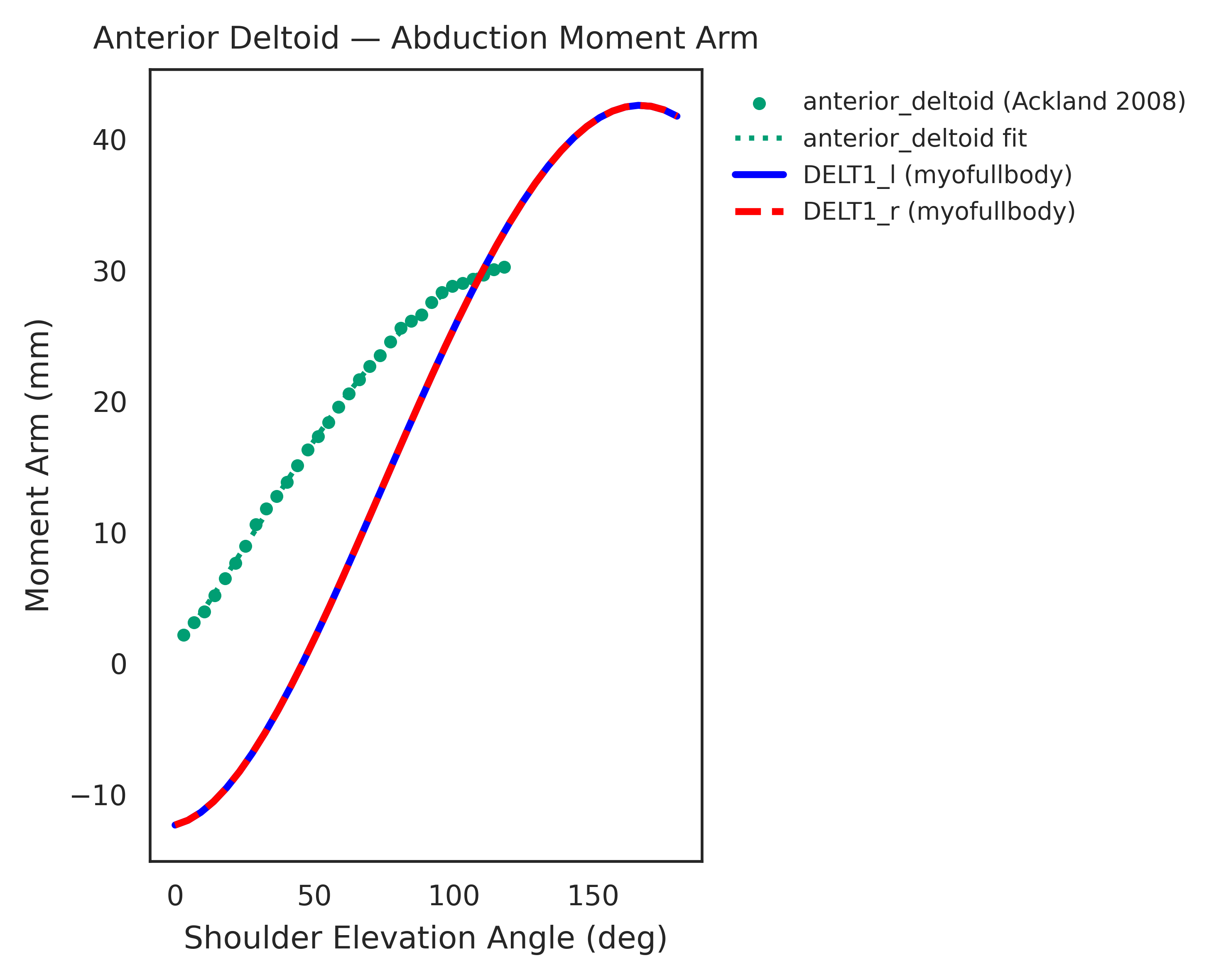}
        \label{fig:comparexp3}
    \end{subfigure}
    \hfill
    \begin{subfigure}[b]{0.61\textwidth}
        \centering
        \includegraphics[width=\linewidth]{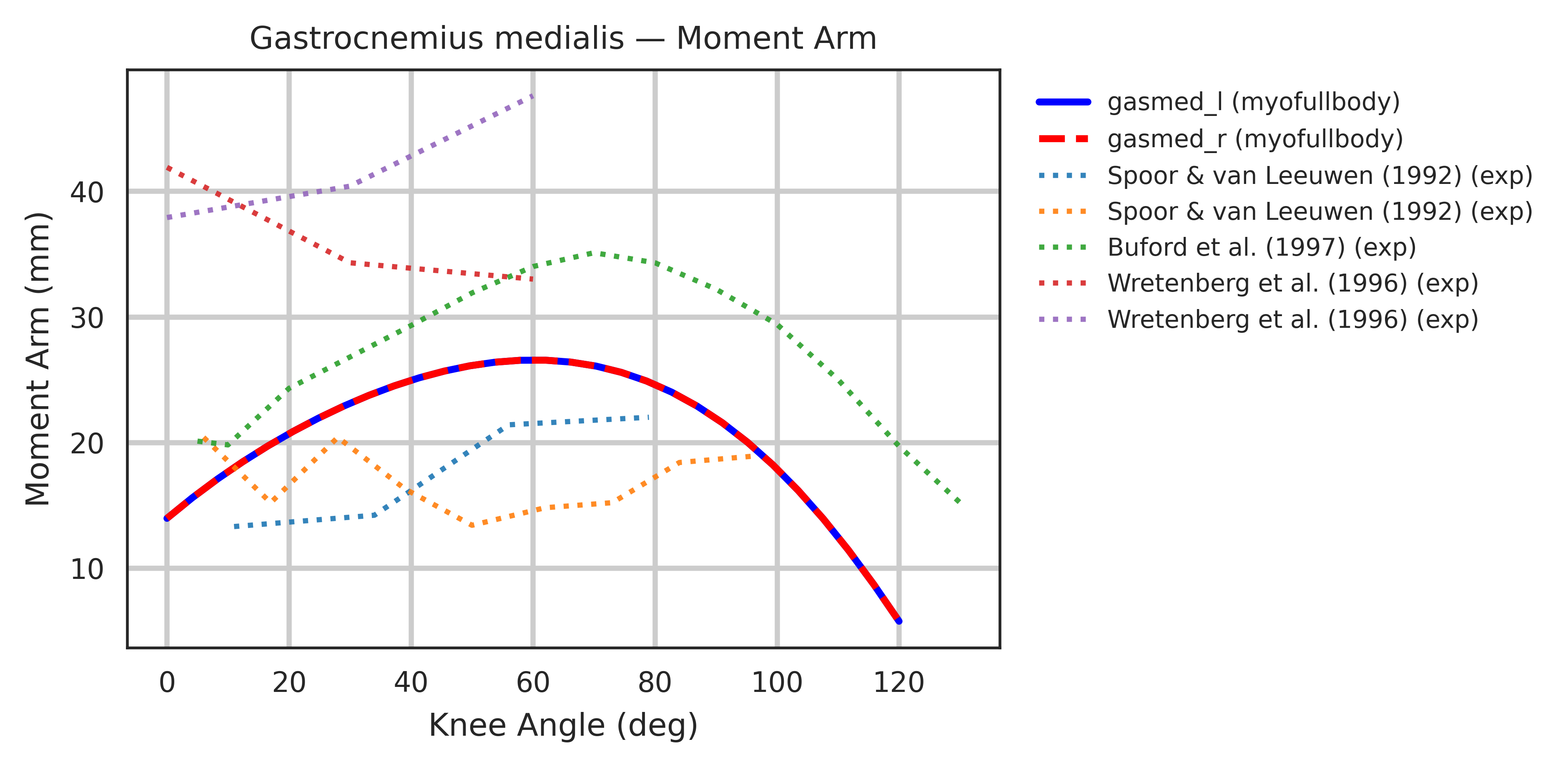}
        \label{fig:comparexp4}
    \end{subfigure}
    \caption{Example validation comparing MyoFullBody muscle moment arms against experimental data from prior studies for selected shoulder, elbow, and lower-limb muscles. Despite inter-individual variability in moment arms and attachment sites, our model’s profiles remain within the reported experimental ranges. (first part)}
    \label{fig:muscle_exp_comparison}
\end{figure*}

\begin{figure*}[!t]
    \ContinuedFloat
    \centering
    \begin{subfigure}[b]{0.48\textwidth}
        \centering
        \includegraphics[width=\linewidth]{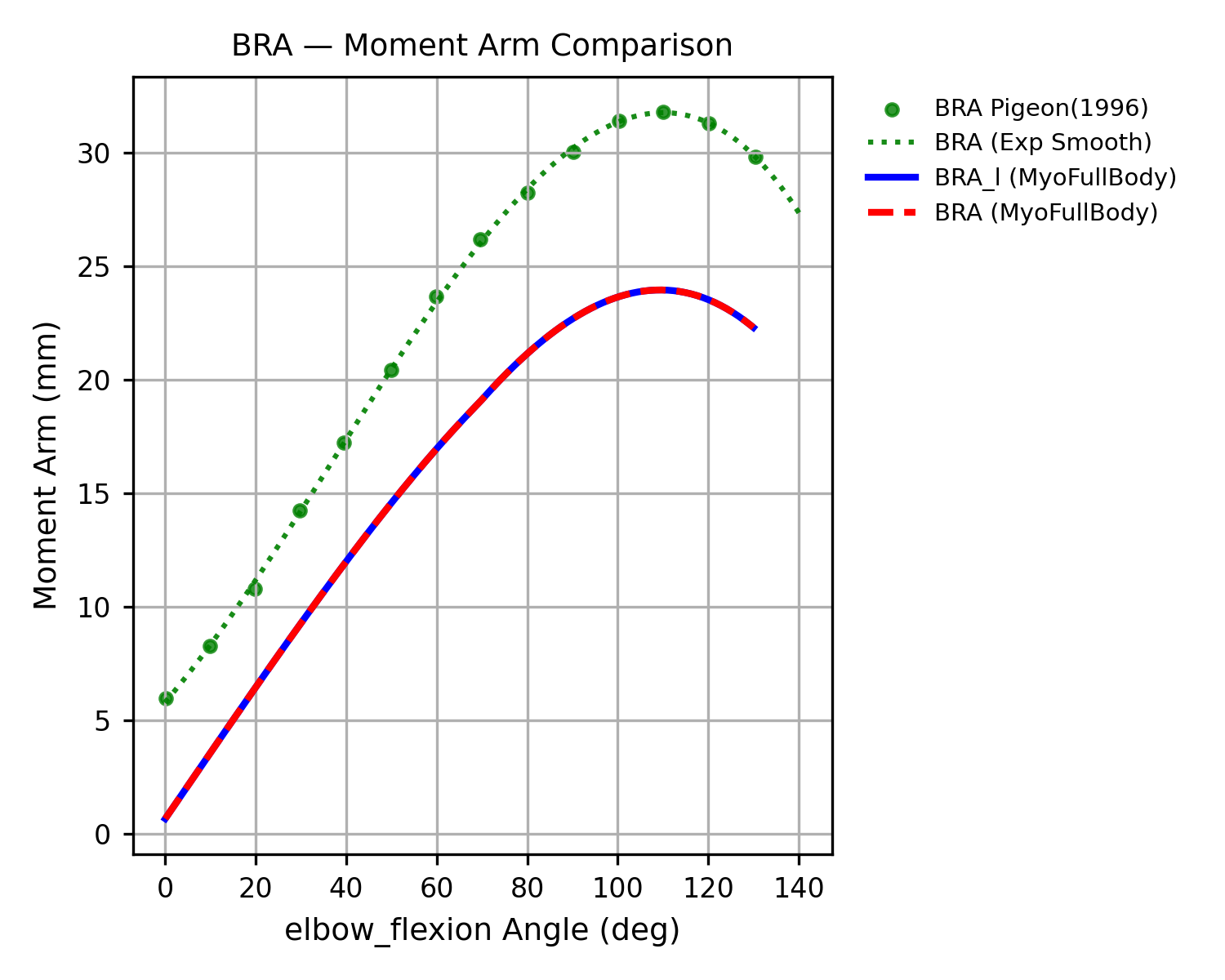}
        \label{fig:comparexp3}
    \end{subfigure}
    \hfill
    \begin{subfigure}[b]{0.48\textwidth}
        \centering
        \includegraphics[width=\linewidth]{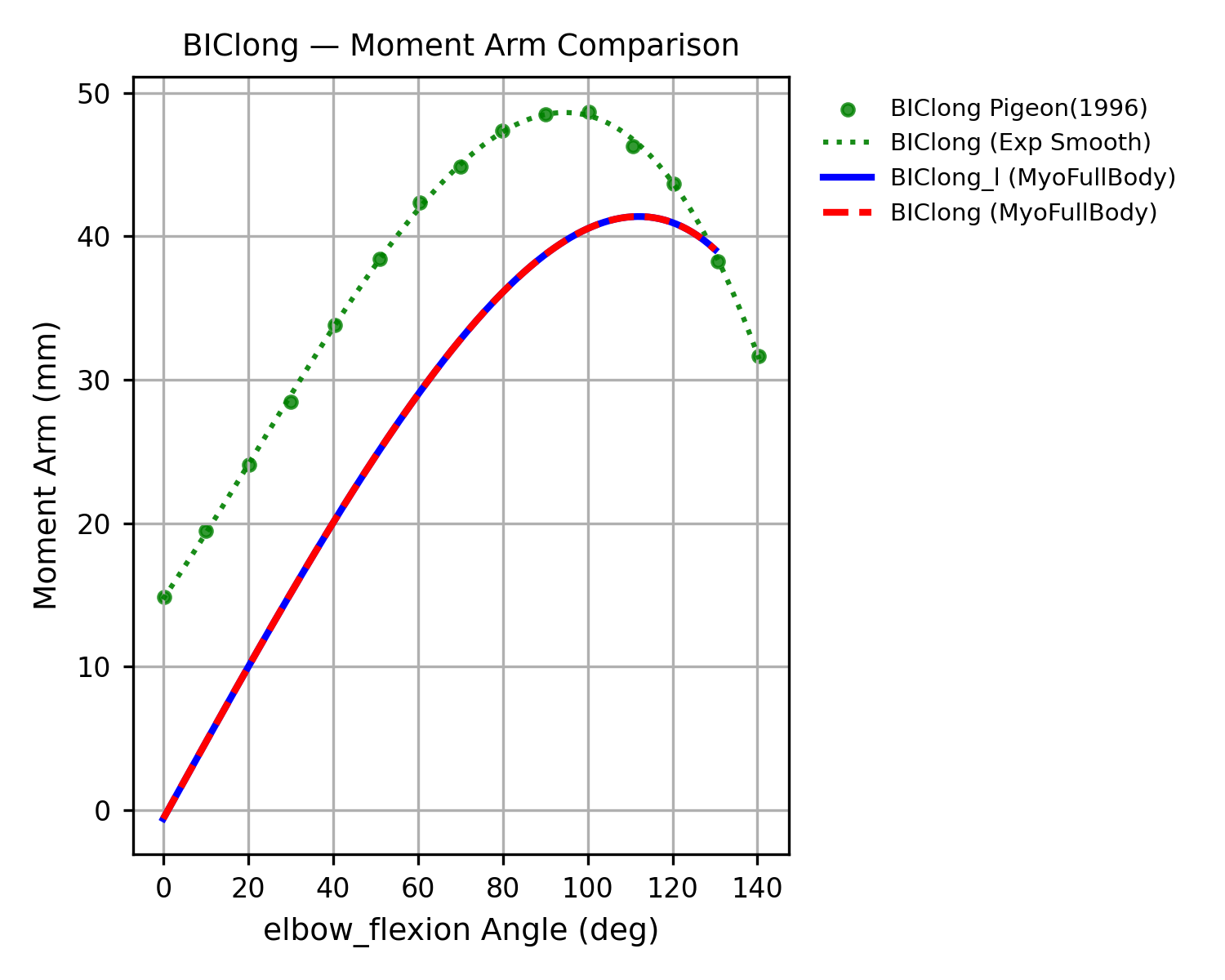}
        \label{fig:comparexp4}
    \end{subfigure}

    \vspace{0.5em}

    \begin{subfigure}[b]{0.38\textwidth}
        \centering
        \includegraphics[width=\linewidth]{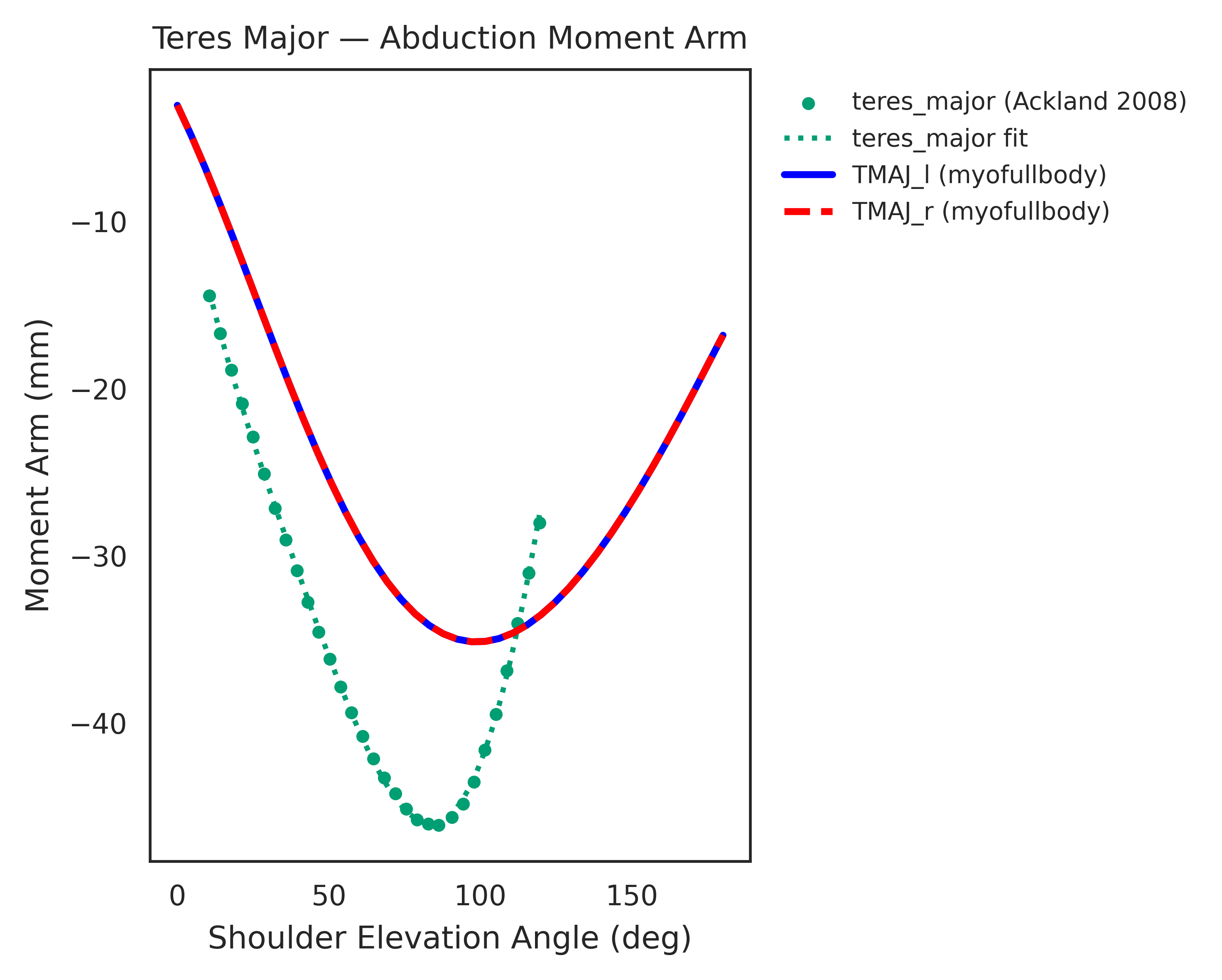}
        \label{fig:comparexp3}
    \end{subfigure}
    \hfill
    \begin{subfigure}[b]{0.61\textwidth}
        \centering
        \includegraphics[width=\linewidth]{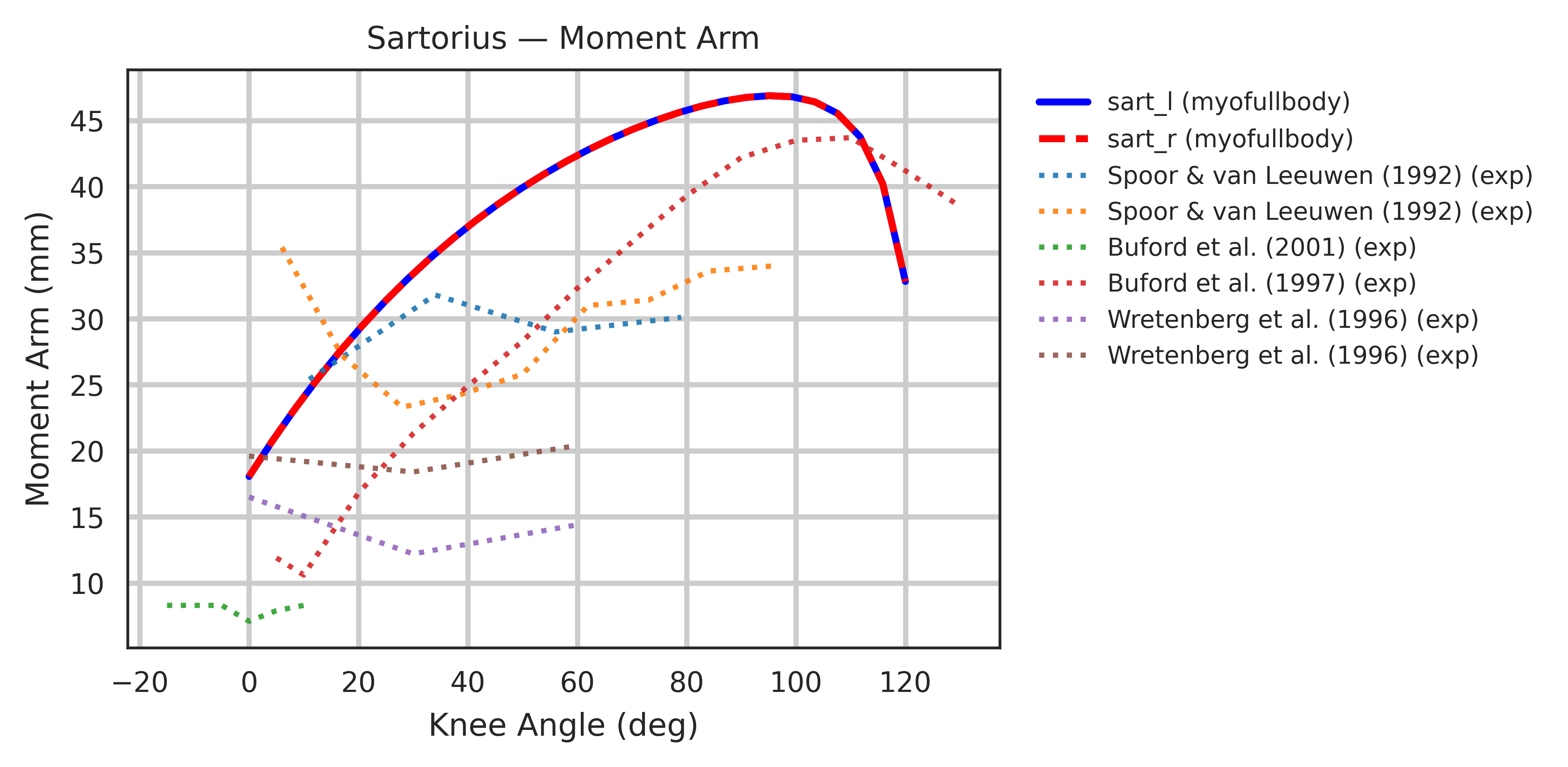}
        \label{fig:comparexp4}
    \end{subfigure}
    \caption{Example validation comparing MyoFullBody muscle moment arms against experimental data from prior studies for selected shoulder, elbow, and lower-limb muscles. Despite inter-individual variability in moment arms and attachment sites, our model’s profiles remain within the reported experimental ranges. (last part)}
    \label{fig:muscle_exp_comparison}
\end{figure*}

\section{MSK Model Parameters}\label{app:muscle_parameters}
The MyoFullBody model declares 123 joints, consisting of one free joint, 112 hinge joints, and 10 slide joints; 51 equality constraints enforce anatomical couplings, resulting in 72 independent degrees of freedom. The MyoBimanualArm model declares 76 hinge joints with 22 equality constraints, yielding 54 independent degrees of freedom. In both models, equality constraints reduce the effective control dimensionality while preserving anatomically realistic coupled motion. The segmental mass distribution of our model is summarized in Table~\ref{tab:segmental_mass_comparison}, together with corresponding values from prior biomechanical studies.

\begin{table}[h]
\centering
\caption{The MyoFullBody model is compared against biomechanical reference data from \cite{Winter2009}. In the reference studies \cite{miller1973biomechanics}, the abdominal segment is defined as spanning T12–L1 to L4–L5, while the thoracic segment spans C7–T1 to T12–L1, and the pelvic segment spans L4–L5 to the greater trochanter. In contrast, the MyoFullBody model defines the abdomen (or hereby lumbar) as L1–L5, with the thorax defined as C7 until T12 - L1. In addition, the MyoFullBody does not represent the neck as a separate segment, and its mass is therefore included within the head segment. The sacrum is classified as part of the pelvis body segment in the table below. These differences in segment definitions should be considered when interpreting discrepancies in segmental mass percentages.}
\label{tab:segmental_mass_comparison}
\begin{tabular}{lccc}
\toprule
\textbf{Body Segment} 
& \textbf{MyoFullBody (kg)} & \textbf{MyoFullBody (\%)} 
& \textbf{Winter et al., 2009 \cite{Winter2009} (\%)} 
\\
\midrule
Head (\& Neck)   &2.4 & 2.8 & 8.1 \\
Thorax           &18.6 & 22.1 & 21.6 \\
Abdomen / Lumbar &9.2 & 10.9 & 13.9  \\
Pelvis          &18.4 & 21.8 & 14.2  \\
Arms (Both)     &9.3 & 11.0 & 10.0 \\
Legs (Both)     &26.9 & 31.9 & 32.2 \\
\bottomrule
\end{tabular}
\end{table}

\section{Training Hyperparameters}\label{app:train_hparams}
We summarize the training hyperparameters of our released checkpoints that is used to evaluate on motion imitation with the MyoFullBody model in Table~\ref{tab:hyperparams_fullbody} and MyoBimanualArm model in Table~\ref{tab:hyperparams_bimanual}.

\begin{table}[h]
\centering
\caption{Training hyperparameters for MyoFullBody motion imitation pretrained checkpoint.}
\label{tab:hyperparams_fullbody}
\small
\begin{tabular}{llr}
\toprule
\textbf{Category} & \textbf{Parameter} & \textbf{Value} \\
\midrule
\multirow{4}{*}{Environment}
    & Number of parallel environments & 8,192 \\
    & Episode horizon & 1,000 steps \\
    & Total training timesteps & 15.87 billion \\
    & Backend & MuJoCo Warp \\
\midrule
\multirow{3}{*}{Network Architecture}
    & Actor hidden layers & [2048, 4096, 4096, 4096, 4096, 4096, 2048, 1024, 512] \\
    & Critic hidden layers & [2048, 4096, 4096, 4096, 4096, 4096, 2048, 1024, 512] \\
    & Activation function & SiLU \\
\midrule
\multirow{2}{*}{Architecture Features}
    & Layer normalization & Yes \\
    & Residual connections & Gated (init -2.0) \\
\midrule
\multirow{7}{*}{PPO Optimization}
    & Learning rate & $4 \times 10^{-4}$ \\
    & Learning rate schedule & Linear annealing to $4 \times 10^{-5}$ \\
    & Rollout steps per environment & 50 \\
    & Number of minibatches & 32 \\
    & Minibatch size & 12,800 \\
    & Discount factor $\gamma$ & 0.99 \\
    & GAE parameter $\lambda$ & 0.95 \\
    & Policy clip coefficient & 0.2 \\
    & Value function clip coefficient & 0.2 \\
    & Max gradient norm & 1.0 \\
    & Weight decay & 0 \\
    & Update epochs per iteration & 1 \\
    & Initial std deviation & 3.0 \\
    & Learnable std & Yes \\
    & Entropy coefficient & 0.0 \\
    & Value function coefficient & 0.5 \\
\midrule
\multirow{5}{*}{Reward Weights ($w_\cdot$)}
    & Site position $w_p$ & 0.6 \\
    & Joint position $w_q$ & 0.1 \\
    & Root velocity $w_{v_{\text{root}}}$ & 0.1 \\
    & Site orientation $w_\theta$ & 0.01 \\
    & Site velocity $w_v$ & 0.1 \\
    & Joint velocity $w_{\dot{q}}$ & 0.1 \\
\midrule
\multirow{2}{*}{Termination}
    & Mean site deviation threshold & 0.5 m \\
    & Root deviation threshold & 0.5 m \\
\bottomrule
\end{tabular}
\end{table}

\begin{table}[h]
\centering
\caption{Training hyperparameters for MyoBimanualArm motion imitation pretrained checkpoint.}
\label{tab:hyperparams_bimanual}
\small
\begin{tabular}{llr}
\toprule
\textbf{Category} & \textbf{Parameter} & \textbf{Value} \\
\midrule
\multirow{4}{*}{Environment}
    & Number of parallel environments & 8,192 \\
    & Episode horizon & 1,000 steps \\
    & Total training timesteps & 2.048 billion \\
    & Backend & MuJoCo Warp \\
\midrule
\multirow{3}{*}{Network Architecture}
    & Actor hidden layers & \multicolumn{1}{c}{$16$ layers: $12{\times}1024 \to 3{\times}2048 \to 1024$} \\
    & Critic hidden layers & \multicolumn{1}{c}{$16$ layers: $12{\times}1024 \to 3{\times}2048 \to 1024$} \\
    & Activation function & SiLU \\
\midrule
\multirow{2}{*}{Architecture Features}
    & Layer normalization & Yes \\
    & Residual connections & Not specified \\
\midrule
\multirow{7}{*}{PPO Optimization}
    & Learning rate & $4 \times 10^{-4}$ \\
    & Learning rate schedule & Warmup cosine \\
    & Rollout steps per environment & 10 \\
    & Number of minibatches & 32 \\
    & Minibatch size & 2,560 \\
    & Discount factor $\gamma$ & 0.99 \\
    & GAE parameter $\lambda$ & 0.95 \\
    & Policy clip coefficient & 0.2 \\
    & Value function clip coefficient & 0.2 \\
    & Max gradient norm & 1.0 \\
    & Weight decay & 0.001 \\
    & Update epochs per iteration & 1 \\
    & Initial std deviation & 0.2 \\
    & Learnable std & Yes \\
    & Entropy coefficient & 0.0 \\
    & Value function coefficient & 0.5 \\
\midrule
\multirow{5}{*}{Reward Weights ($w_\cdot$)}
    & Site position $w_p$ & 0.6 \\
    & Joint position $w_q$ & 0.1 \\
    & Site orientation $w_\theta$ & 0.1 \\
    & Site velocity $w_v$ & 0.1 \\
    & Joint velocity $w_{\dot{q}}$ & 0.1 \\
    & Root velocity $w_{v_{\text{root}}}$ & 0.0 \\
\midrule
\multirow{1}{*}{Termination}
    & Mean site deviation threshold & 0.25 m \\
\bottomrule
\end{tabular}
\end{table}

\section{Validation Metrics}
\label{app:validation-metrics}

This section defines the evaluation metrics reported in Table~\ref{tab:validation-metrics}. All metrics are computed by comparing simulated trajectories against the reference motion after applying the same root-frame alignment used during training.

\begin{itemize}
    \setlength{\itemsep}{0pt}
    \setlength{\parskip}{0pt}
    \setlength{\parsep}{0pt}
    
    \item \textbf{Success rate:} Fraction of episodes that complete the entire episode without exceeding thresholds on mean relative site error or root position deviation.

    \item \textbf{Joint angle error:} Root-mean-square (RMS) error between simulated and reference joint angle (excluding the root). Quaternion joints are compared using angular distance.

    \item \textbf{Joint velocity error:} RMS error between simulated and reference joint velocities (excluding the root).

    \item \textbf{Root position error:} RMS Euclidean distance between simulated and reference root positions in world coordinates after removing the initial XYZ offset.

    \item \textbf{Root yaw error:} Absolute wrapped angular difference between simulated and reference root yaw angles.

    \item \textbf{Relative site position error:} RMS error of site positions expressed in the root frame, measuring articulated-body geometric consistency.

    \item \textbf{Absolute site position error:} Mean Euclidean distance between world-frame site positions after initial root alignment.

    \item \textbf{Mean episode length:} Average number of simulation steps completed per episode.

    \item \textbf{Mean episode return:} Average cumulative reward per episode using the same reward formulation as training.
\end{itemize}

\bibliography{sample}
\end{document}